\documentclass{article} 
\usepackage[final]{neurips_2020}

\usepackage{amsmath,amsfonts,bm}

\def\1{\bm{1}}

\def\vg{{\bm{g}}}

\DeclareMathAlphabet{\mathsfit}{\encodingdefault}{\sfdefault}{m}{sl}
\SetMathAlphabet{\mathsfit}{bold}{\encodingdefault}{\sfdefault}{bx}{n}

\newcommand{\E}{\mathbb{E}}

\newcommand{\R}{\mathbb{R}}

\usepackage{url}
\usepackage{amsthm,amsfonts,braket}
\usepackage{amsmath}
\usepackage{algorithm,algorithmic}
\usepackage{booktabs}
\usepackage{multirow, subcaption, adjustbox}
\usepackage{graphicx}
\usepackage{mathtools}
\usepackage{amssymb}
\usepackage{color}
\usepackage{wrapfig}
\usepackage{enumitem}

\usepackage[dvipsnames]{xcolor}
\definecolor{darkblue}{rgb}{0, 0, 0.5}
\usepackage[colorlinks=true,linkcolor=darkblue,citecolor=darkblue,urlcolor=darkblue]{hyperref}



\newtheorem{thm}{Theorem}[section]
\newtheorem{defi}{Definition}[section]
\newtheorem{prp}{Proposition}[section]
\newtheorem{lem}{Lemma}[section]

\newtheorem{assumption}{Assumption}[section]
\newtheorem{cor}{Corollary}[section]

\title{Stein Self-Repulsive Dynamics: Benefits from Past Samples}

\author{%
  \large Mao Ye \thanks{Equal Contribution} \\
  UT Austin \\
  \texttt{my21@cs.utexas.edu} \\
  \And
  \large Tongzheng Ren \textsuperscript{*} \\
  UT Austin\\
  \texttt{tongzheng@utexas.edu} \\
  \And
  \large Qiang Liu \\
  UT Austin \\
  \texttt{lqiang@cs.utexas.edu} \\
}

\begin{document}
\maketitle

\global\long\def\th{\boldsymbol{\theta}}%
\global\long\def\Th{\boldsymbol{\Theta}}%
\global\long\def\lt{\underline{t}}%
\global\long\def\ls{\underline{s}}%
\global\long\def\bth{\bar{\th}}%
\global\long\def\tth{\tilde{\th}}%
\vspace{-2em}
\begin{abstract}
We propose a new Stein self-repulsive dynamics for obtaining diversified samples from intractable un-normalized distributions. 
Our idea is to introduce Stein variational gradient as a repulsive force to push the samples of Langevin dynamics away from the past trajectories. This simple idea allows us to significantly decrease the auto-correlation in Langevin dynamics and hence increase the effective sample size. Importantly, 
as we establish in our theoretical analysis, 
the asymptotic stationary distribution remains correct even with the addition of the repulsive force, 
thanks to the special properties of the Stein variational gradient. 
We perform extensive 
empirical studies of our new algorithm, showing that our method yields much higher sample efficiency and better uncertainty estimation than vanilla Langevin dynamics. 
\end{abstract}
\vspace{-1em}
\section{Introduction}

Drawing samples from complex un-normalized distributions 
is one of the most basic problems in statistics and machine learning, 
with broad applications to enormous research fields that rely on probabilistic modeling. 
%
Over the past decades, large amounts of methods have been 
proposed for approximate sampling,  
including both Markov Chain Monte Carlo (MCMC)
\citep[e.g.,][]{brooks2011handbook} and variational inference \citep[e.g.,][]{wainwright2008graphical, blei2017variational}.

MCMC works by simulating Markov chains whose stationary distributions match the distributions of interest. 
Despite nice asymptotic theoretical properties, 
MCMC is widely criticized for its slow convergence rate in practice.  
In difficult problems, 
the samples drawn from MCMC are often found to have high auto-correlation across time,  
meaning that the Markov chains explore very slowly in the configuration space. 
When this happens, 
the samples returned by MCMC only approximate a small local region, and under-estimate the probability of the regions un-explored by the chain.  

Stein variational gradient descent (SVGD) \citep{liu2016stein} 
is a different type of approximate sampling methods 
designed to overcome the limitation of MCMC. 
Instead of drawing random samples sequentially, 
SVGD evolves a pre-defined number of particles (or sample points) in parallel 
with a special interacting particle system to match the distribution of interest by minimizing the KL divergence. 
In SVGD, the particles 
interact with each other to 
simultaneously move towards the high probability regions following the gradient direction, 
and also move away from each other due to a special repulsive force. 
As a result, SVGD allows us to obtain diversified samples 
that correctly represent the variation of the distribution of interest. 
SVGD has found applications in various challenging problems 
\citep[e.g.,][]{feng2017learning, haarnoja2017reinforcement, pu2017vae, liu2017steinP, gong2019quantile}. 
See    \citet[e.g.,][]{han2018stein, DBLP:conf/uai/ChenZWLC18, liu2019understanding, wang2019stein} for examples of extensions. 

However, one problem of SVGD is that it theoretically requires to run an infinite number of chains in parallel
in order to approximate the target distribution asymptotically \citep{liu2017stein}.  
With a finite number of particles,
the fixed point of SVGD does still provide a \emph{prioritized, partial} approximation to the distribution in terms of the expectation of a special case of functions  \citep{liu2018stein}. Nevertheless, it is still desirable to develop a variant of ``single-chain SVGD'', which only requires to run a single chain sequentially like MCMC to achieve the correct stationary distribution asymptotically in time, with no need to take the limit of infinite number of parallel particles.

In this work, 
we propose an example of \emph{single-chain SVGD} 
by integrating the special repulsive mechanism of SVGD with gradient-based MCMC such as Langevin dynamics. Our idea is to
use repulsive term of SVGD to enforce the samples in MCMC
away from the past samples visited at previous iterations. 
Such a new \emph{self-repulsive dynamics} allows us to decrease the auto-correlation in MCMC and hence increase the mixing rate, but still obtain the same stationary distribution thanks to the special property of the SVGD repulsive mechanism. 

We provide thorough theoretical analysis of our new method, 
establishing its asymptotic convergence to the target distribution. 
Such result is highly non-trivial, as our new self-repulsive dynamic is a non-linear high-order Markov process. 
Empirically, 
we evaluate our methods on an array of challenging sampling tasks, showing that our method yields much better uncertainty estimation and larger effective sample size.

\section{Background: Langevin dynamics \& SVGD} 
\label{sec:background}
This section gives 
a brief introduction on Langevin dynamics \citep{rossky1978brownian} and Stein Variational Gradient Descent (SVGD) \citep{liu2016stein}, which we integrate together to develop our new self-repulsive dynamics for more efficient sampling.  

\textbf{Langevin Dynamics} \quad
Langevin dynamics is 
a basic gradient based MCMC method. 
For a given target distribution supported on $\R^d$ with density function  $\rho^*(\th)\propto \exp({-V(\th)})$, 
where $V\colon \R^d \mapsto \R $ is the potential function, 
the (Euler-discrerized) Langevin dynamics simulates a Markov chain with the following rule: 
\begin{equation*}
\th_{k+1}=\th_{k}-\eta \nabla V(\th_{k})+\sqrt{2\eta}\boldsymbol{e}_{k},
~~~~~~\boldsymbol{e}_{k}\sim\mathcal{N}(0,\mathbf{I}), 
\label{eq:disLD}
\end{equation*}
where $k$ denotes the number of iterations, $ 
\{\boldsymbol{e}_k\}_{k=1}^\infty$ are independent standard Gaussian noise,  and $\eta$ is a step size parameter. 
It is well known that the limiting distribution of $\th_k$ when  $k\to \infty$ approximates the target distribution when $\eta$ is sufficiently small. 

Because the updates in Langevin dynamics 
are local and incremental, new points generated by the dynamics can be highly correlated to the past sample, 
in which case
we need to run Langevin dynamics sufficiently long 
in order to obtain diverse samples.

\textbf{Stein Variatinal Gradient Descent (SVGD)} \quad
Different from Langevin dynamics, 
SVGD iteratively evolves a pre-defined number of particles in parallel. Starting from an initial set of particles $\{\th_0^i\}_{i=1}^M$, 
SVGD updates the $M$ particles in parallel  by 
\begin{equation*}
\th_{k+1}^{i}=\th_{k}^{i}+\eta
\vg(\th_{k}^{i}; ~\hat{\delta}_k^M),  ~~~~\forall i =1,\ldots, M, 
\end{equation*}

where $\vg(\boldsymbol\theta_k^i; ~ \hat{\delta}_k^M)$ is 
a velocity field that depends the empirical distribution of the current set of particles 
$\hat{\delta}_k^M := \frac{1}{M}\sum_{j=1}^{M}\delta_{\th_{k}^{j}}$ in the following way, 
\begin{align*} 
\vg(\th_k^i;\hat{\delta}_k^M) 
=
\E_{\th \sim \hat \delta_k^M}\!\!\bigg [ \underset{\mathrm{Confining\ Term}}{\text{\ensuremath{\underbrace{-K(\th,\th_{k}^i) \nabla V(\th)}}}}+\underset{\mathrm{Repulsive\ Term}}{\underbrace{\nabla_{\th}K(\th,\th_{k}^i)}} \!\bigg]\!\!.
\end{align*}
Here $\delta_{\th}$ is the Dirac measure centered at $\th$, 
and hence $\E_{\th \sim \hat \delta_k^M}[\cdot]$ denotes averaging on the particles.  
The $K(\cdot,\cdot)$ is a positive definite kernel, such as the RBF kernel, that can be specified by users. 

Note that $\vg(\th_k^i;~ \hat{\delta}_k^M)$ consists of 
a confining term and repulsive term: 
the confining term pushes particles to move towards high density region, and the repulsive term prevents the particles from colliding with each other. It is the balance of these two terms that allows us to asymptotically approximate the target  distribution $\rho^*(\th)\propto \exp(-V(\th))$ at the fixed point,   when the number of particles goes to infinite. 
We refer the readers to \citet{liu2016stein, liu2017stein, liu2018stein} for thorough theoretical justifications of SVGD. But a quick, informal way to justify the SVGD update is 
through the \emph{Stein's identity}, which shows that  
the velocity field $\vg(\th;~ \rho)$ equals zero when $\rho$ equals the true distribution $\rho^*$, that is, $\forall \th^\prime\in \R^d$,
\begin{eqnarray} 
\label{equ:steinidentity}
\vg(\th^\prime; \rho^*) = \E_{\th\sim \rho^*} \left [ 
-K(\th,\th') \nabla V(\th) +  \nabla_{\th}K(\th,\th^\prime)\right ] = 0.
\end{eqnarray}

This equation shows that, the target distributions forms a fixed point of the update, and SVGD would converge if the particle distribution $\hat \delta_{k}^M$ gives a close approximation to the target distribution $\rho^*$. 

\section{Stein Self-Repulsive Dynamics}
\label{sec:methods}
In this work, we propose to integrate Langevin dynamics and SVGD to simultaneously decrease the auto-correlation of Langevin dynamics and eliminate the need for running parallel chains in SVGD. 
The idea is to use Stein repulsive force between the the current particle and the particles from previous iterations,  
hence forming a new self-avoiding dynamics with fast convergence speed. 

Specifically, assume we run a single Markov chain like Langevin dynamics, where $\th_k$ denotes the sample at the $k$-th iteration. Denote by 
$\tilde \delta_{k}^M$ the empirical measure of $M$ samples from the past iterations: 
\begin{equation*}
\small
\tilde \delta_{k}^M := \frac{1}{M}\sum_{j=1}^M \delta_{\th_{k-j c_\eta}}, ~~~~~~~~
c_\eta = c/\eta, 
\end{equation*}
where $c_\eta$ is a thinning factor, which scales inversely with the step size $\eta$, introduced to slim the sequence of past samples. 
Compared with the $\hat \delta_k^M$ in SVGD, which is averaged over $M$ parallel particles,
$\tilde \delta_k^M$ is averaged across time over $M$ past samples. 
Given this, our Stein self-repulsive dynamics updates the sample via 
\begin{align} \label{equ:self}
\th_{k+1} \gets \th_{k}~ + ~
\underbrace{(-\eta V(\th_k)  + \sqrt{2\eta} \boldsymbol e_k)}_{\text{Langevin}} ~+~ 
\underbrace{\eta \alpha \vg(\th_k;~\tilde \delta_{k}^M)}_{\text{Stein Repulsive}}, 
\end{align}
in which the particle is updated with the typical Langevin gradient, 
plus a Stein repulsive force against the particles from the previous iterations. Here $\alpha \geq 0$ is a parameter that controls the magnitude of the Stein repulsive term. 
In this way, the particles are pushed away from the past samples, 
and hence admits lower auto-correlation and 
faster convergence speed. 
Importantly, 
the addition of the repulsive force \emph{does not impact} the asymptotic stationary distribution, thanks to Stein's identity in \eqref{equ:steinidentity}. 
This is because if the self-repulsive dynamics has converged to the target distribution $\rho^*$, 
such that $\theta_k\sim \rho^*$ for all $k$,
the Stein self-repulsive term would equal to zero in expectation due to Stein's identity and hence does not introduce additional bias over Langevin dynamics. Rigorous theoretical analysis of this idea is developed in Section~\ref{sec:theoretical}.

\textbf{Practical Algorithm} \quad
Because $\tilde \delta_{k}^M$ is averaged across the past samples, it is necessary to introduce a burn-in phase with the repulsive dynamics. Therefore, our overall procedure works as follows,  

\vspace{-1.5em}
\begin{align}
\th_{k+1} 
= \begin{cases}
\th_{k}\!-\!\eta\nabla V(\th_{k})\!+\!\sqrt{2\eta}\boldsymbol{e}_{k}, & k<Mc_\eta,\\
\th_{k}\!+\!\eta\left[-\nabla V(\th_{k})\!+\!\alpha \vg(\th_k; \tilde \delta_{k}^M) 
\right]\!+\!\sqrt{2\eta}\boldsymbol{e}_{k}, & k\geq Mc_\eta. 
\end{cases}
\label{eq:srld}
\end{align}
It includes two phases. The first phase is the same as the Langevin dynamics which collects the initial $M$ samples used in the second phase while serves as a warm start. 
The repulsive gradient update is  introduced in the second phase to encourage the dynamics to visit the under-explored density region.  
We call this particular instance of our algorithm Self-Repulsive Langevin dynamics (SRLD),
self-repulsive variants of more general dynamics is discussed in Section~\ref{sec:extend}.  

\textbf{Remark} \quad
Note that the first phase is introduced to collect the initial $M$ samples. However, it's not really necessary to generate the initial $M$ samples with Langevin dynamics. We can simply use some other initialization distribution and get $M$ initial samples from that distribution. In practice, we find using Langevin dynamics to collect the initial samples is natural and it can also be viewed as the burn-in phase before sampling, so we use \eqref{eq:srld} in all of the other experiments.

\textbf{Remark} \quad
The general idea of introducing self-repulsive terms inside MCMC or other iterative algorithms is not new itself. For example, 
in molecular dynamics simulations, 
an algorithm called metadynamics \citep{laio2002escaping} has been widely used, 
in which the particles are repelled away from the past samples in a way similar to our method, 
but with a typical repulsive function, such as $\sum_{j} D(\theta_k, \theta_{k-jc_\eta})$, where $D(\cdot,\cdot)$ can be any kind of dis-similarity. However, introducing an arbitrary repulsive force would alter the stationary distribution of the dynamics, introducing a harmful bias into the algorithm. Besides, the self-adjusting mechanism in \citet{deng2020contour} can also be viewed as a repulsive force using the multiplier in gradient. The key highlight of our approach, as reflected by our theoretical results in Section~\ref{sec:theoretical}, is the unique property of the Stein repulsive term, that allows us to obtain the correct stationary distribution even with the addition of the repulsive force. 

\textbf{Remark} \quad
Recent works \citep{gallego2018stochastic,zhang2018stochastic} also combine SVGD with Langevin dynamics, in which, however, the Langevin force is directly added to a set of particles that evolve in parallel with SVGD. Using our terminology, their system is 
\begin{align*}
    \th_{k+1}^i = \th_k^i + 
{(-\eta V(\th_k^i)  + \sqrt{2\eta} \boldsymbol e_k^i)} 
+ 
 {\eta \alpha \vg(\th_k^i;~\hat \delta_{k}^M)},\quad
 \boldsymbol{e}_{k}\sim\mathcal{N}(0,\mathbf{I}),
 \quad \forall i=1,\ldots, M. 
\end{align*}
This is significantly different from our method on both motivation and practical algorithm. 
Their algorithm still requires to simulate $M$ parallel chains of particles  like SVGD, and was proposed to obtain easier theoretical analysis than the deterministic dynamics of SVGD. 
Our work is instead  motivated by the practical need of decreasing the auto-correlation in Langevin dynamics, 
and avoiding the need of running multiple chains in SVGD, and hence must be based on \emph{self-repulsion} against past samples along a single chain.   

\begin{wrapfigure}{r}{0.6\textwidth}
\vspace{-1\baselineskip}
\begin{centering}
\includegraphics[width=0.5\linewidth]{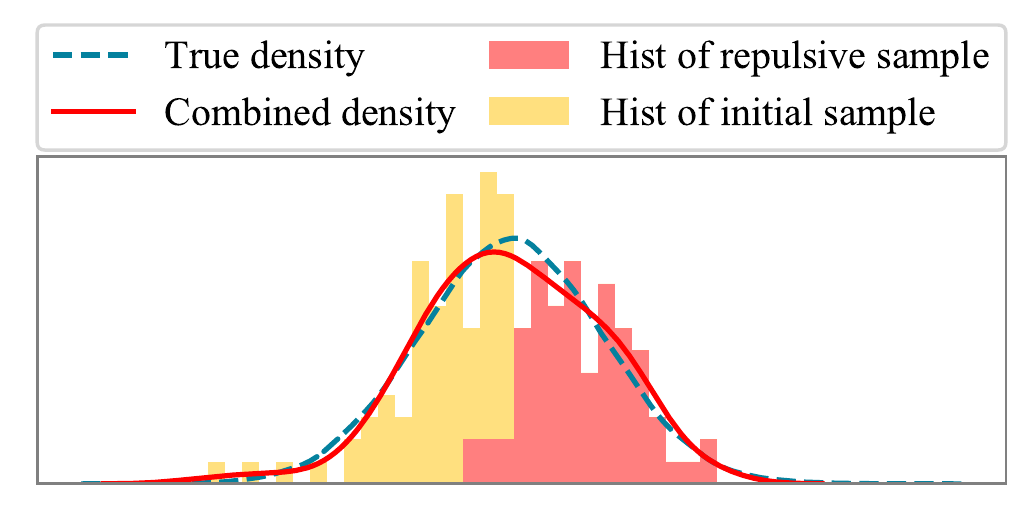}\includegraphics[width=0.5\linewidth]{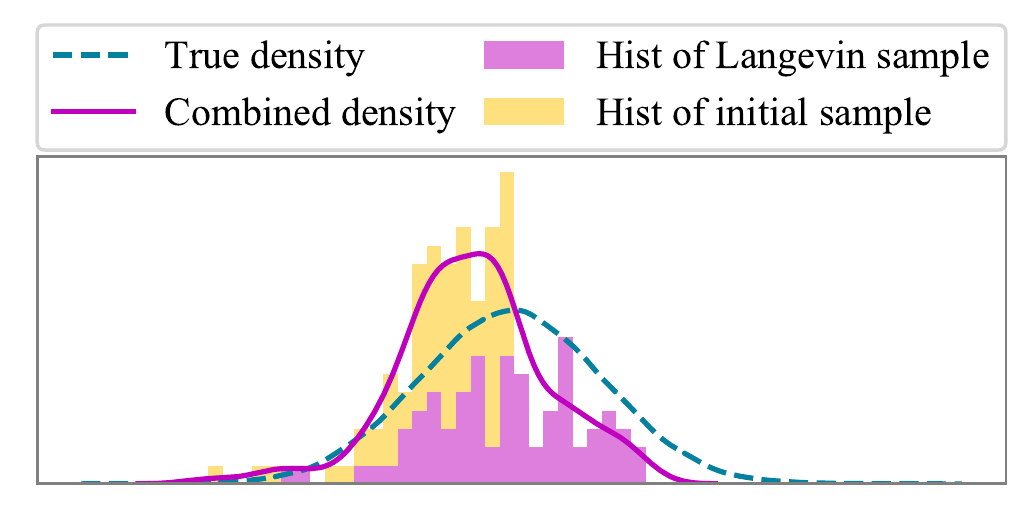}
\par\end{centering}
\begin{centering}
\includegraphics[width=0.5\linewidth]{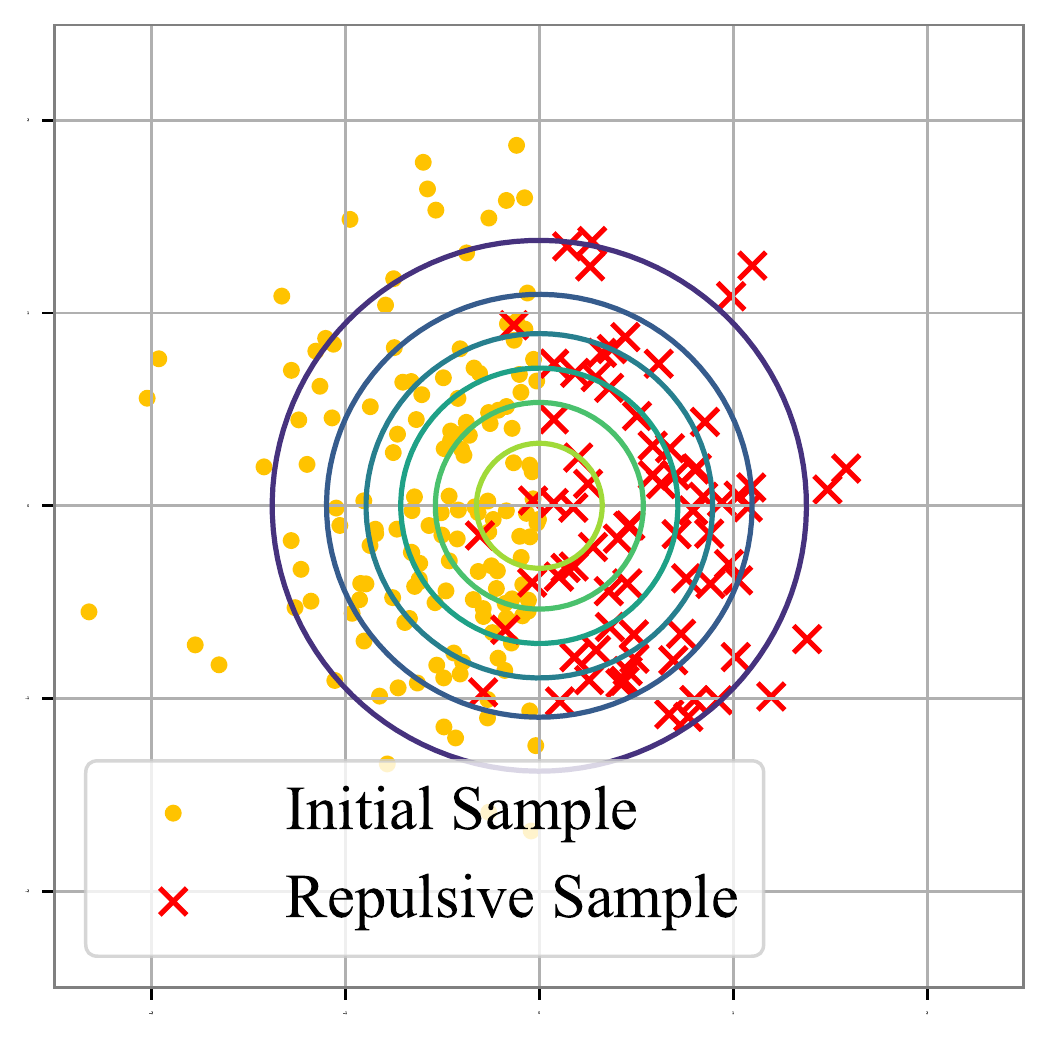}\includegraphics[width=0.5\linewidth]{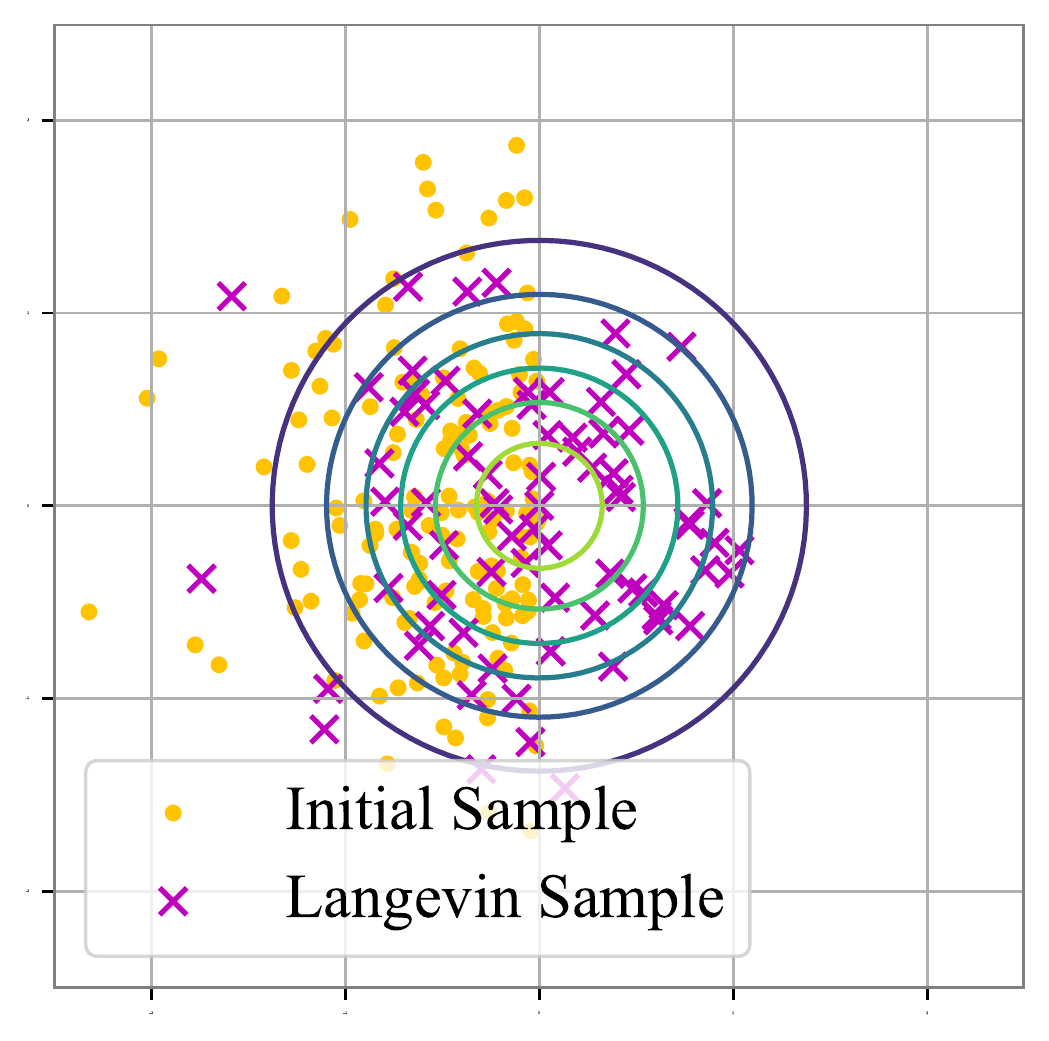}
\par\end{centering}
\caption{
Illustrating the advantage of our Self-Repulsive Langevin dynamics. 
With a set of initial examples locating on the left part of the target distribution (yellow dots), Self-Repulsive Langevin dynamics is forced to explore the right part more frequently, yielding an accurate approximation when combined with the initial samples. 
Langevin dynamics, however, does not take the past samples into account and yields a poor overall approximation.}\label{fig:toy-example}
\end{wrapfigure}

\textbf{An Illustrative Example} \quad
We give an illustrative example to show the key advantage of our self-repulsive dynamics. 
Assume that we want to sample from a bi-variate Gaussian distribution shown in Figure~\ref{fig:toy-example}. 
Unlike 
standard settings,  
we assume that we have already obtained some initial samples (yellow
dots in Figure \ref{fig:toy-example}) before running the dynamics. 
The initial samples are assumed to concentrate on the left part of the target distribution as shown in Figure~\ref{fig:toy-example}. 
In this extreme case, since the left part of the distribution 
is over-explored by the initial samples,  it is desirable to have the subsequent new samples 
to concentrate more on the un-explored part of the distribution. 
However, standard Langevin dynamics does not take this into account, and hence yielding a biased overall representation of the true distribution (left panel). 
With our self-repulsive dynamics, 
the new samples are forced to explore the un-explored region more frequently, allowing us to obtain a much more accurate approximation when combining the new and initial samples. 

\section{Theoretical Analysis of Stein Self-Repulsive Dynamics} \label{sec:theoretical} 
We provide theoretical analysis of the self-repulsive dynamics. 
We establish that our self-repulsive dynamics converges to the correct target distribution asymptotically,
in the limit when 
particle size
$M$ approaches to infinite and 
 the step size $\eta$ approaches to $0$.  
This is a highly non-trivial task, as the self-repulsive dynamics is 
a highly complex, non-linear and high order Markov stochastic process. 
We attack this problem by breaking the proof into the following three steps:
\vspace{-0.5em}
\begin{enumerate}[leftmargin=1.5em, label=(\arabic*)]
    \item At the limit of $M\to\infty$ (called the \textbf{mean field limit}), 
    we show that 
    {practical dynamics} in \eqref{eq:srld} 
    is closely approximated by a \textbf{discrete-time mean-field dynamics} characterized by \eqref{eq:dy2}. 
    \item By further taking the limit of $\eta\to 0^+$ (called the \textbf{continuous-time limit}), 
    the dynamics in \eqref{eq:dy2} converges to a \textbf{continuous-time mean-field dynamics} characterized by \eqref{eq:dy3}. 
    \item We show that the dynamics in \eqref{eq:dy3} converges to the target distribution. 
\end{enumerate}
\vspace{-0.5em}

\textbf{Remark} \quad
As we mentioned in Section \ref{sec:methods}, we introduce the first phase to collect the initial $M$ samples for the second phase, and our theoretical analysis follows this setting to make our theory as close to the practice. However, the theoretical analysis can be generalized to the setting of drawing $M$ initial samples from some initialization distribution with almost identical argument.

\textbf{Notations} \quad We use $\left\Vert \cdot\right\Vert $
and $\left\langle \cdot,\cdot\right\rangle $ to represent the $\ell_{2}$
vector norm and inner product,  respectively. The
Lipschitz norm and bounded Lipschitz norm of a function $f$ are defined by $\left\Vert f\right\Vert _{\mathrm{Lip}}$
and $\left\Vert f\right\Vert _{\mathrm{BL}}$. The KL divergence,
Wasserstein-2 distance and Bounded Lipschitz distance between distribution $\rho_{1}$, $\rho_{2}$ are denoted
as $\mathbb{D}_{\mathrm{KL}}[\rho_{1}\|\rho_{2}]$, $\mathbb{W}_{2}[\rho_{1},\rho_{2}]$ and $\mathbb{D}_{\mathrm{BL}}[\rho_{1},\rho_{2}]$, respectively.

\subsection{Mean-Field and Continuous-Time Limits} 
To fix the notation, 
we denote by  $\rho_{k}:=\mathrm{Law}(\th_{k})$ the distribution of $\th_{k}$ at time $k$ of the practical self-repulsive dynamics \eqref{eq:srld}, which we refer to the \textbf{practical dynamics} in the sequel, when the initial particle $\th_0$ is drawn from an initial continuous distribution $\rho_0$.    
Note that given $\rho_0$, the subsequent $\rho_k$ can be recursively defined through dynamics \eqref{eq:srld}. 
Due to the diffusion noise in Langevin dynamics, all $\rho_k$ are continuous distributions supported on $\R^d$.  
We now introduce the limit dynamics when we take
the mean-field limit 
$(M\to +\infty)$ and then the continuous-time limit $(\eta  \to 0^+)$. 

\textbf{Discrete-Time Mean-Field Dynamics ($M\to +\infty$)}
In the limit of $M\to\infty$, 
our practical dynamics \eqref{eq:srld} 
approaches to 
the following limit dynamics,
in which the delta measures  
on the particles are replaced by the actual continuous distributions 
of the particles,  
\begin{align}
\tilde{\th}_{k+1} 
= \begin{cases}
\tilde{\th}_{k}\!-\!\eta\nabla V(\tilde{\th}_{k})\!+\!\sqrt{2\eta}\boldsymbol{e}_{k}, & k\le Mc_\eta,\\
\tilde{\th}_{k}\!+\!\eta\left[-\nabla V(\tilde{\th}_{k})\!+\!\alpha \vg(\tilde \th_k, \tilde \rho_{k}^M)
\right]\!+\!\sqrt{2\eta}\boldsymbol{e}_{k}, & k\ge Mc_\eta.
\end{cases}\label{eq:dy2}
\end{align}
where $\tilde \rho_k^M =\frac{1}{M}\sum_{j=1}^{M}\tilde{\rho}_{k-jc_\eta}$ and  $\tilde{\rho}_k := \mathrm{Law}(\tilde{\th}_{k})$ is the (smooth) distribution of $\tth_k$ at time-step $k$ when the dynamics is initialized with $\tth_0\sim \tilde \rho_0 = \rho_0$. 
Compared with the practical dynamics in \eqref{eq:srld}, the difference is that the empirical distribution $\tilde \delta_{k}^M$ is replaced by the smooth distribution $\tilde \rho_k^M$.  
Similar to the recursive definition of $\rho_k$ following dynamics \eqref{eq:srld}, 
$\tilde \rho_k$ is also recursively defined through dynamics \eqref{eq:dy2}, starting from $\tilde \rho_0=\rho_0$. 

As we show in Theorem \ref{thm: particle}, if the auto-correlation of $\th_{k}$ decays fast enough and $M$ is sufficiently large, $\tilde {\rho}_{k}^M$ is well approximated by the empirical distribution $\tilde \delta_{k}^M$ in \eqref{eq:srld}, and further the two dynamics (\eqref{eq:srld} and \eqref{eq:dy2}) converges to each other in the sense that $\mathbb{W}_{2}[\rho_{k},\tilde{\rho}_{k}]\to0$
as $M\to\infty$ for any $k$. Note that in taking the limit of $M\to\infty$, we need to ensure that we run the dynamics for more than $Mc_\eta$ steps. Otherwise, SRLD degenerates to Langevin dynamics as we stop the chain before we finish collecting the $M$ samples.

\textbf{Continuous-Time Mean-Field Dynamics ($\eta\to 0^+$)} \quad
In the limit of zero step size $(\eta\to 0^+)$, the discrete-time mean field dynamics in \eqref{eq:dy2} can be shown to 
converge to the following continuous-time mean-field dynamics: 
\begin{align}
d\bar{\th}_{t} & =\begin{cases}
-\nabla V(\bar{\th}_{t})dt+d\mathcal{B}_{t}, & t\in[0,Mc),\\
\left[-\nabla V(\bar{\th}_{t})+\alpha
\vg(\bar \th_t, ~ \bar \rho_t^M)
\right]dt+d\mathcal{B}_{t}, & t\ge Mc.
\end{cases}\label{eq:dy3}
\end{align}
where $\bar \rho_t^M :=
\frac{1}{M}\sum_{j=1}^{M}\bar{\rho}_{t-jc}(\cdot)$, $\mathcal{B}_{t}$ is the Brownian motion and 
$\bar{\rho}_{t} =\mathrm{Law}\left(\bar{\th}_{t}\right)$ is the distribution of $\bth_t$ at a continuous time point $t$ with $\th_0$ initialized by  
$\bar{\th}_{0}\sim\tilde{\rho}_{0}=\rho_0$. 
We prove that 
\eqref{eq:dy3} is closely approximated by 
\eqref{eq:dy2} with small step size in the sense that 
$\mathbb{D}_{\mathrm{KL}}[\tilde{\rho}_{k}\| \bar{\rho}_{k\eta}]\to0$ as $\eta\to 0$ in Theorem \ref{thm: discrete}, and importantly, the stationary distribution of \eqref{eq:dy3} equals to the target distribution  $\rho^*(\th) \propto \exp(-V(\th)).$
\subsection{Assumptions}
We first introduce the techinical assumptions used in our theoretical analysis.

\begin{assumption} [RBF Kernel]\label{asm: RBF}

We use RBF kernel, i.e. $K(\th_{1},\th_{2})=\exp({-\left\Vert \th_{1}-\th_{2}\right\Vert ^{2}/\sigma})$,
for some fixed $0<\sigma<\infty$.
\end{assumption}
We only assume the RBF kernel for the simplicity of our analysis. However, it is straightforward to generalize our theoretical result to other positive definite kernels.

\begin{assumption} [$V$ is dissipative and smooth] \label{asm: dis}

Assume that $\left\langle \th,-\nabla V(\th)\right\rangle \le b_{1}-a_{1}\left\Vert \th\right\Vert ^{2}$
and $\left\Vert \nabla V(\th_{1})-\nabla V(\th_{2})\right\Vert \le b_{1}\left\Vert \th_{1}-\th_{2}\right\Vert $.
We also assume that $\left\Vert \nabla V(\boldsymbol{0})\right\Vert \le b_{1}$.
Here $a_{1}$ and $b_{1}$ are some finite positive constant.
\end{assumption}

\begin{assumption} [Regularity Condition] \label{asm: reg}

Assume $\mathbb{E}_{\th\sim\rho_{0}}[\left\Vert \th\right\Vert ^{2}]>0$.
Define $\rho_k^M = \sum_{j=1}^M \rho_{k-jc_\eta}/M$, assume there exists $a_2, B<\infty$ such that 
\vspace{-0.5em}
\begin{small}
\begin{equation*}
\sup_{k\ge Mc_{\eta}}\frac{\E\left\Vert g(\boldsymbol{\theta}_{k};\tilde{\delta}_{k}^{M})-g(\boldsymbol{\theta}_{k};\rho_{k}^{M})\right\Vert ^{2}}{\sup_{\left\Vert \boldsymbol{\theta}\right\Vert \le B}\E\left\Vert g(\boldsymbol{\theta};\tilde{\delta}_{k}^{M})-g(\boldsymbol{\theta};\rho_{k}^{M})\right\Vert ^{2}}\le a_{2}.
\end{equation*}
\end{small}
\end{assumption}
\vspace{-0.5em}

\begin{assumption} [Strong-convexity] \label{asm: logcon}

Suppose that $\left\langle \nabla V(\th_{1})-\nabla V(\th_{2}),\th_{1}-\th_{2}\right\rangle \geq L\left\Vert \th_{1}-\th_{2}\right\Vert ^{2}$ for a positive constant $L$. 
\end{assumption}

\textbf{Remark} Assumption \ref{asm: dis} is standard in the existing Langevin dynamics analysis \citep[see][]{dalalyan2017theoretical,DBLP:conf/colt/RaginskyRT17,deng2020non}. 
Assumption \ref{asm: reg} is a weak condition as it assumes that the dynamics can not degenerate into one local mode and stop moving anymore.
This assumption is expected to be true when we have diffusion terms like the Gaussian noises 
in our self-repulsive dynamics.  
%
Assumption \ref{asm: logcon} is a classical assumption on the existing Langevin dynamics analysis with convex potential \cite{dalalyan2017theoretical,durmus2019analysis}. 
Although being a bit strong, 
this assumption broadly applies to posterior inference problem in the limit of big data, 
as the posterior distribution converges to Gaussian distributions for large training set as shown by Bernstein-von Mises theorem. 
It is technically possible to 
further generalize our results to the non-convex settings with a refined analysis, 
which we leave as future work. This work focuses on the classic convex setting for simplicity.
\subsection{Main Theorems}
All of the proofs in this section can be found in Appendix \ref{sec:theory_apx}. We first prove that the limiting distribution of the continuous-time mean field dynamics \eqref{eq:dy3} is the target distribution.
This is achieved by writing  dynamics \eqref{eq:dy3}
into the following (non-linear) partial differential equation: 
\begin{small}
\begin{equation*}
\partial_{t}\bar{\rho}_{t}=\begin{cases}
\nabla\cdot\left(-\nabla V\bar{\rho}_{t}\right)+\Delta\bar{\rho}_{t} & t\in[0,Mc)\\
\nabla\cdot\left[\left(-\nabla V+
\alpha \vg(\cdot, \bar{\rho}_t^M)\right)
\bar{\rho}_{t}
\right]+\Delta\bar{\rho}_{t}, & t\ge Mc.
\end{cases}
\label{eq:continuity_equation}
\end{equation*}
\end{small}
\begin{thm}[Stationary Distribution] \label{thm: stationary} 
Given some finite $M$, $c$ and $\alpha$, and suppose that the limit
distribution of \eqref{eq:dy3} exists. Then the limit distribution is unique and satisfies $\rho^*(\th) \propto \exp({-V(\th)})$. 
\end{thm}

We then give the upper bound on the discretization error, which can be characterized
by analyzing the KL divergence between $\tilde{\rho}_{k}$
and $\bar{\rho}_{k\eta}$. 

\begin{thm} [Time Discretization Error] \label{thm: discrete}
Given some sufficiently small step size $\eta$ and choose $\alpha < a_2/(2b_1 + 4/\sigma)$. Under Assumption \ref{asm: RBF},
\ref{asm: dis}, \ref{asm: reg} and $c_\eta = c/\eta$. 
we have for some constant $C$, 
\begin{small}
\begin{align*}
    \underset{l\in\{0,...,k\}}{\max}\ \mathbb{D}_{\mathrm{KL}}\left[\bar{\rho}_{l\eta}\|\tilde{\rho}_{l}\right]
\le \begin{cases}
\mathcal{O}\left( \eta +  k \eta^2\right) & k\le Mc_\eta-1\\
\mathcal{O}\left(\eta + Mc\eta +\alpha^{2}Mce^{C\alpha^{2}(k\eta-Mc)}\eta^{2}
\right) & k\ge Mc_\eta. 
\end{cases}
\end{align*}
\end{small}
\vspace{-1.5em}
\end{thm}
With this theorem, we can know that if $\eta$ is small enough, then the discretization error is small and $\tilde{\rho}$ approximates $\bar{\rho}$ closely. Next we give result on the mean field limit of $M\to\infty$.
\begin{thm} [Mean-Field Limit]\label{thm: particle} 
Under Assumption \ref{asm: RBF}, \ref{asm: dis}, \ref{asm: reg}, and \ref{asm: logcon}, suppose that we choose $\alpha$ and $\eta$ such that 
\textcolor{black}{$-(a_{1}-2\alpha b_{1}/\sigma) +\eta b_{1}<0$; $\frac{2\alpha\eta}{\sigma}(b_{1}+1)<1$; 
$a_{2}-\alpha\left(2b_{1}+\frac{4}{\sigma}\right)>0$};
Then there exists a constant $c_2$, 
such that when $L/a \geq c_2$ and 
we have 
\begin{small}
\begin{eqnarray*}
\mathbb{W}_{2}^{2}[\rho_{k},\tilde{\rho}_{k}] = \begin{cases}\mathcal{O}\left({\alpha^{2}/{M}}+\eta^{2}\right) & \ge Mc_\eta, \\
0 & k \le Mc_\eta-1. 
\end{cases}
\end{eqnarray*}
\end{small}
\vspace{-1.5em}
\end{thm}
Thus, if $M$ is sufficiently large, $\rho_k$ can well approximate the $\tilde{\rho}_k$. 
Combining all the above theorems, we have the following Corollary showing the convergence of the proposed practical algorithm to the target distribution.

\begin{cor}[Convergence to Target Distribution]
Under the assumptions of Theorem \ref{thm: stationary}, \ref{thm: discrete} and \ref{thm: particle}, by choosing $k,\eta,M$ such that $k\eta\to\infty$, $\exp(C\alpha^{2}k\eta)\eta^{2}=o(1)$ and $\frac{k\eta}{Mc}=\gamma\left(1+o(1)\right)$ with $\gamma>1$, we have
\begin{small}
\begin{equation*}
\lim_{k,M\to\infty,\eta\to0^{+}}\mathbb{D}_{\text{BL}}\left[\rho_{k},\rho^{*}\right]=0.
\end{equation*}
\end{small}
\vspace{-1.5em}
\end{cor}

\textbf{Remark} \quad
A careful choice of parameters is needed as our system is a complicated longitudinal particle system. Also notice that if $\gamma\le1$, the repulsive dynamics reduces to Langevin dynamics, as only the samples from the first phase will be collected.
\vspace{-0.5em}
\section{Extension to General Dynamics} \label{sec:extend}
Although we have focused on self-repulsive Langevin dynamics, our Stein self-repulsive idea can be broadly combined with general gradient-based MCMC. 
Following \citet{ma2015complete}, we consider the following general class of sampling dynamics for drawing samples from $p(\th) \propto \exp(-V(\th))$: 
\begin{align*}
& d\th_{t}  = -\boldsymbol{f}(\th)dt+\sqrt{2\boldsymbol{D}(\th)}d\mathcal{B}_{t},\\
&\text{with}\quad \boldsymbol{f}(\th) =  [\boldsymbol{D}(\th)+\boldsymbol{Q}(\th)] \nabla V(\th)-\boldsymbol{\Gamma}(\th),\quad \Gamma_{i}(\th) = \sum_{j=1}^{d}\frac{\partial}{\partial\text{\ensuremath{\th}}_{j}}\left(D_{ij}(\th)+Q_{ij}(\th)\right).
\end{align*}
where $\boldsymbol{D}$ is a positive semi-definite diffusion matrix that determines the strength of the Brownian motion and $\boldsymbol{Q}$ is a skew-symmetric curl matrix that can represent the traverse effect \citep[e.g. in][]{neal2011mcmc, DBLP:conf/nips/DingFBCSN14}.
By adding the Stein repulsive force, we obtain the following general self-repulsive dynamics 
\begin{align}
d\bar{\th}_{t} & =\begin{cases}
-\boldsymbol{f}(\th)dt+\sqrt{2\boldsymbol{D}(\th)}d\mathcal{B}_{t}, & t\in[0,Mc)\\
-\left(\boldsymbol{f}(\th)+\alpha
\vg(\bar\th_t; ~ \bar\rho_{t}^M)
\right)dt+d\mathcal{B}_{t}, & t\ge Mc
\end{cases}\label{eq:dy4}
\end{align}
where $\bar\rho_t:=\mathrm{Law}(\bar\th_t)$ is again the distribution 
of $\bar\th_t$ following \eqref{eq:dy4} when initalized at $\bar\th_0\sim\bar\rho_0$. 
Similar to the case of Langevin dynamics, 
this process also converges to the correct target distribution,  
and can be simulated by practical dynamics similar to \eqref{eq:srld}. 
\begin{thm}[Stationary Distribution] \label{thm: stationary2}

Given some finite $M$, $c$ and $\alpha$, and suppose that the limiting
distribution of dynamics \eqref{eq:dy4} exists. 
Then the limiting distribution is unique and equals the target distribution 
$\rho^*(\th)\propto \exp({-V(\th)})$. 
\end{thm}
\vspace{-0.5em}
\section{Experiments}
In this section, we evaluate the proposed method in various challenging tasks. We demonstrate the effectiveness of SRLD in  high dimensions  by applying it to sample the posterior of Bayesian Neural Networks. To demonstrate the superiority of the SRLD in obtaining diversified samples, we apply SRLD on contextual bandits problem, which requires the sampler efficiently explores the distribution in order to give good uncertainty estimation.

We include discussion on the parameter tuning and additional experiment on sampling high dimensional Gaussian and Gaussian mixture in Appendix \ref{sec:addexp}. Our code is available at \url{https://github.com/lushleaf/Stein-Repulsive-Dynamics}. 
\vspace{-0.5em}
\subsection{Synthetic Experiment} 
\label{sec:toy_example}
We first show how the repulsive gradient helps explore the whole distribution using a synthetic distribution that is easy to visualize. Following \citet{ma2015complete}, we compare the sampling efficiency on the following correlated 2D distribution with density 
\begin{equation*}
    \rho^*([\theta_1,\theta_2])\propto-\theta_{1}^{4}/10-\left(4\left(\theta_{2}+1.2\right)-\theta_{1}^{2}\right)^{2}/2.
\end{equation*}

We compare the SRLD with vanilla Langevin dynamics, and evaluate the sample quality by Maximum Mean Discrepancy (MMD) \citep{gretton2012kernel}, Wasserstein-1 Distance and effective sample size (ESS). Notice that the finite sample quality of gradient based MCMC method is highly related to the step size. Compared with Langevin dynamics, we have an extra repulsive gradient and thus we implicitly have larger step size. To rule out this effect, we set different step sizes of the two dynamics so that the gradient of the two dynamics has the same magnitude.

In addition, 
to decrease the influence of random noise, 
the two dynamics are set to have the same initialization and use the same sequence of Gaussian noise. 
We collect the sample of every iteration. 
We repeat the experiment 20 times with different initialization and sequence of Gaussian noise.

\begin{figure}[t]
\centering
    \includegraphics[scale=0.25]{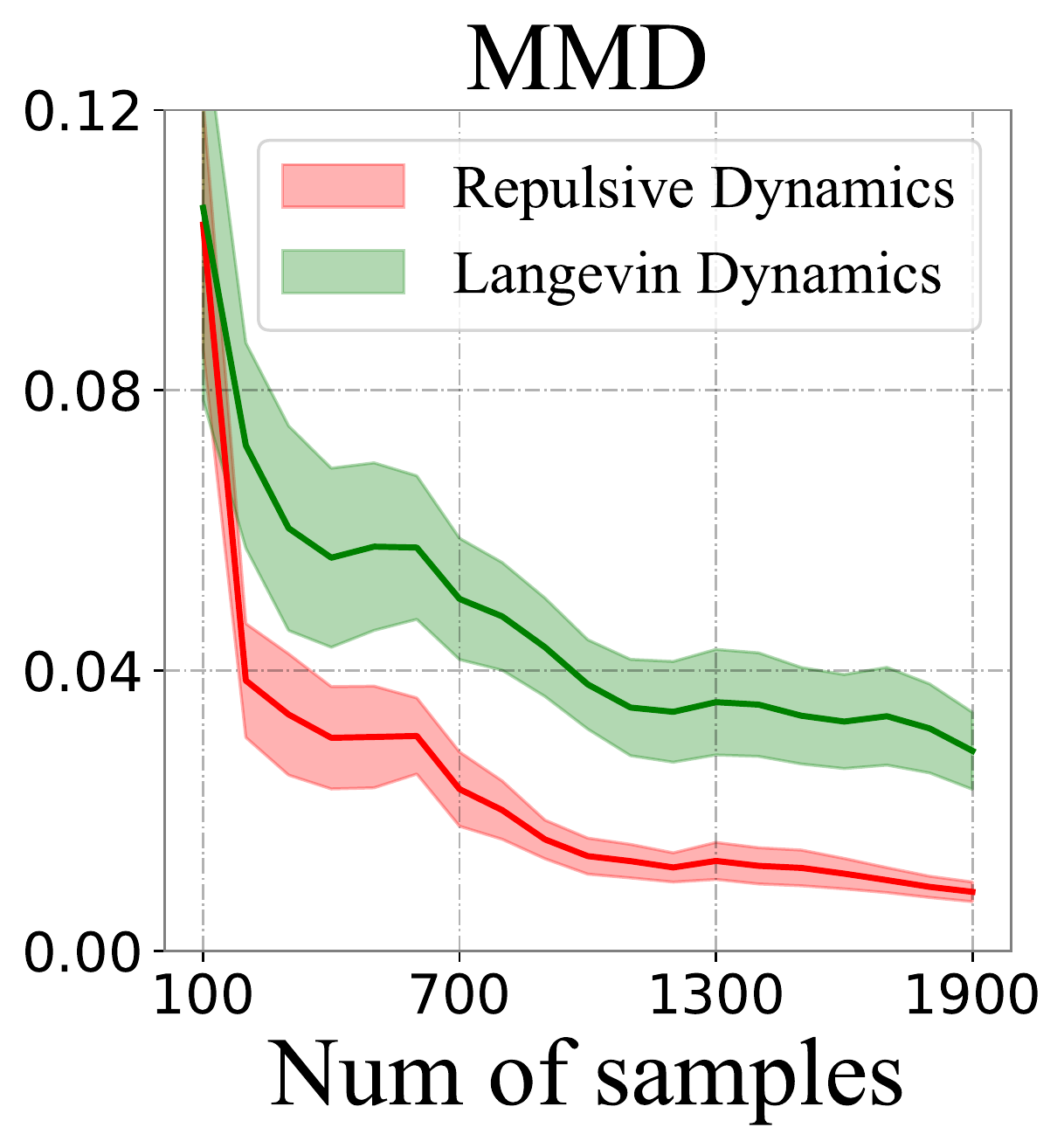}
    \includegraphics[scale=0.25]{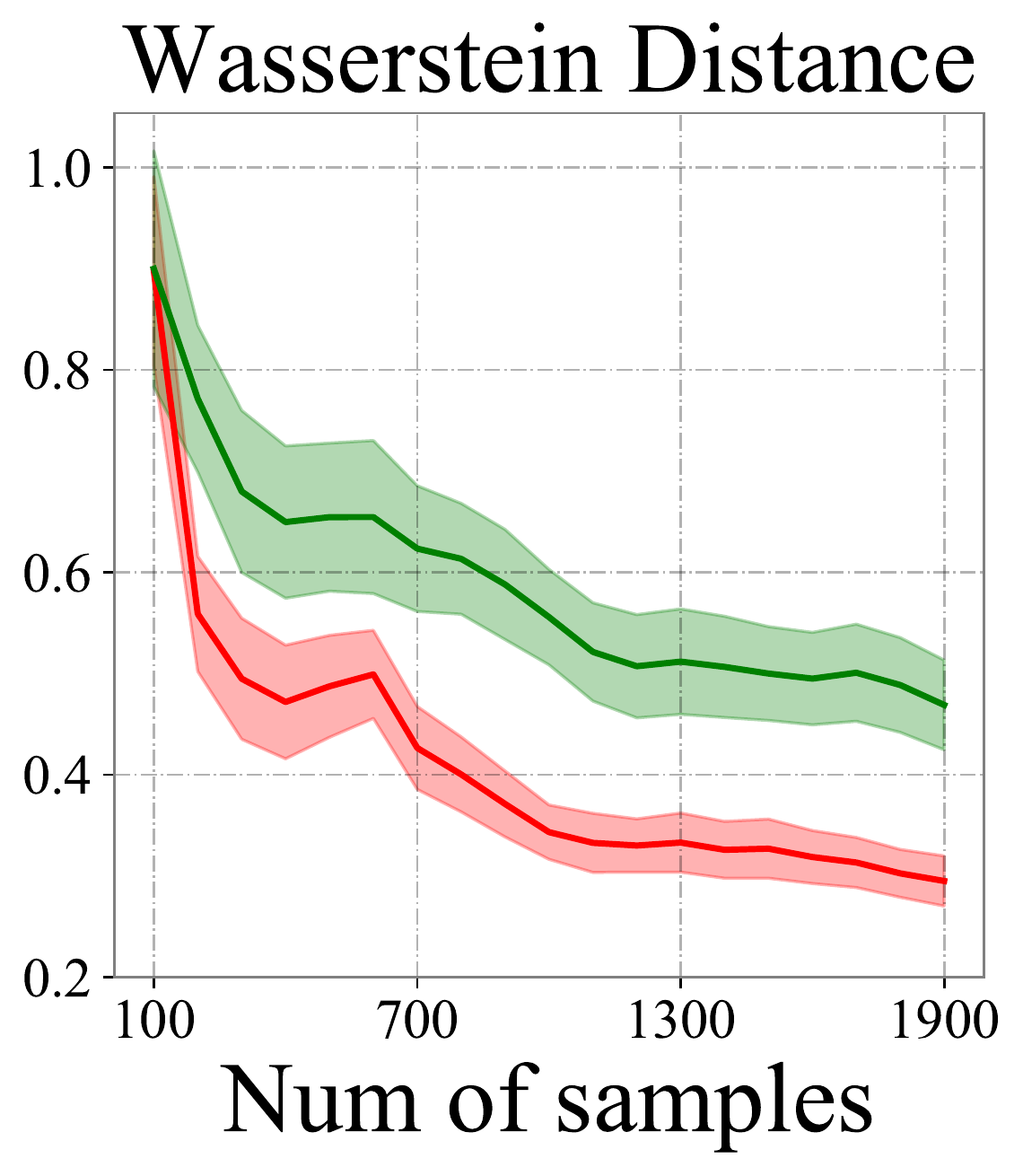}
    \includegraphics[scale=0.25]{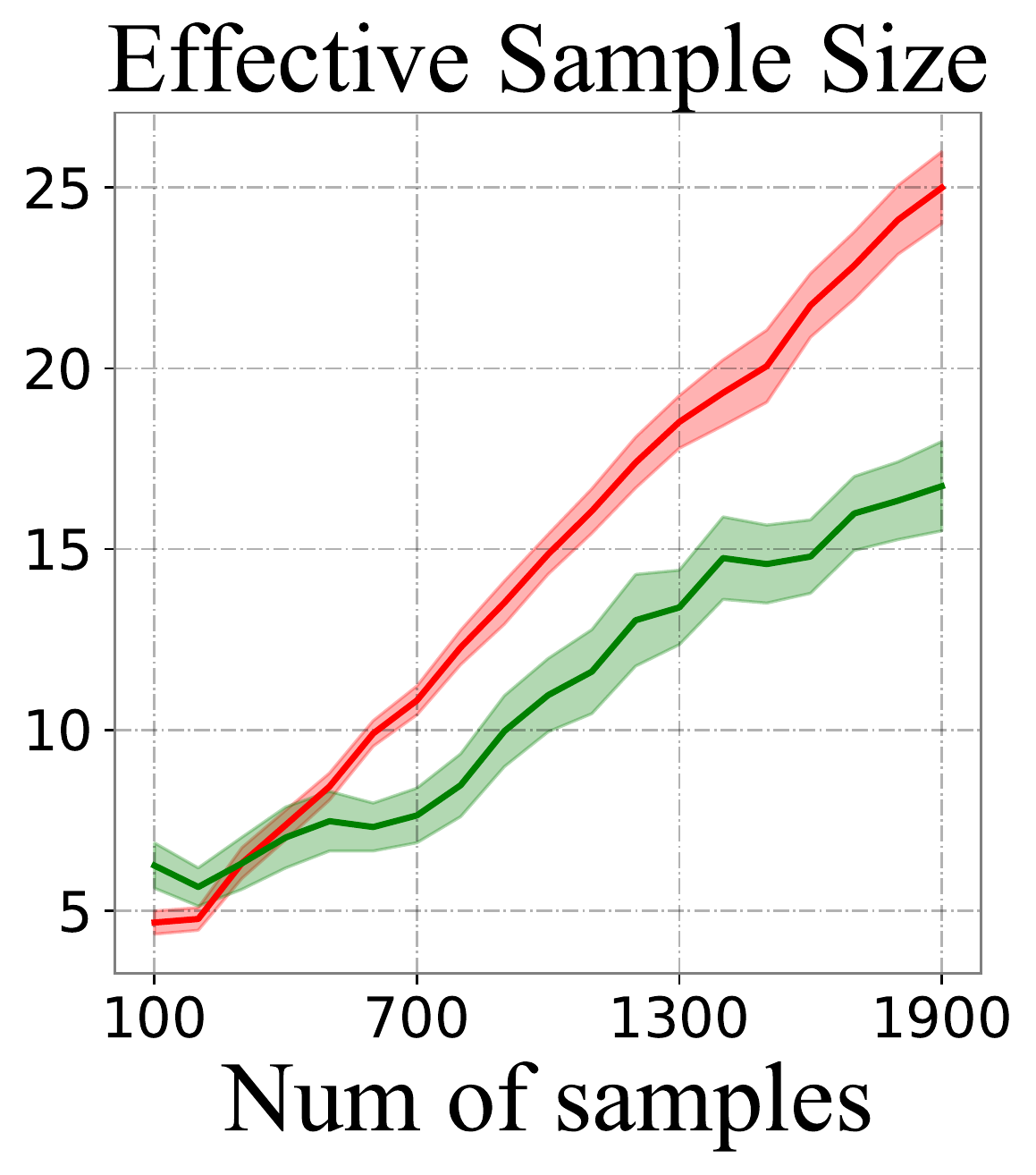}
    \includegraphics[scale=0.25]{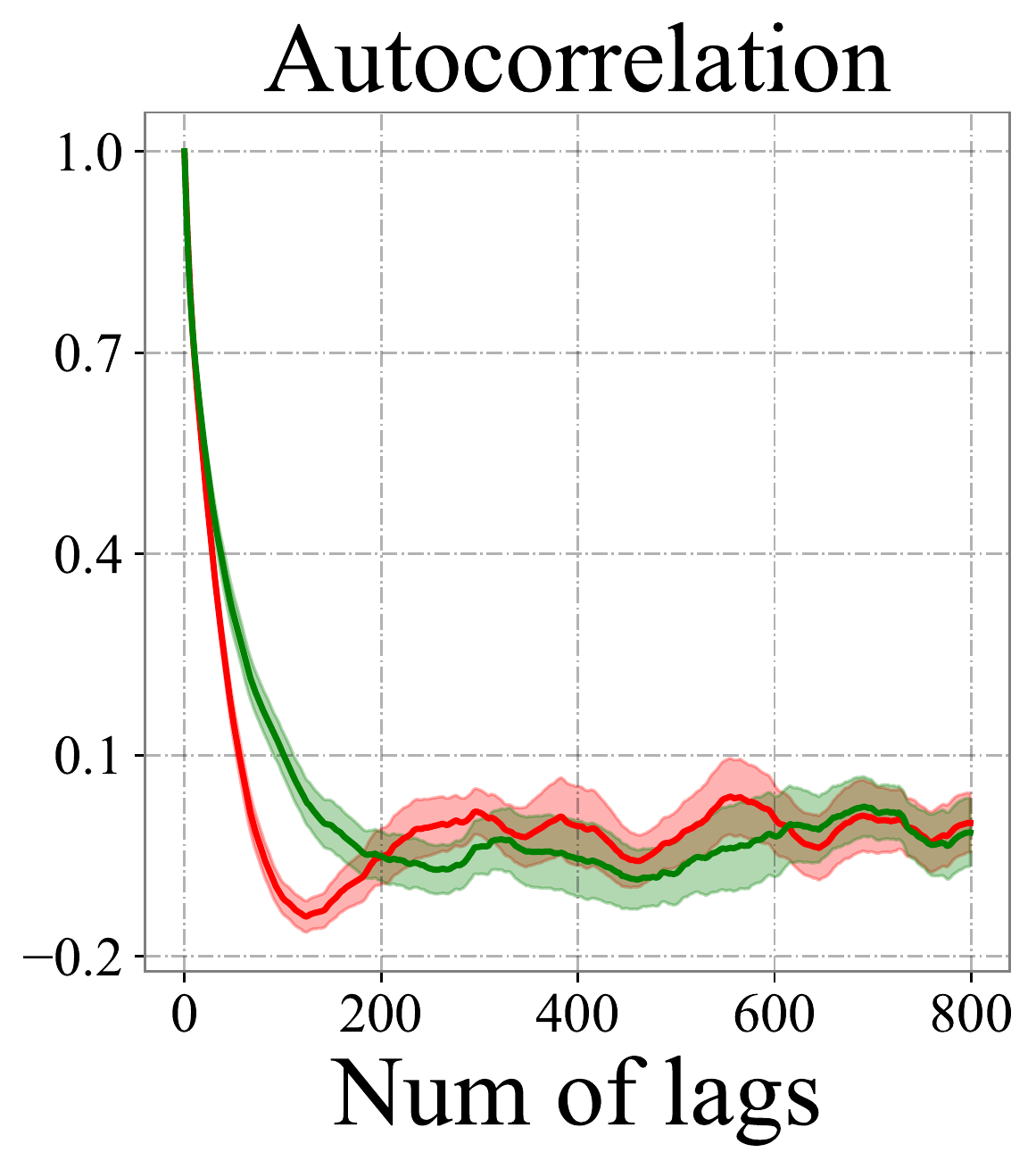}
\caption{Sample quality of SRLD and Langevin dynamics for sampling the correlated 2D distribution. The auto-correlation is the averaged auto-correlation of the two dimensions.
\label{fig:Sample-quality-corr}}
\end{figure}

\begin{figure*}[t]
\begin{centering}
\includegraphics[scale=0.2]{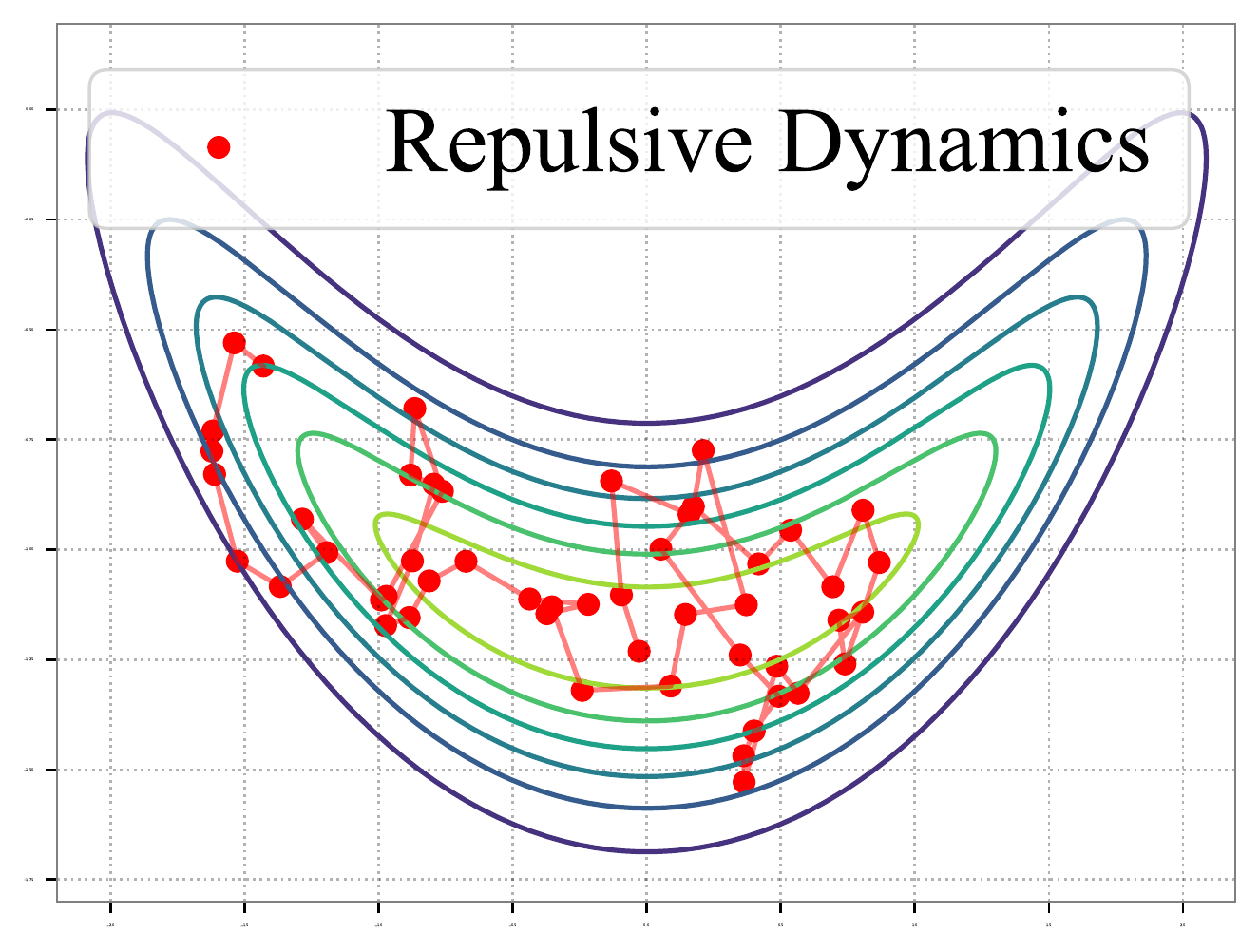}
\includegraphics[scale=0.2]{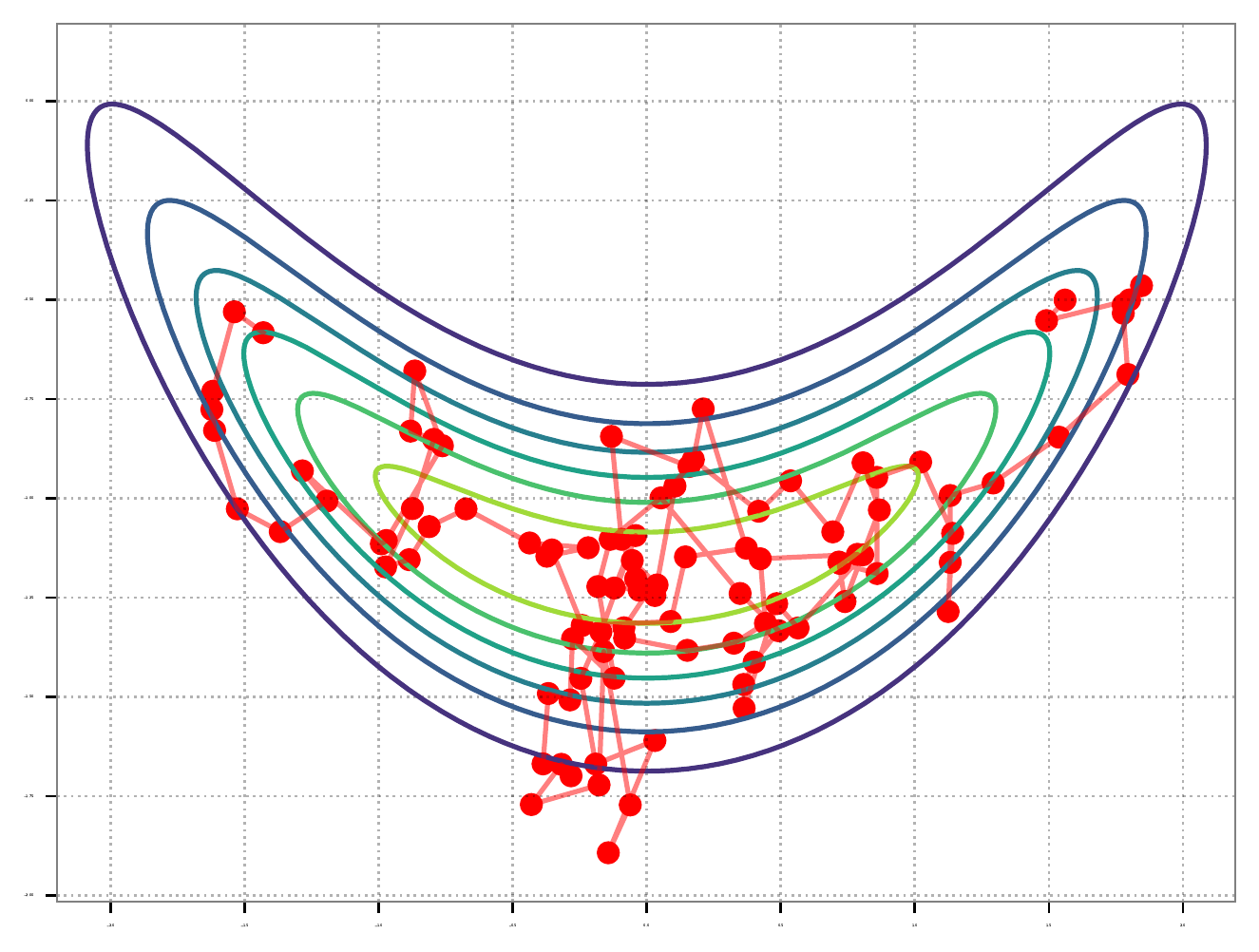}
\includegraphics[scale=0.2]{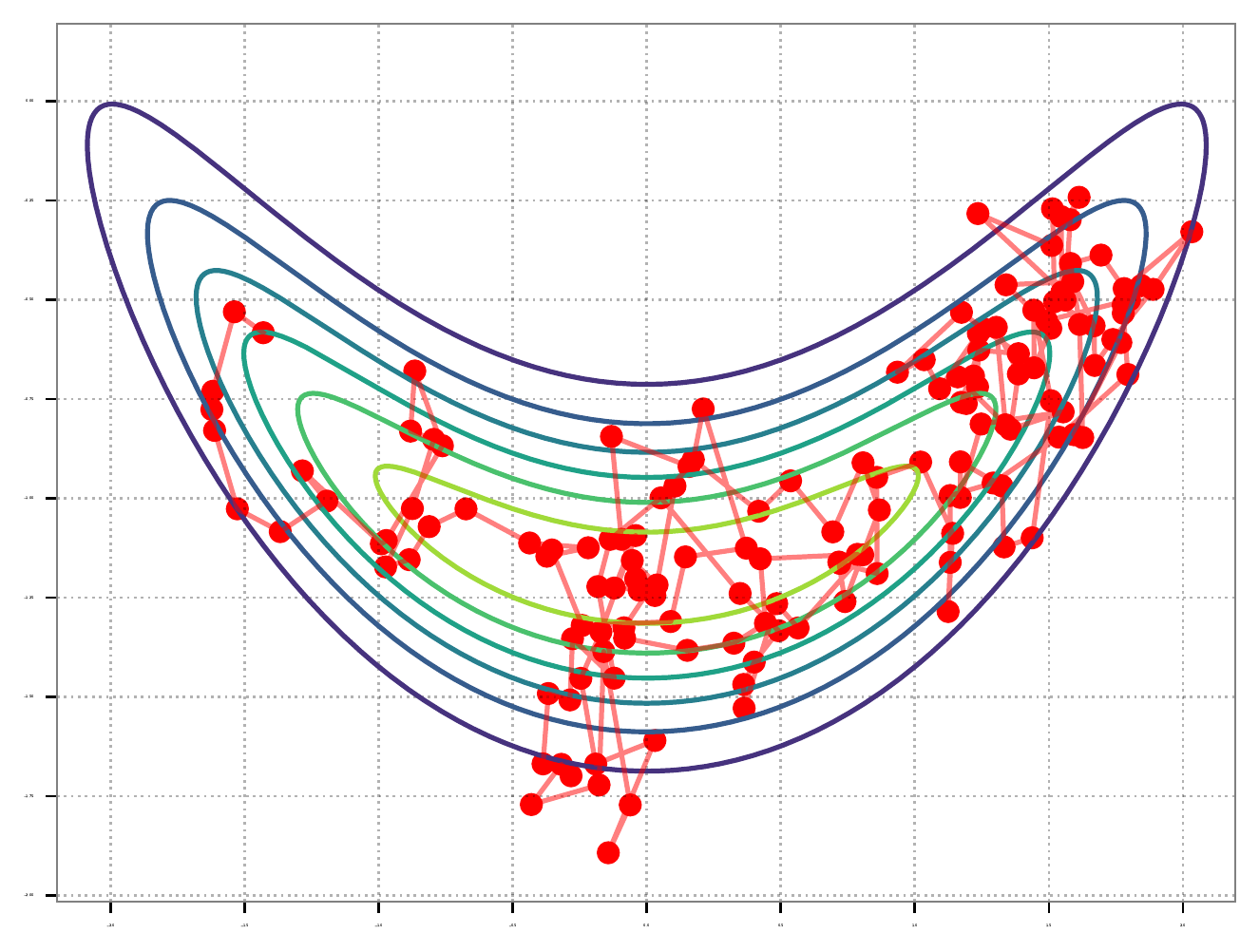}
\includegraphics[scale=0.2]{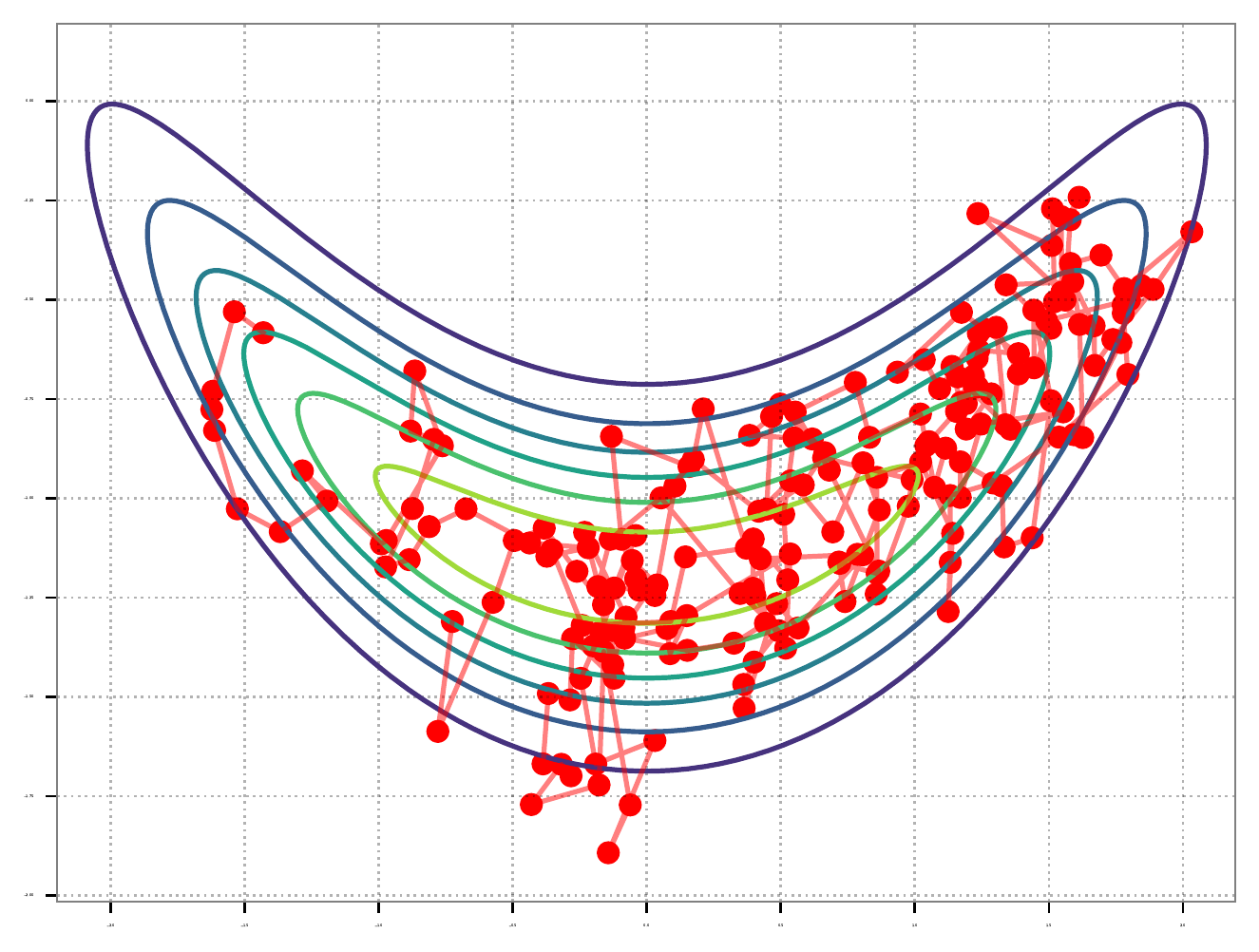}
\includegraphics[scale=0.2]{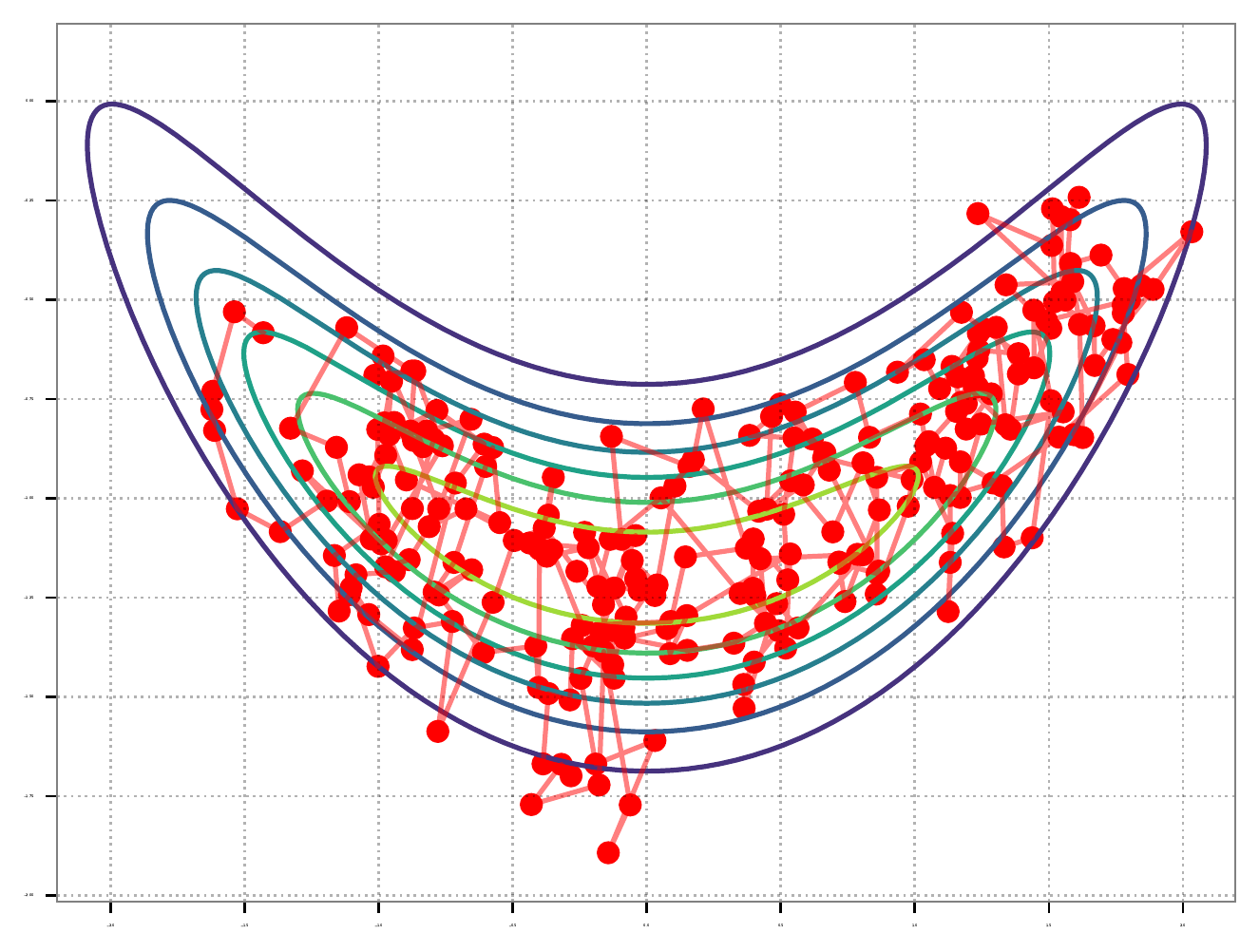}
\par\end{centering}
\begin{centering}
\includegraphics[scale=0.2]{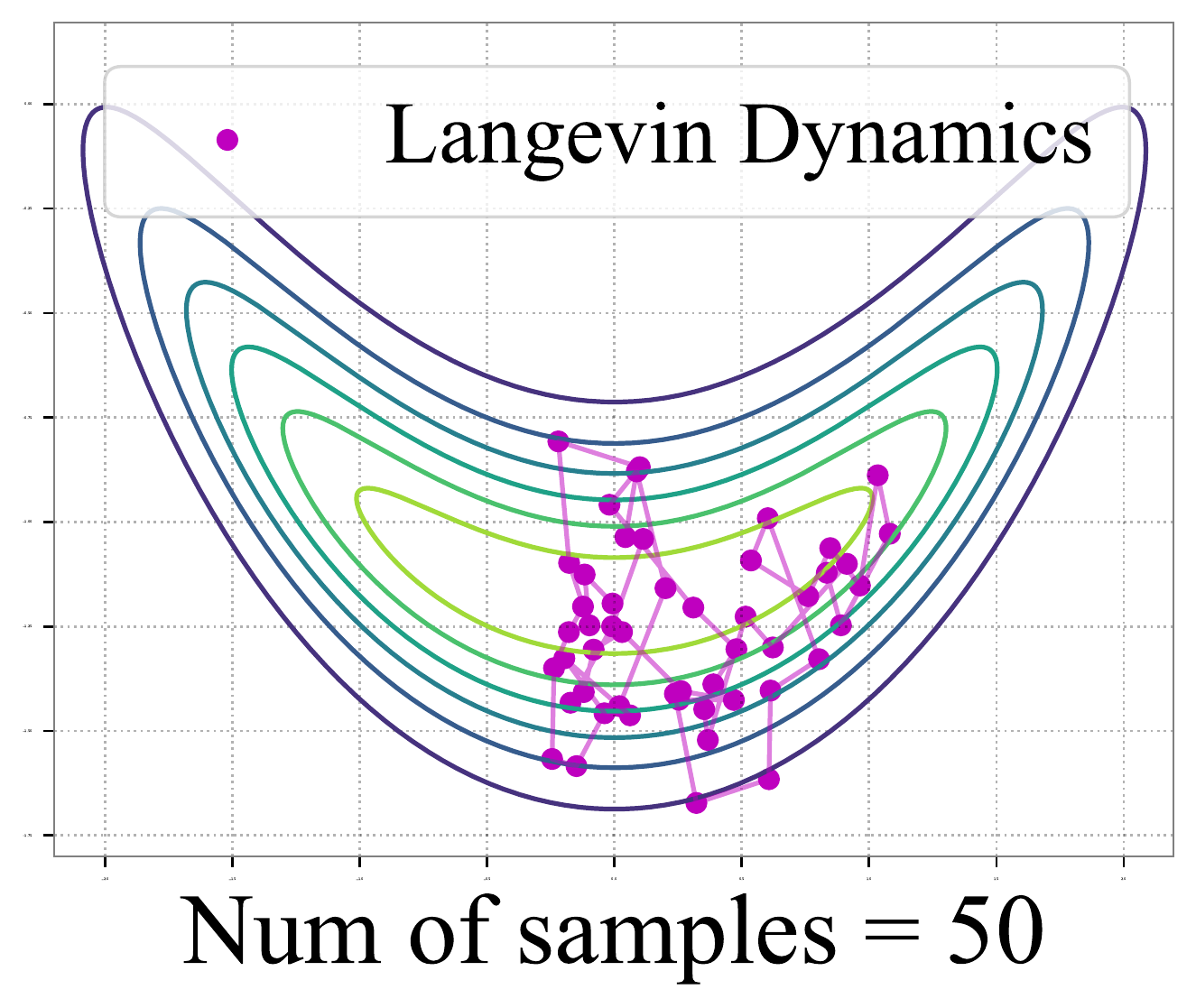}
\includegraphics[scale=0.2]{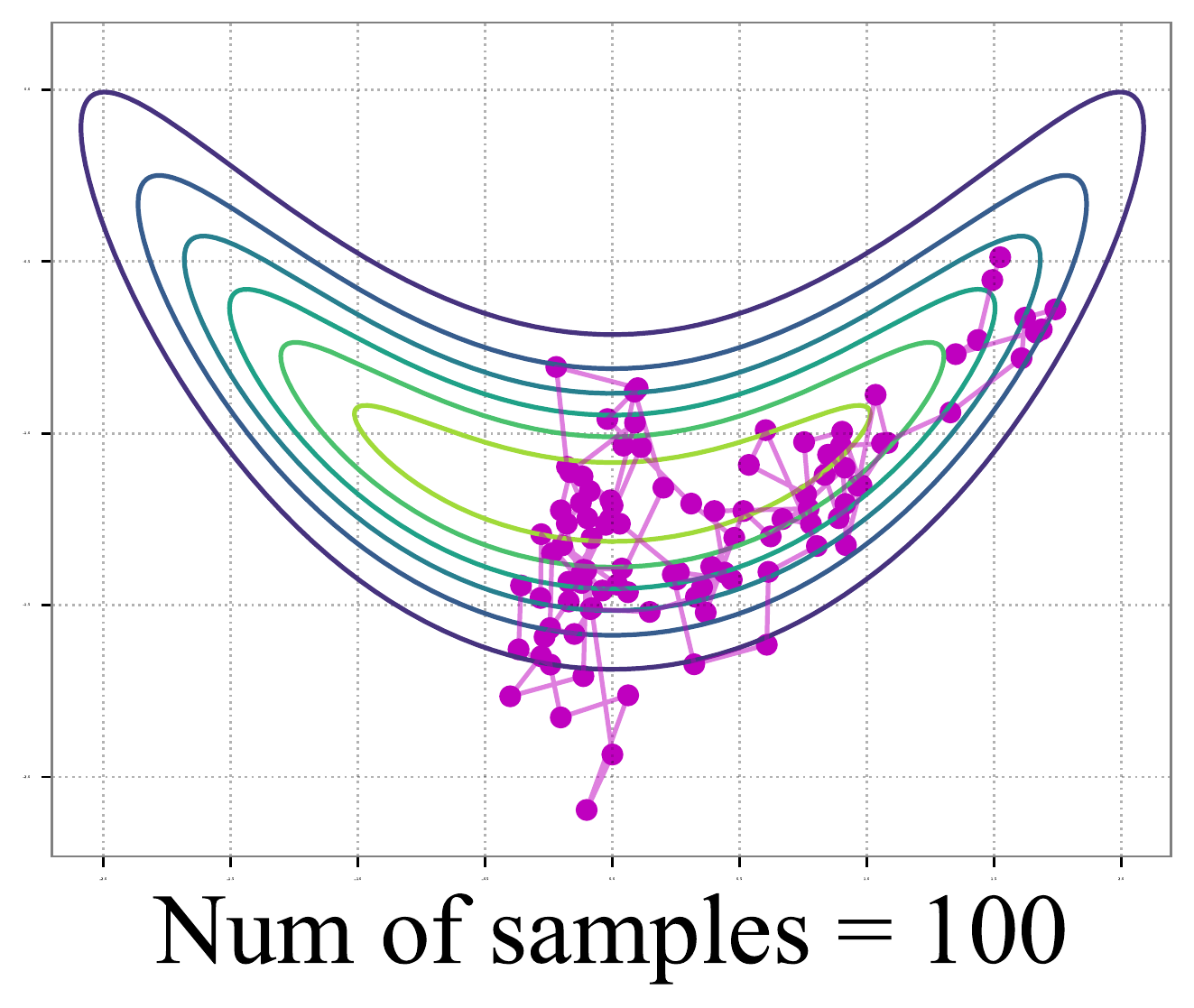}
\includegraphics[scale=0.2]{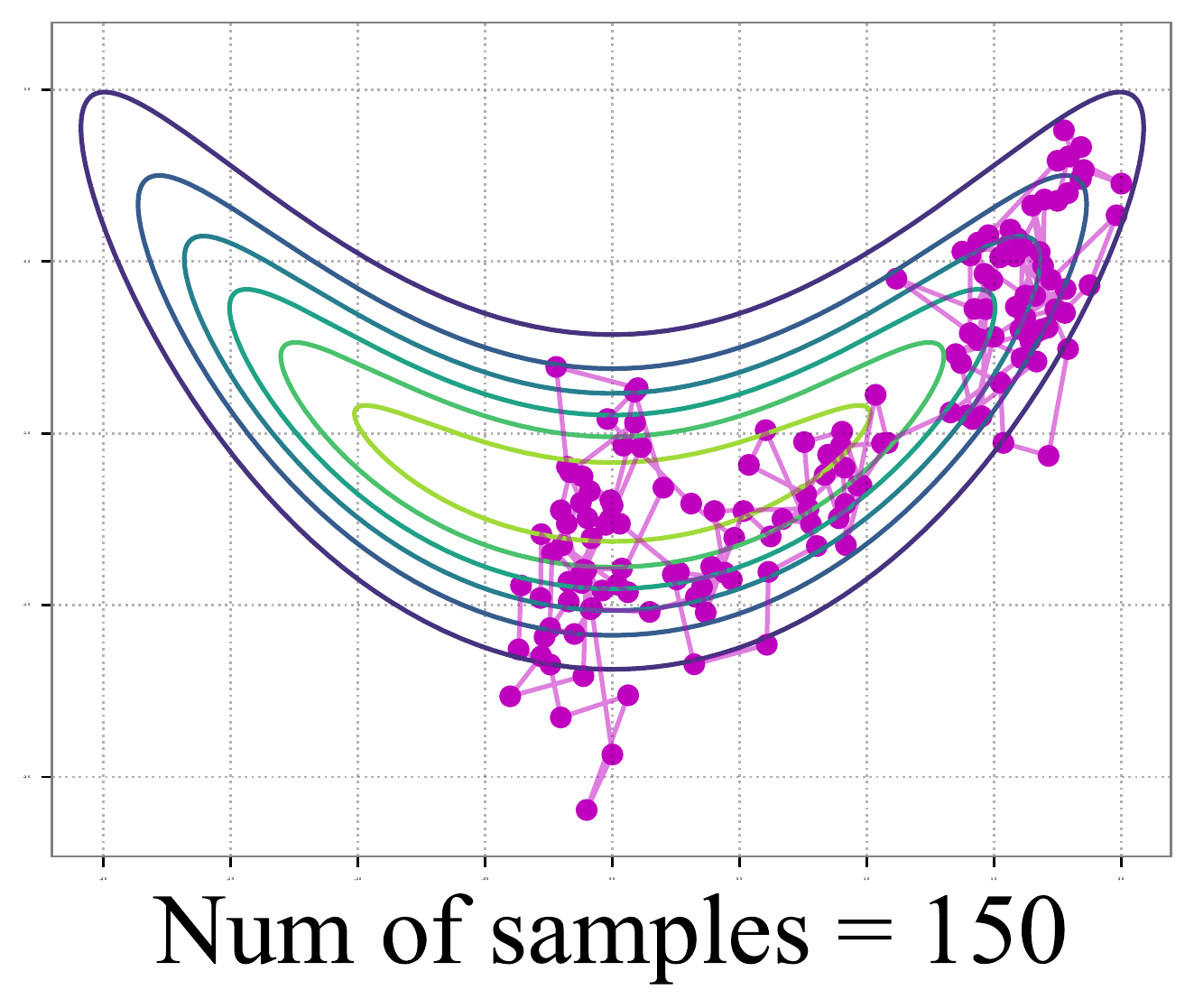}
\includegraphics[scale=0.2]{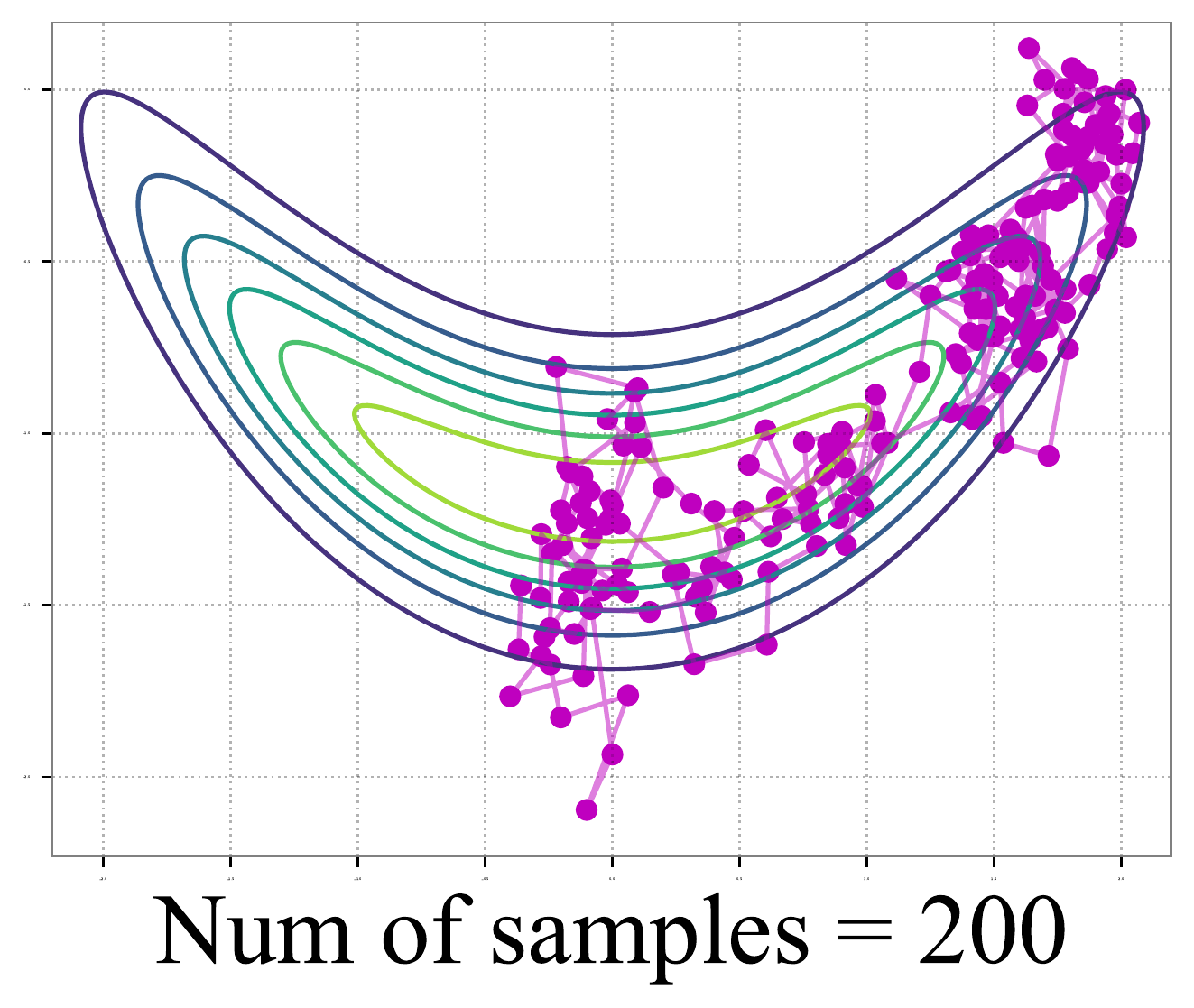}
\includegraphics[scale=0.2]{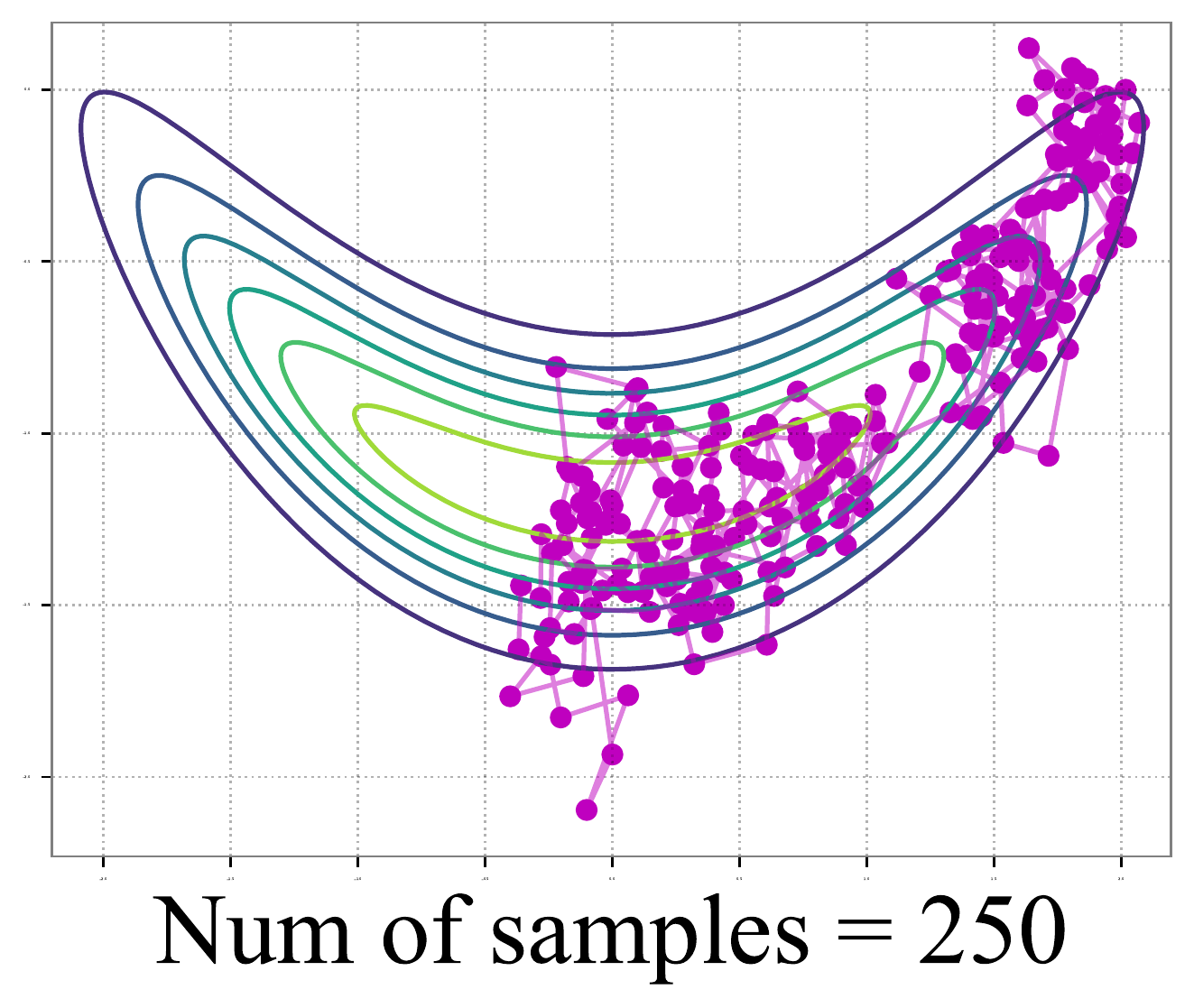}
\par\end{centering}
\caption{Sampling trajectory of the correlated 2D distribution.
\label{fig:Sampling-Trajectory-corr}}
\end{figure*}
Figure \ref{fig:Sample-quality-corr} summarizes the result with different metrics. We can see that SRLD has a significantly smaller MMD and Wasserstein-1 Distance as well as a larger ESS compared with the vanilla Langevin dynamics. Moreover, the introduced repulsive gradient creates a negative auto-correlation between samples. Figure \ref{fig:Sampling-Trajectory-corr} shows a typical trajectory of the two sampling dynamics. We can see that SRLD have a faster mixing rate than vanilla Langevin dynamics. Note that since we use the same sequence of Gaussian noise for both algorithms, the difference is mainly due to the use of repulsive gradient rather than the randomness.

\vspace{-0.5em}
\subsection{Bayesian Neural Network} \label{sec:uci}
Bayesian Neural Network is one of the most important methods in Bayesian Deep Learning with wide application in practice. 
Here we test the performance of SRLD on sampling the posterior of Bayesian Neural Network on the UCI datasets \citep{Dua:2019}. 
We assume the output is normal distributed, with a two-layer neural network with 50 hidden units and  $\mathrm{tanh}$ activation to predict the mean of outputs. 
All of the datasets are randomly partitioned into $90\%$ for training and $10\%$ for testing. 
The results are averaged over 20 random trials. We refer readers to Appendix \ref{sec:uci_appendix} for hyper-parameter tuning and other experiment details.
\begin{table*}[h]
\scriptsize
\setlength\tabcolsep{3.pt}
\centering{}%
\begin{tabular}{c|ccc|ccc}
\hline 
\multirow{2}{*}{Dataset} & \multicolumn{3}{c|}{Ave Test RMSE} & \multicolumn{3}{c}{Ave Test LL}\tabularnewline
 & \multicolumn{1}{c}{SVGD} & \multicolumn{1}{c}{LD} & SRLD & SVGD & LD & SRLD\tabularnewline
\hline 
Boston & $3.300\pm0.142$ & $3.342\pm0.187$ & $\underline{{\bf 3.086\pm0.181}}$ & $-4.276\pm0.217$ & $-2.678\pm0.092$ & $\underline{{\bf -2.500\pm0.054}}$\tabularnewline
Concrete & $4.994\pm0.171$ & $4.908\pm0.113$ & ${\bf 4.886\pm0.108}$ & $-5.500\pm0.398$ & $-3.055\pm0.035$ & $\underline{{\bf -3.034\pm0.031}}$\tabularnewline
Energy  & $0.428\pm0.016$ & $0.412\pm0.016$ & $\underline{{\bf 0.395\pm0.016}}$ & $-0.781\pm0.094$ & $-0.543\pm0.014$ & $\underline{{\bf -0.476\pm0.036}}$\tabularnewline
Naval & $0.006\pm0.000$ & $0.006\pm0.002$ & $\underline{{\bf 0.003\pm0.000}}$ & $3.056\pm0.034$ & $4.041\pm0.030$ & $\underline{{\bf 4.186\pm0.015}}$\tabularnewline
WineRed & $0.655\pm0.008$ & $0.649\pm0.009$ & $\underline{{\bf 0.639\pm0.009}}$ & $-1.040\pm0.018$ & $-1.004\pm0.019$ & $\underline{{\bf -0.970\pm0.016}}$\tabularnewline
WineWhite & $\underline{{\bf 0.655\pm0.008}}$ & $0.692\pm0.003$ & $0.688\pm0.003$ & $\underline{{\bf -1.040\pm0.019}}$ & $-1.047\pm0.004$ & $-1.043\pm0.004$\tabularnewline
Yacht & $0.593\pm0.071$ & $0.597\pm0.051$ & ${\bf 0.578\pm0.054}$ & $-1.281\pm0.279$ & $-1.187\pm0.307$ & $\underline{{\bf -0.458\pm0.036}}$\tabularnewline
\hline 
\end{tabular}
\caption{Averaged test RMSE and test log-likelihood on UCI datasets. Results are averaged over 20 trails. The boldface indicates the method has the best average performance and the underline marks the methods that perform the best with a significance level of $0.05$.}
\label{tb: UCI_main}
\end{table*}
Table \ref{tb: UCI_main} shows the average test RMSE and test log-likelihood and their standard deviation. The method that has the best average performance is marked as boldface. 
We observe that a large portion of the variance is due to the random partition of the dataset. 
Therefore, to show the statistical significance, 
we use the matched pair $t$-test to test the statistical significance, mark the methods that perform the best with a significance level of 0.05 with underlines. 
%
Note that the results of SRLD/LD and SVGD is not very comparable, because   
SRLD/LD are single chain methods which averages across time, and SVGD is a multi-chain method that only use the results of the last iteration.  
We provide additional results in Appendix \ref{sec:uci_appendix} that SRLD averaged on 20 particles (across time) can also 
achieve similar or better results as SVGD with 20 (parallel) particles. 

\vspace{-0.5em}
\subsection{Contextual Bandits}
We consider the posterior sampling (a.k.a Thompson sampling) algorithm with Bayesian neural network as the function approximator, to demonstrate the uncertanty estimation provided by SRLD. We follow the experimental setting from \citet{DBLP:conf/iclr/RiquelmeTS18}. The only difference is that we change the optimization of the objective (e.g. evidence lower bound (ELBO) in variational inference methods) into running MCMC samplers. We compare the SRLD with the Langevin dynamics on two benchmarks from \citep{DBLP:conf/iclr/RiquelmeTS18}, and include SVGD as a baseline. 
For more detailed introduction, setup, hyper-parameter tuning and experiment details; see Appendix \ref{sec:contexual_bandits}. 

\begin{table}
    \centering
    \begin{tabular}{c|c|c|c}
    \hline
        Dataset & SVGD & LD & SRLD \\
        \hline
        Mushroom &  $20.7\pm 2.0$ & $4.28 \pm 0.09 $ & $\underline{\mathbf{3.80\pm 0.16}}$\\
        Wheel & $91.32 \pm 0.17$ & $38.07 \pm 1.11 $ & $\underline{\mathbf{32.08 \pm 0.75}}$\\
    \hline
    \end{tabular}
    \vspace{0.3cm}
    \caption{Cumulative Regrets on two bandits problem (smaller is better). Results are averaged over 10 trails. Boldface indicates the methods with best performance and underline marks the best significant methods with significant level $0.05$.}
    \label{tab:bandits}
\end{table}

The cumulative regret is shown in Table \ref{tab:bandits}. SVGD is known to have the under-estimated uncertainty for Bayesian neural network if particle number is limited \citep{wang2019function}, and as a result, has the worst performance among the three methods. SRLD is slightly better than vanilla Langevin dynamics on the simple Mushroom bandits. On the much more harder Wheel bandits, SRLD is significantly better than the vanilla Langevin dynamics, which shows the improving uncertainty estimation of our methods within finite number of samples.

\section{Conclusion}
We propose a Stein self-repulsive dynamics which applies Stein variational gradient to push samples from MCMC dynamics away from its past trajectories. This allows us to significantly decrease the auto-correlation of MCMC, increasing the sample efficiency for better estimation. The advantages of our method are extensive  studied both theoretical and empirical analysis in our work. In future work, we plan to investigate the combination of our Stein self-repulsive idea with more general MCMC procedures, and explore broader applications.

\clearpage

\paragraph{Broader Impact Statement}
This work incorporates Stein repulsive force into Langevin dynamics to improve sample efficiency. It brings a positively improvement to the community such as reinforcement learning and Bayesian neural network that needs efficient sampler. Our work do not have any negative societal impacts that we can foresee in the future.

\paragraph{Acknowledgement}
This paper is supported in part by NSF CAREER 1846421, SenSE 2037267 and EAGER 2041327.

\bibliography{main}
\bibliographystyle{plainnat}

\newpage
\appendix



\section{Discussion on Hyper-parameter Tuning}
The key hyper-parameters of SRLD are 1. $\alpha$, which balance the
confining gradient and repulsive gradient; 2. $M$ the number of particles
used; 3. $\sigma$ the bandwidth of kernel; 4. $\eta$ the stepsize;
5. $c_{\eta}$ the thinning factor. Among which, $\alpha$, $M$ and
$\sigma$ are the hyper-parameter introduced by the proposed repulsive
gradient and thus we mainly discuss these three hyper-parameter. The
number of particles $M$ and bandwidth of kennel $\sigma$ are introduced
by the use the repulsive term in SVGD \citep{liu2016stein}. In practice, we find
using a similar setting for tuning $M$ and $\sigma$ as that in SVGD
\citep{liu2016stein} gives good performance. In specific, in order to obtain
good performance, $M$ does not needs be very large, and similar to
SVGD, $M=10$ already gives good enough particle approximation. A
good choice of bandwidth $\sigma$ is also important for the kernel.
In SVGD, instead of tuning $\sigma$, they propose an adaptive way
to adjust $\sigma$ during the running of the dynamics. Specifically,
they choose $\sigma=\text{med}^{2}/\log(M)$, where $\text{med}$
is the median of the pairwise distance between the particles $\th_{k}^{i}$,
$i\in[M]$. In this way, the bandwidth ensures that $\sum_{j=1}^{M}K(\th_{k}^{i},\th_{k}^{j})\approx1$.
This adaptive way of choosing $\sigma$ is also widely used in current
approximation inference area, e.g., \citet{liu2017policy,han2017stein,wang2019function,wang2019stein}. We also find that applying this adaptive bandwidth is able to give good empirical performance and thus we also use this method in the implementation. Now we discuss how choose $\alpha$. Notice that $\alpha$ serves to balance the confining gradient and repulsive gradient and based on this motivation, we recommend readers to find a proper $\alpha$ using the samples at burn-in phase by setting 
\[
\alpha\approx\frac{\sum_{k=1}^{Mc_{\eta}}\left\Vert \nabla V(\th_{k})\right\Vert }{\sum_{k=1}^{Mc_{\eta}}\left\Vert g(\th_{k},\tilde{\delta}_{k}^{M})\right\Vert }.
\]
In this way, $\alpha$ balances the two kind of gradients. And then we may further tune $\alpha$ by searching around this value. An empirical good choice of $\alpha$ is 10 for the data sets we tested and we use $\alpha=10$ for all the experiments.

The step size is important for gradient based MCMC, as too large step size gives too large discretization error while a too small step size will cause the dynamics converges very slowly. In this paper, we mainly use validation set to tune the step size. The thinning factor is also a common parameter is MCMC methods and usually MCMC methods are not sensitive to this parameter. SRLD is not sensitive to this parameter and we simply set $c_{\eta}=100$ for all experiments.

\section{Additional Experiment Result on Synthetic Data}
\label{sec:addexp}
In this section, we show additional experiment on synthetic data. To further visualize the role of the proposed stein repulsive gradient, we also apply our method to sample a 2D mixture of Gaussian distribution (see section \ref{suppsec: 2dmix}). To further study how different $\alpha$ influences sampling high dimension distribution, we apply SRLD to sample high dimensional Gaussian (section \ref{suppsec: hdgaussian}) and high dimensional mixture of Gaussian (section \ref{sec:highdimgaussian}).

\subsection{Synthetic 2D Mixture of Gaussian Experiment} \label{suppsec: 2dmix}

\begin{figure}[htb]
\begin{centering}
\includegraphics[scale=0.25]{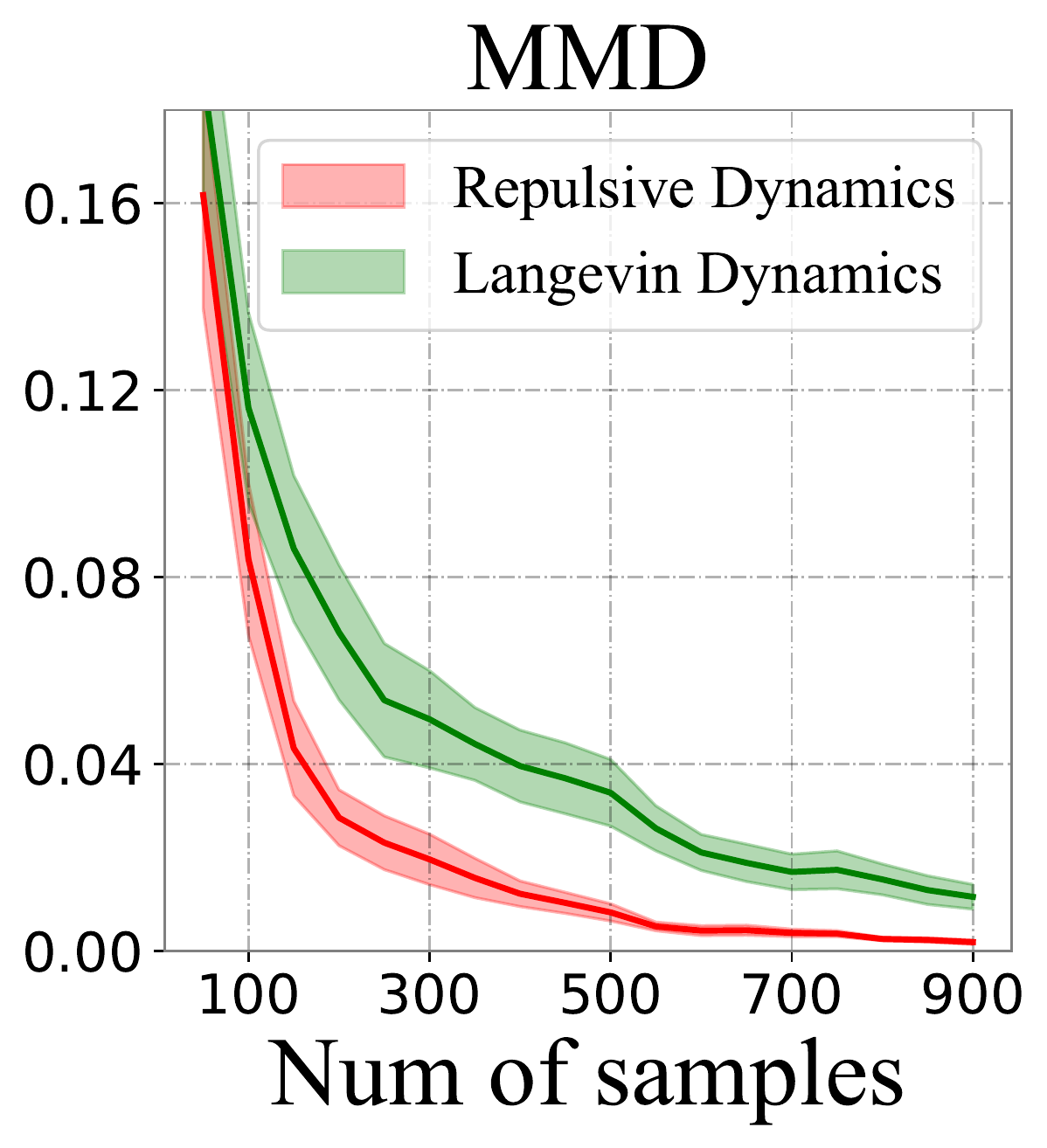}
\includegraphics[scale=0.25]{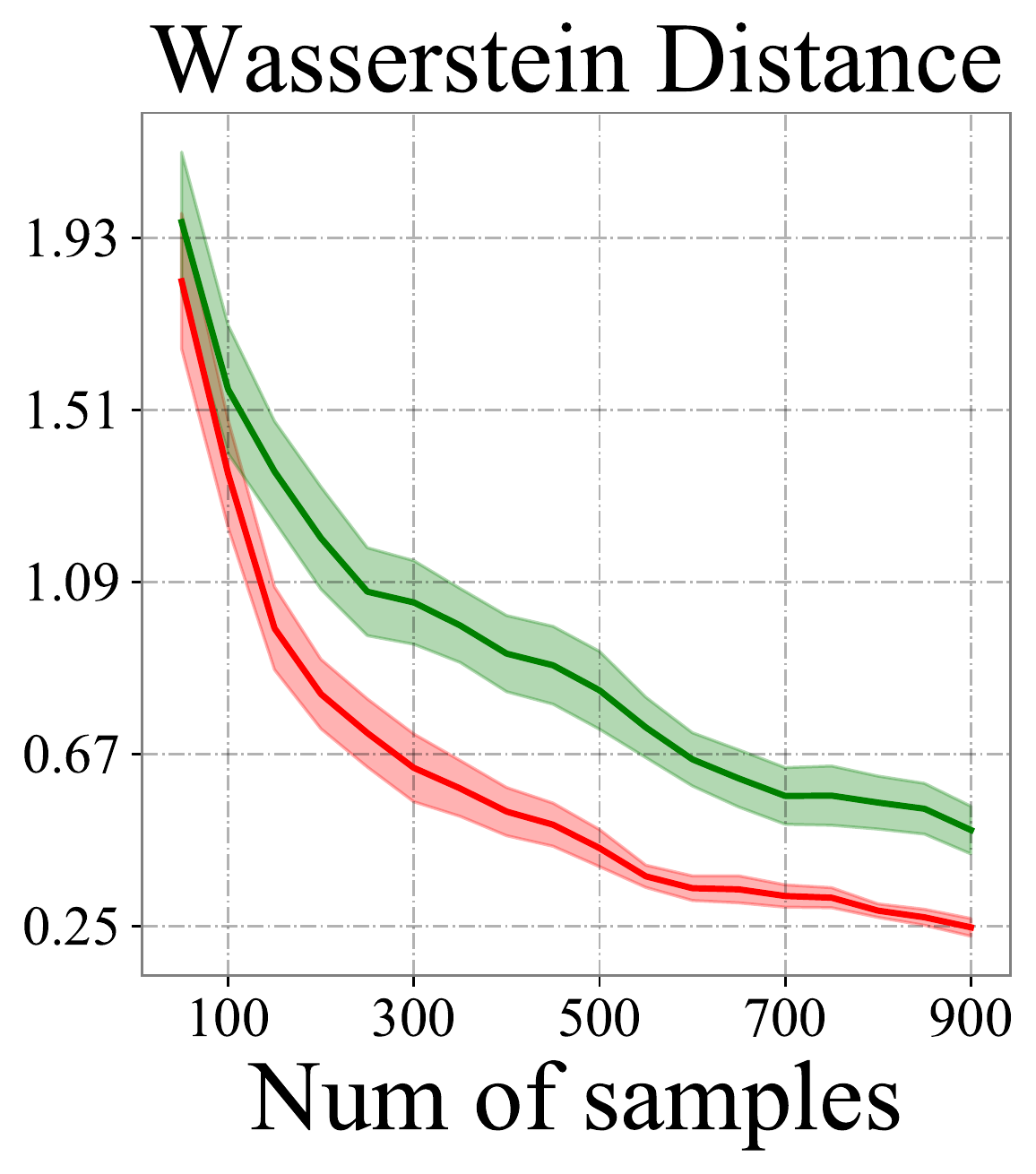}
\includegraphics[scale=0.25]{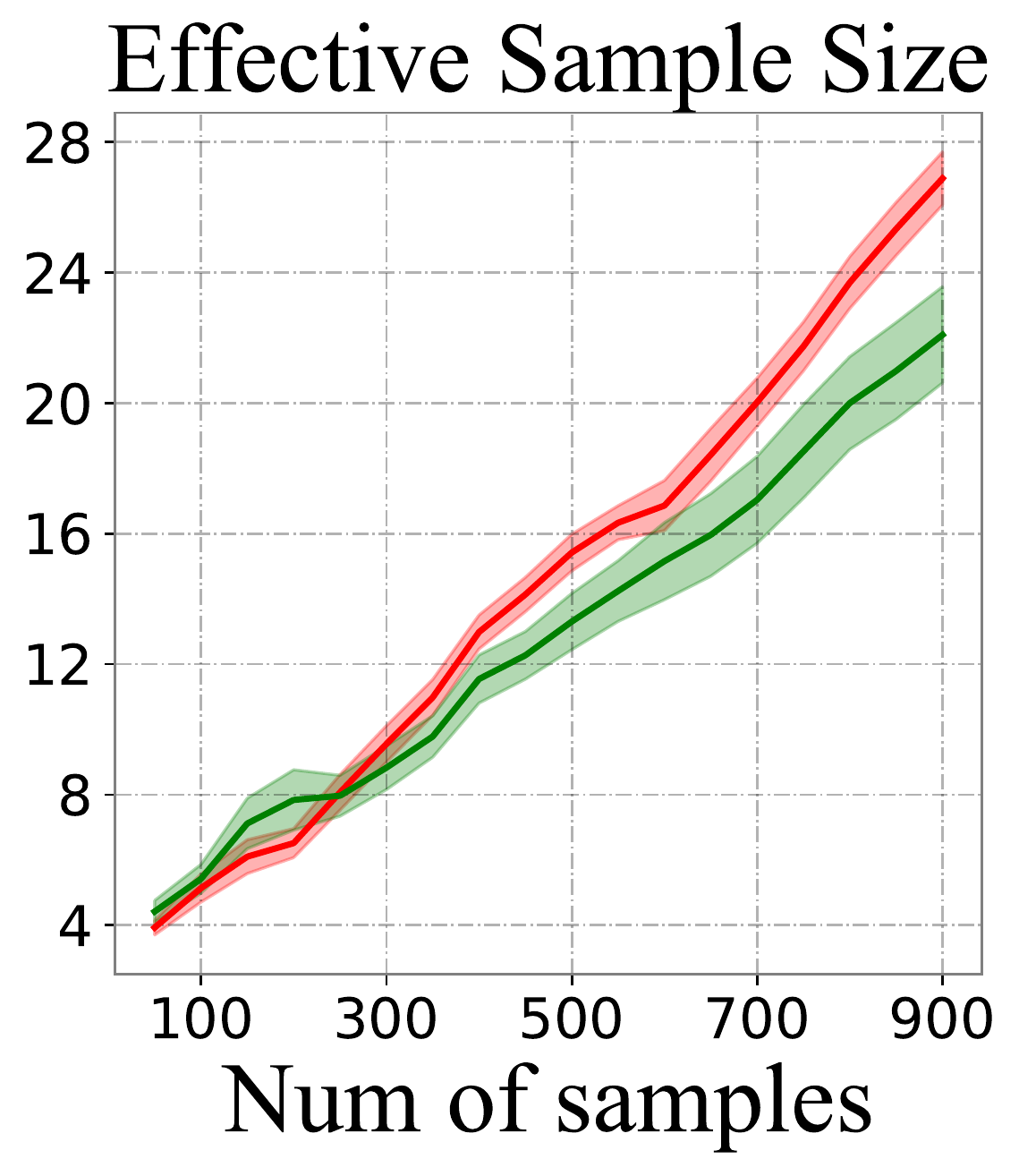}
\includegraphics[scale=0.25]{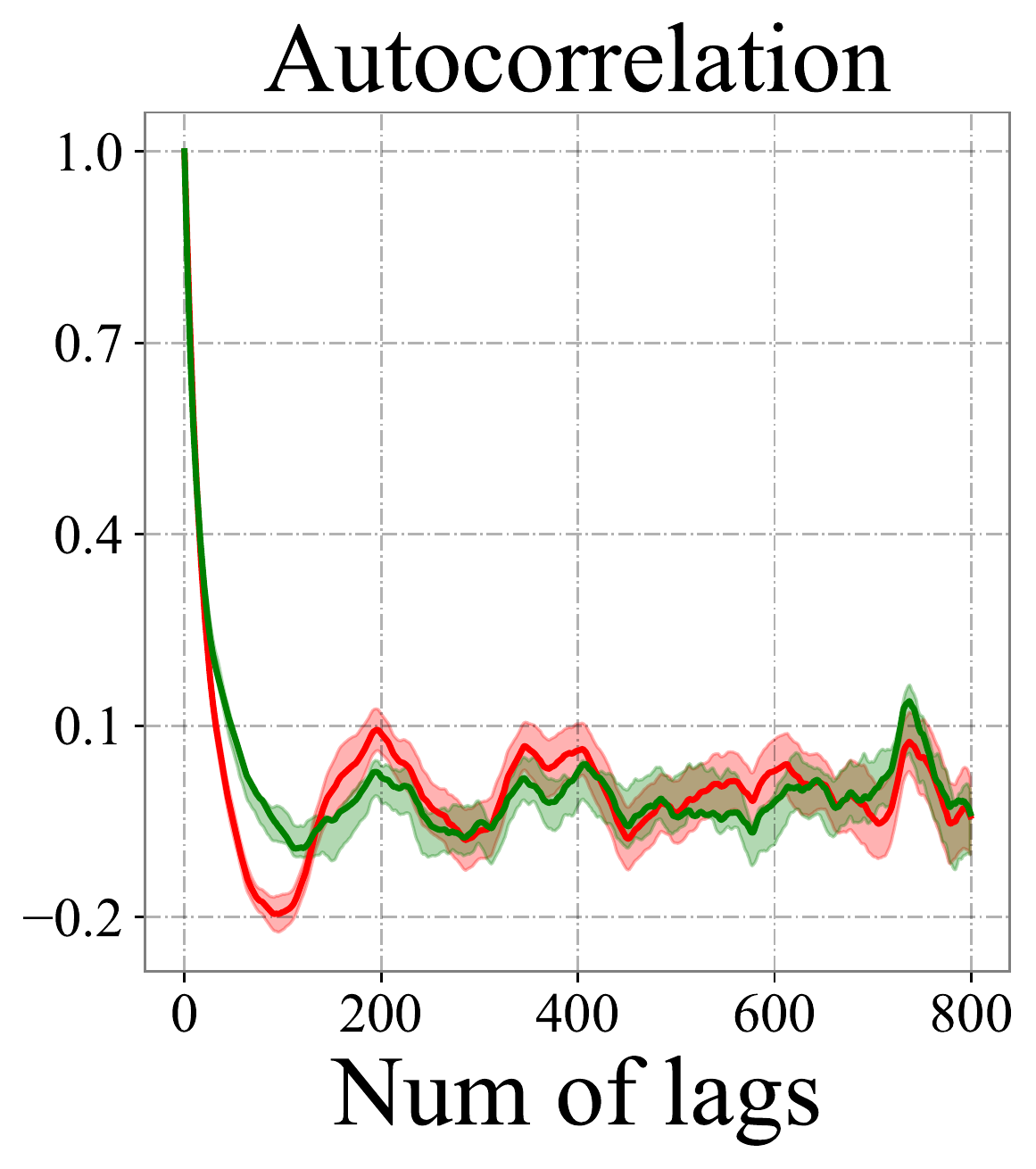}
\par\end{centering}
\caption{Sample quality and autocorrelation of the mixture distribution. The auto-correlation is the averaged auto-correlation of the two dimensions. \label{fig:Sample-quality-mix} }
\end{figure}
\begin{figure}[htb]
\begin{centering}
\includegraphics[scale=0.2]{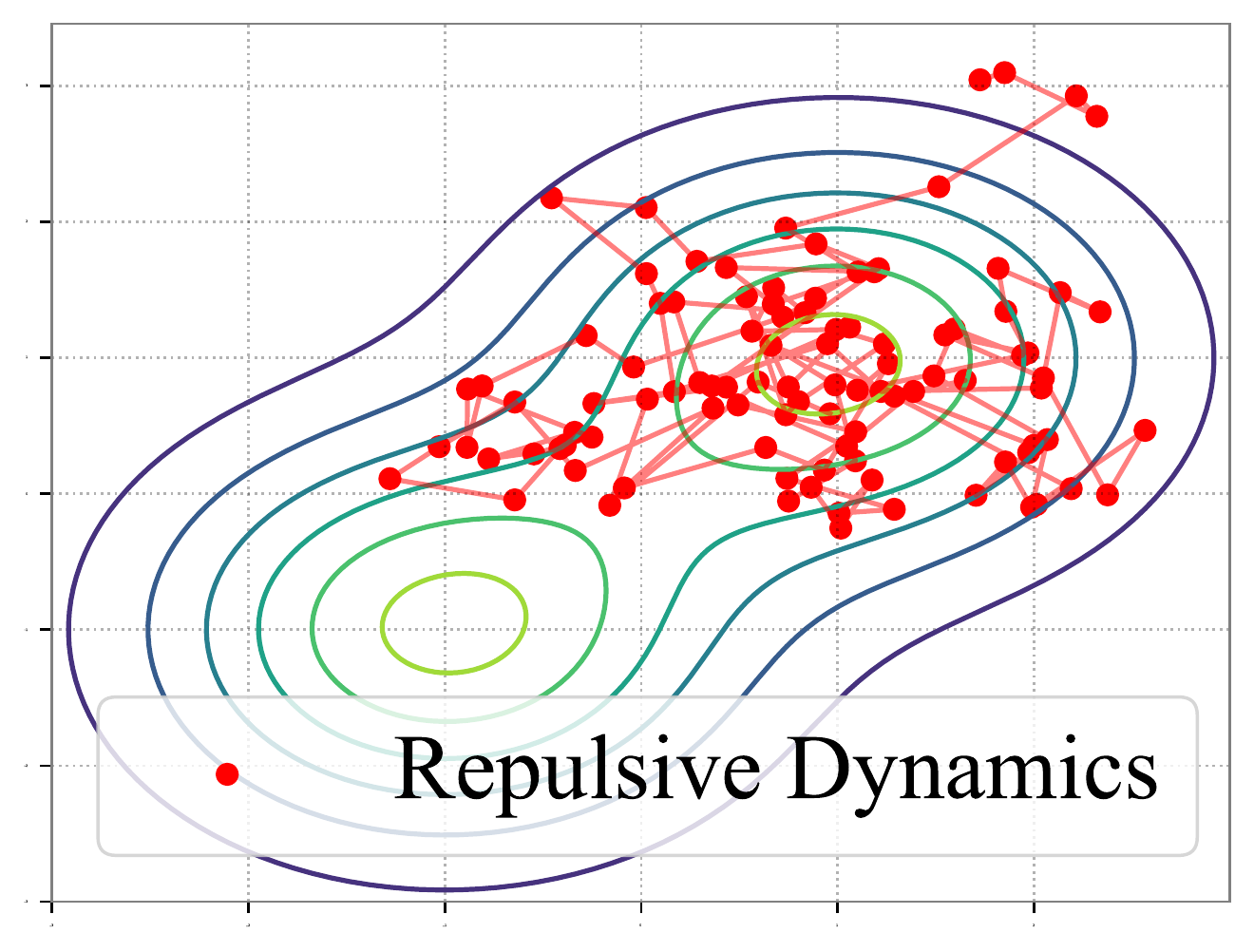}
\includegraphics[scale=0.2]{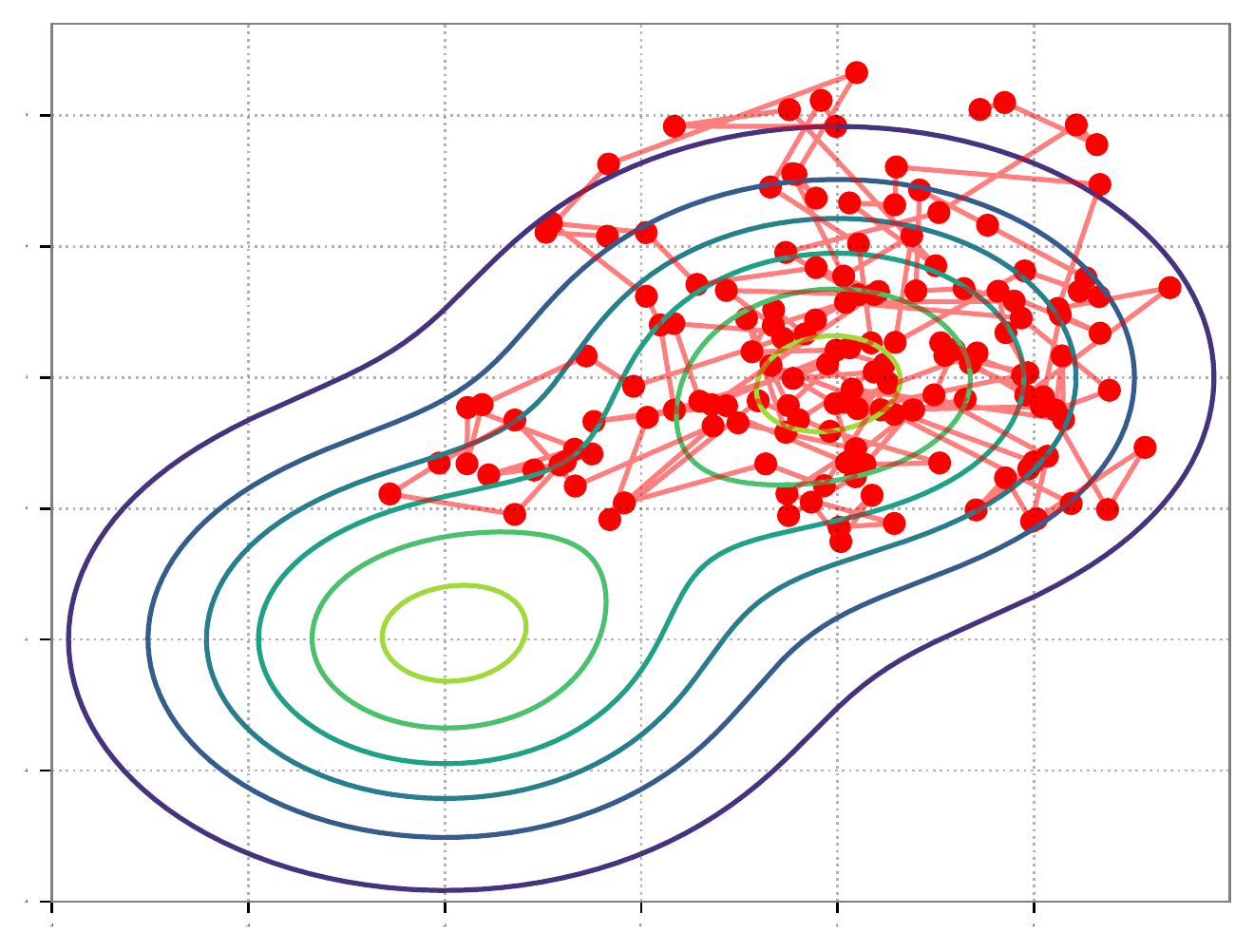}
\includegraphics[scale=0.2]{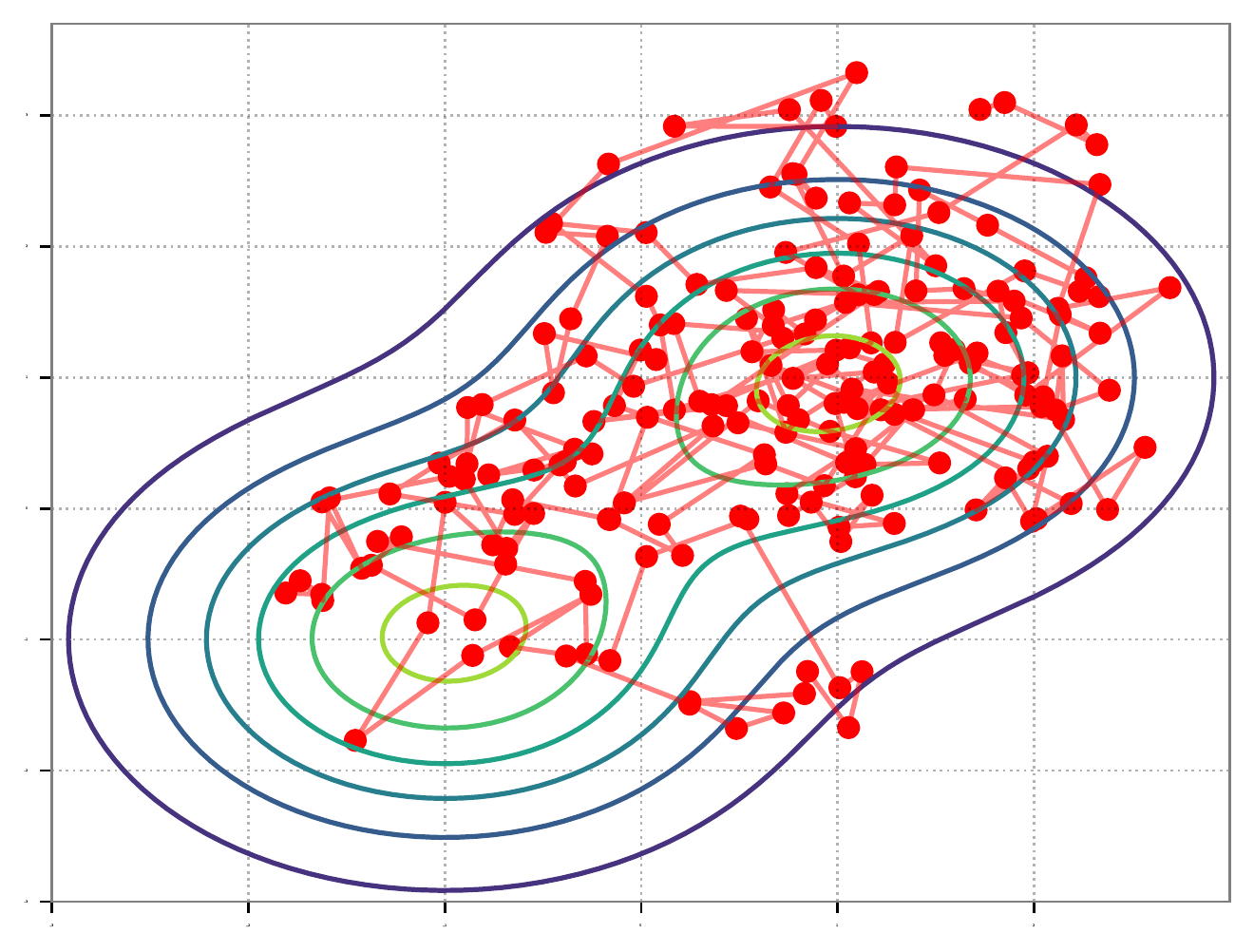}
\includegraphics[scale=0.2]{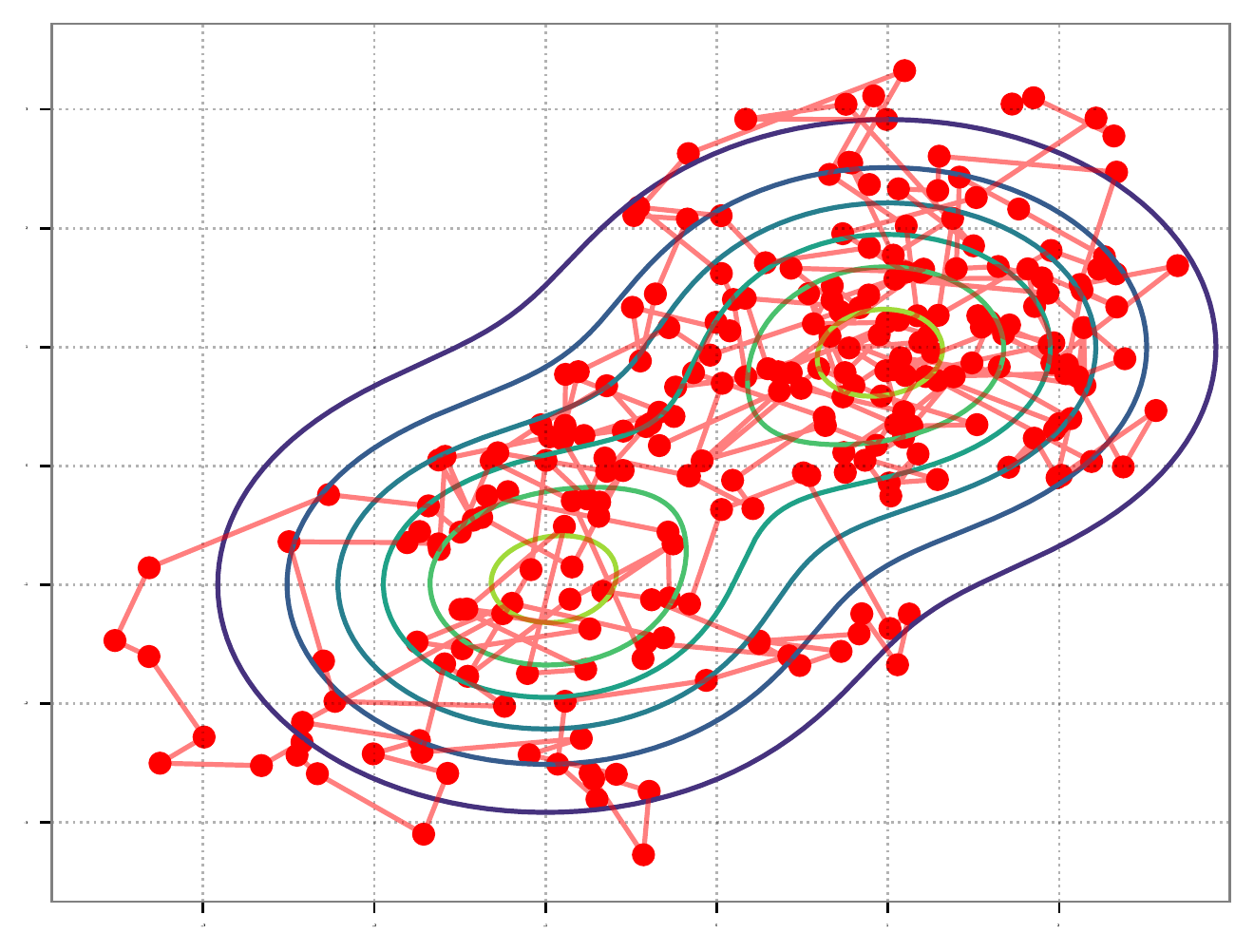}
\includegraphics[scale=0.2]{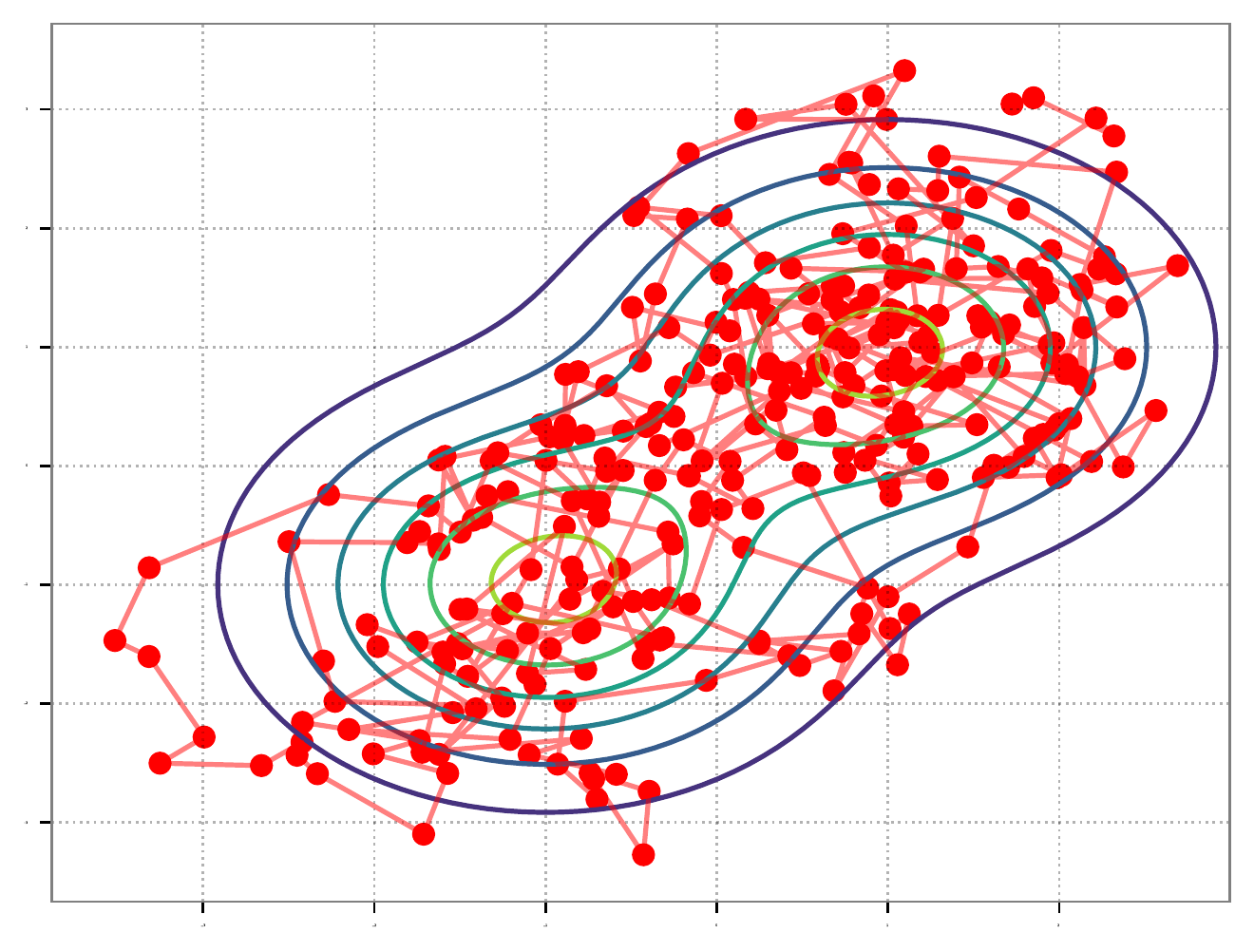}
\par\end{centering}
\begin{centering}
\includegraphics[scale=0.2]{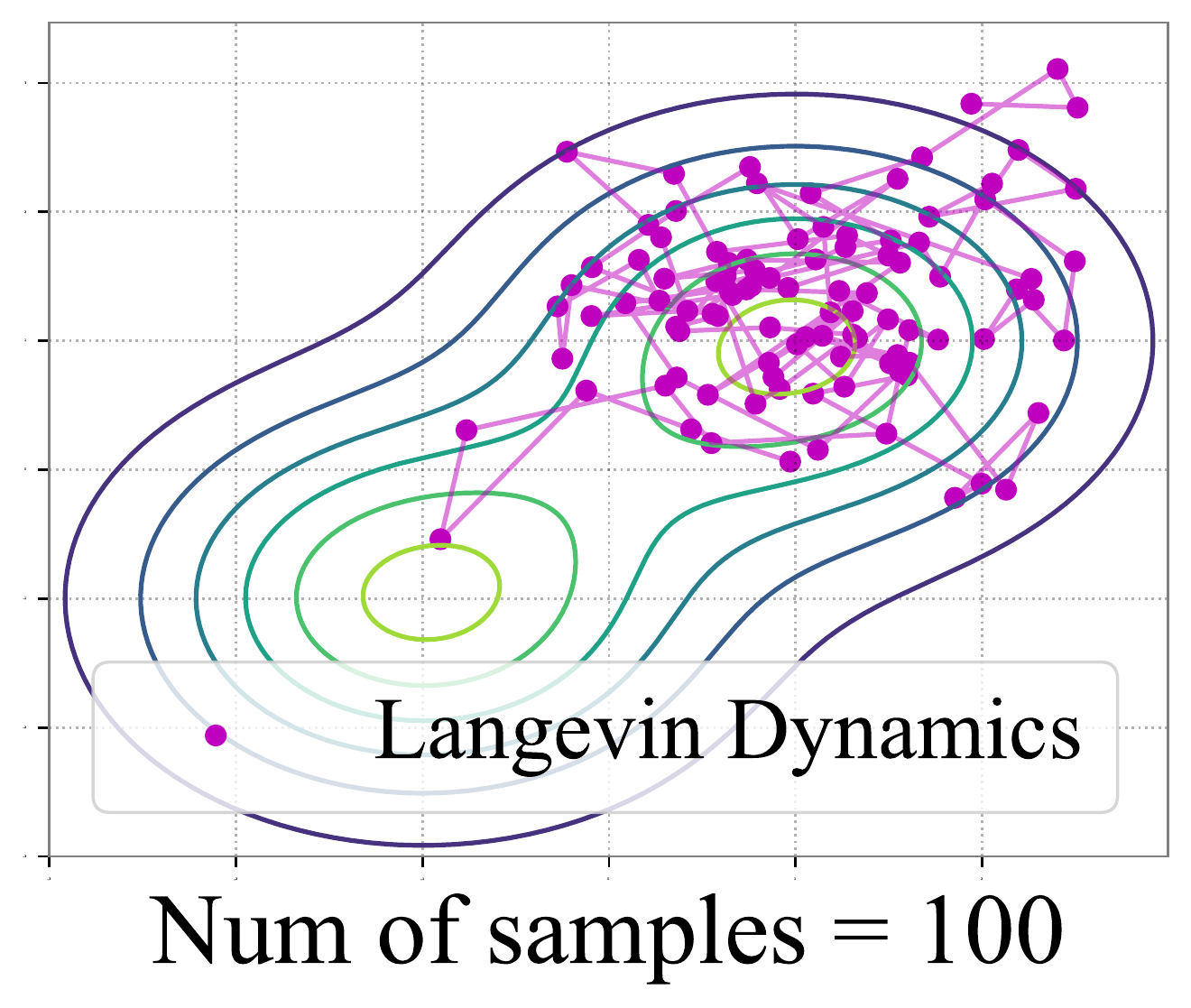}
\includegraphics[scale=0.2]{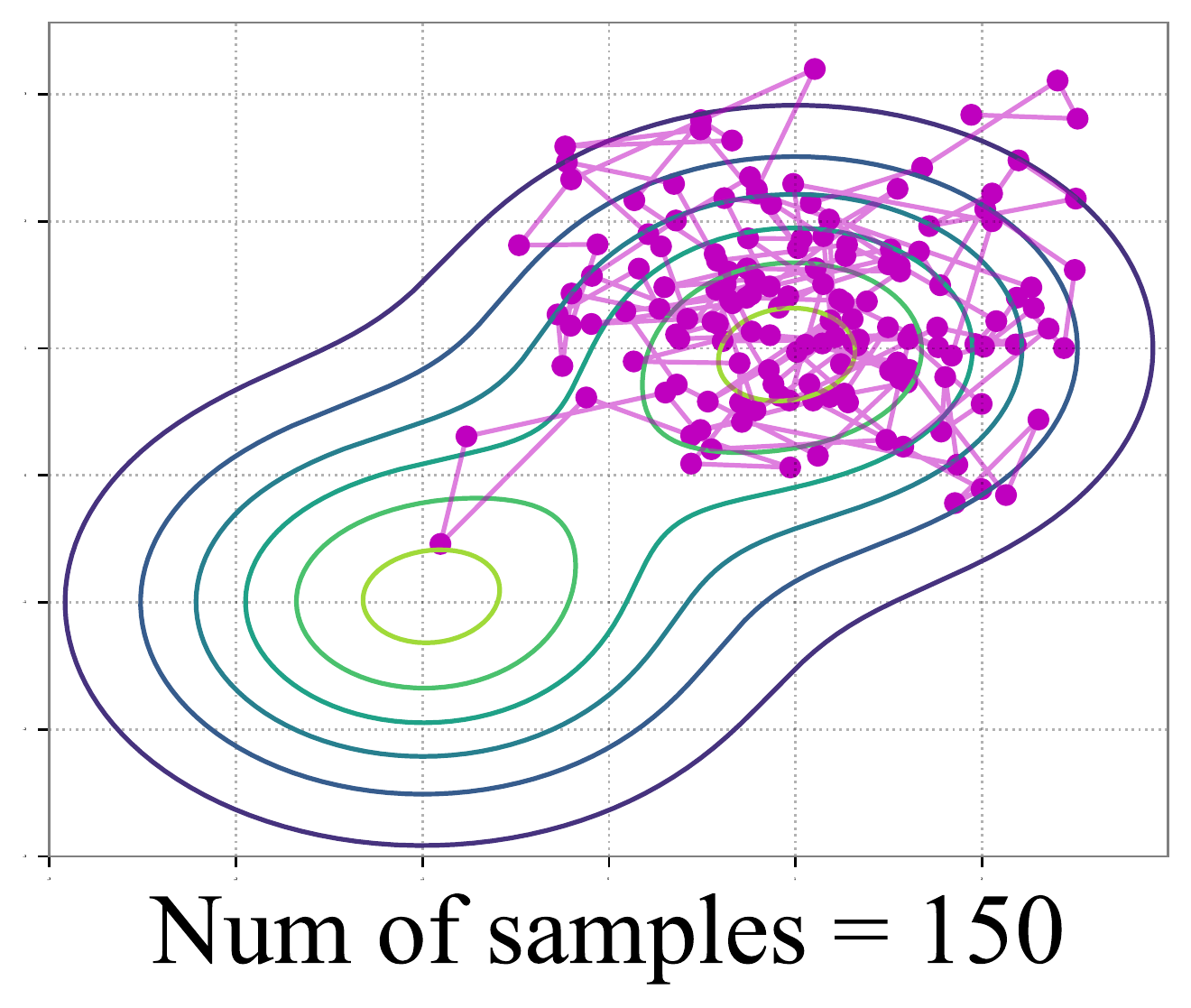}
\includegraphics[scale=0.2]{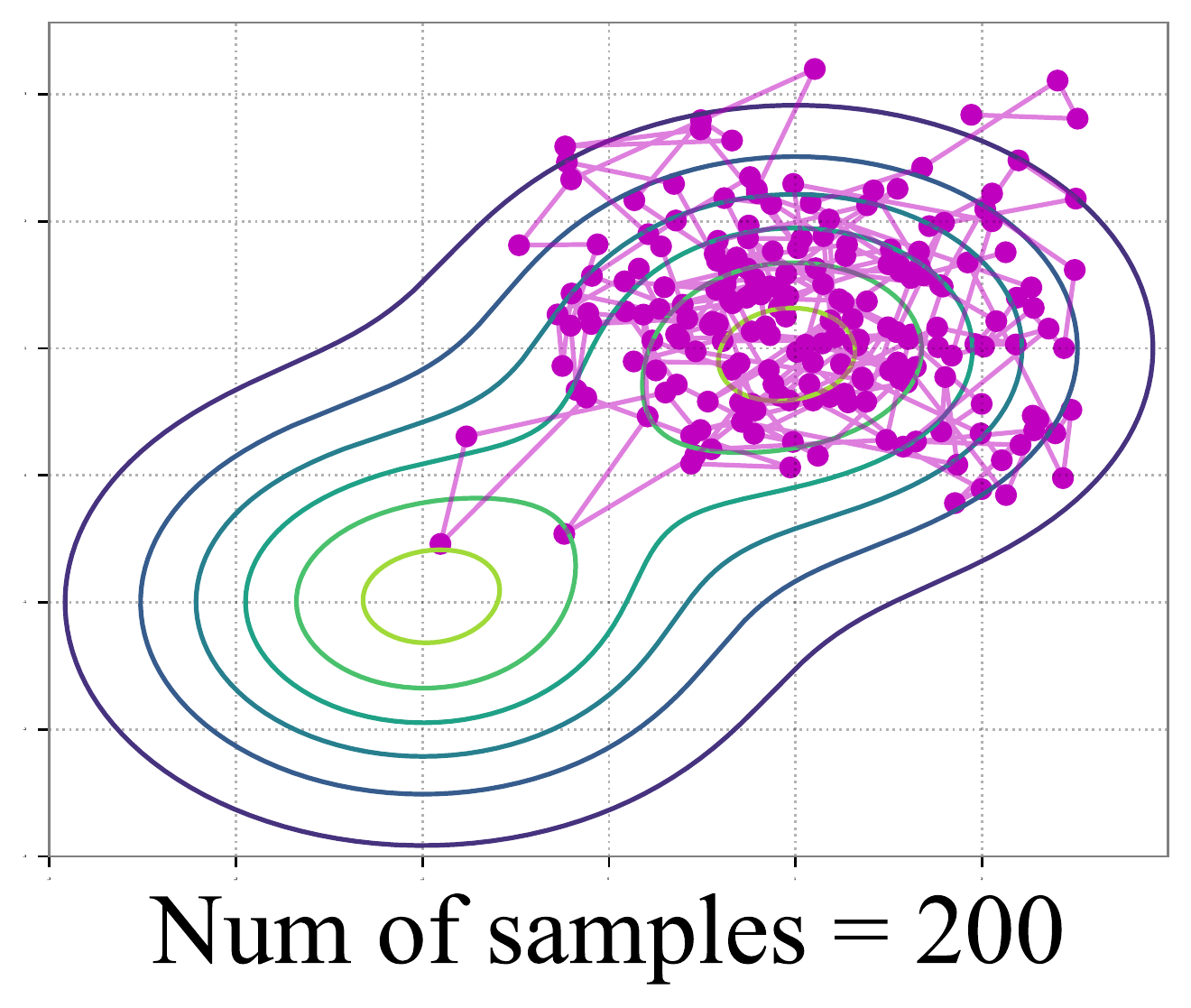}
\includegraphics[scale=0.2]{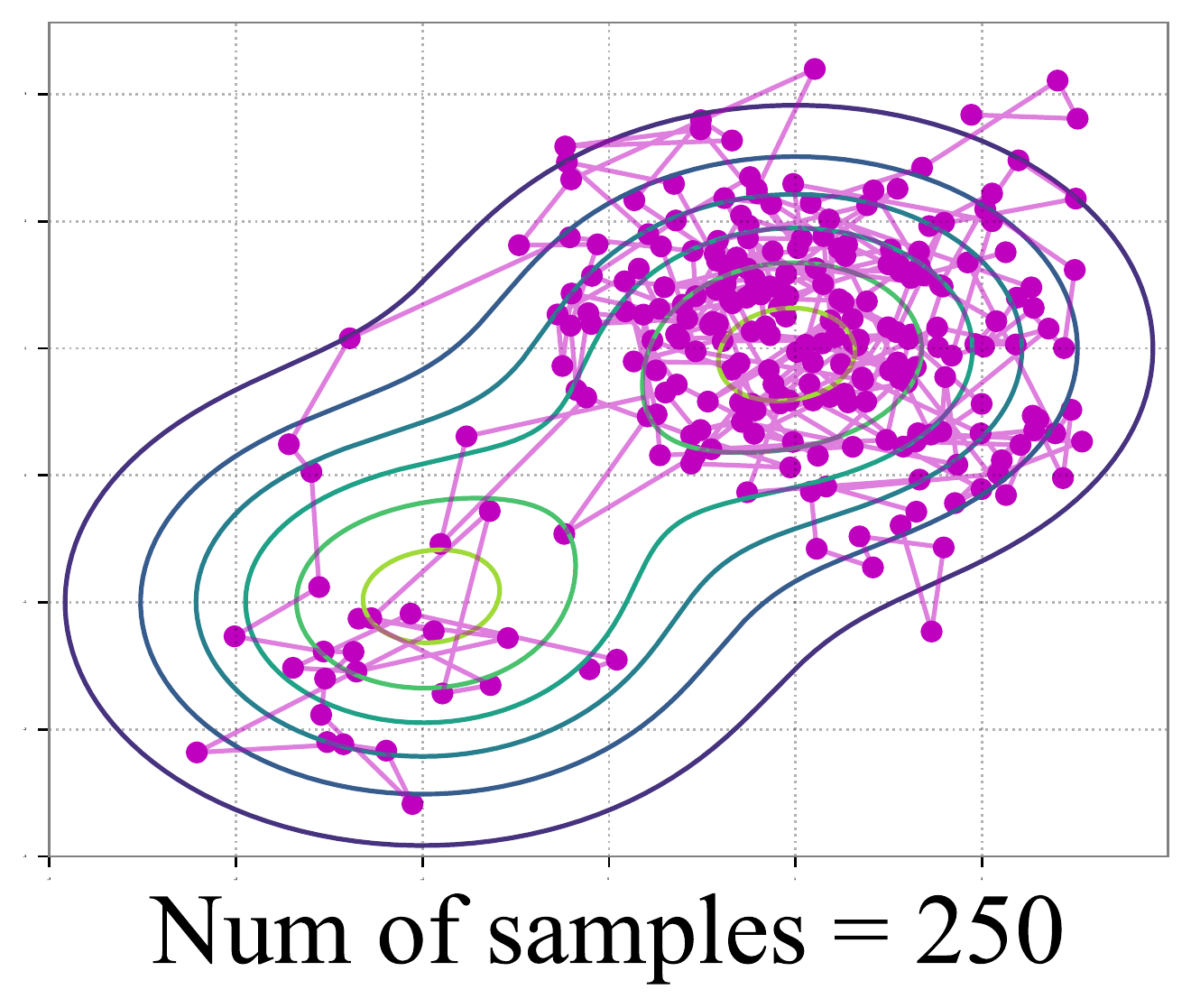}
\includegraphics[scale=0.2]{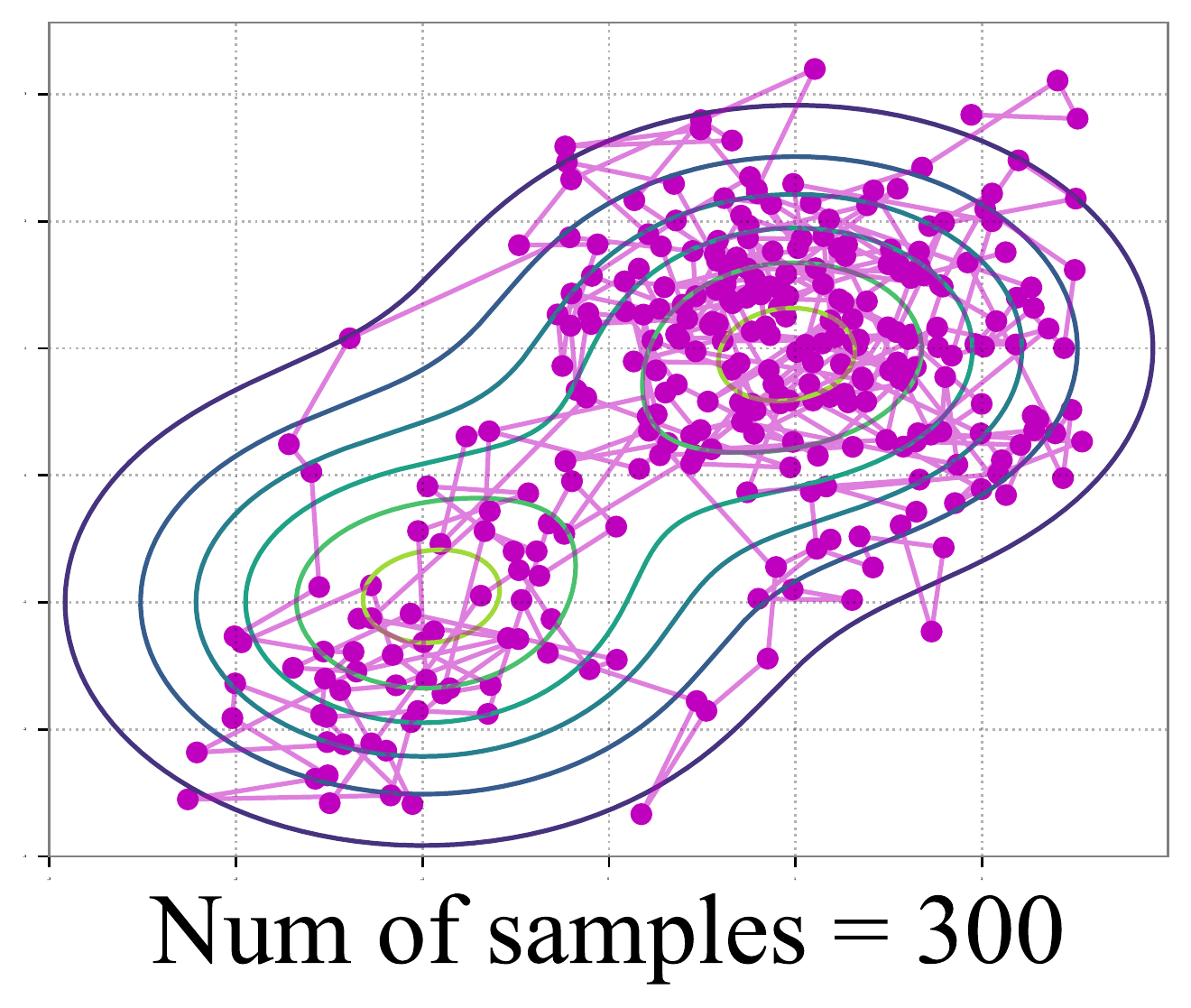}
\par\end{centering}
\begin{centering}
\caption{Sampling trajectory of the mixture of Gaussian.}
\par\end{centering}
\end{figure}

We aim to show how the repulsive gradient helps the particle escape from the local high density region by sampling the 2D mixture of Gaussian distribution using SRLD and Langevin dynamics. The target density is set to be
\begin{align*}
    \rho^{*}(\th)\propto 0.5 \exp{\left(-\left\Vert \th-\boldsymbol{1}\right\Vert ^{2}/2\right)} +0.5\exp{\left(-\left\Vert \th+ \boldsymbol{1}\right\Vert ^{2}/2\right)},
\end{align*}
where $\th=[\theta_1,\theta_2]^\top$ and $\mathbf{1} = [1, 1]^\top$. This target distribution have two mode at $-\boldsymbol{1}$ and $\boldsymbol{1}$, and vanilla Langevin dynamics can stuck in one mode while keeps the another mode under-explored (as the gradient of energy function can dominate the update of samples). We use the same evaluation method, step sizes, initialization and Gaussian noise as the previous experiment. 
We collect one sample every 100 iterations and the experiment is repeated for 20 times. Figure \ref{fig:Sample-quality-mix} shows that 
SRLD consistently outperforms the Langevin dynamics on all of the evaluation metrics.

To provide more evidence on the effectiveness of SRLD on escaping from local high density region, we plot the sampling trajectory of SRLD and vanilla Langevin dynamics on the mixture of Gaussian mentioned in Section \ref{sec:toy_example}. We can find that, when both of the methods obtain 200 samples, SRLD have started to explore the second mode, while vanilla Langevin dynamics still stuck in the original mode. When both of the methods have 250 examples, the vanilla Langevin dynamics just start to explore the second mode, while our SRLD have already obtained several samples from the second mode, which shows our methods effectiveness on escaping the local mode.

\subsection{Synthetic higher dimensional Gaussian Experiment} \label{suppsec: hdgaussian}
\begin{figure}[tb]
\begin{centering}
\includegraphics[scale=0.3]{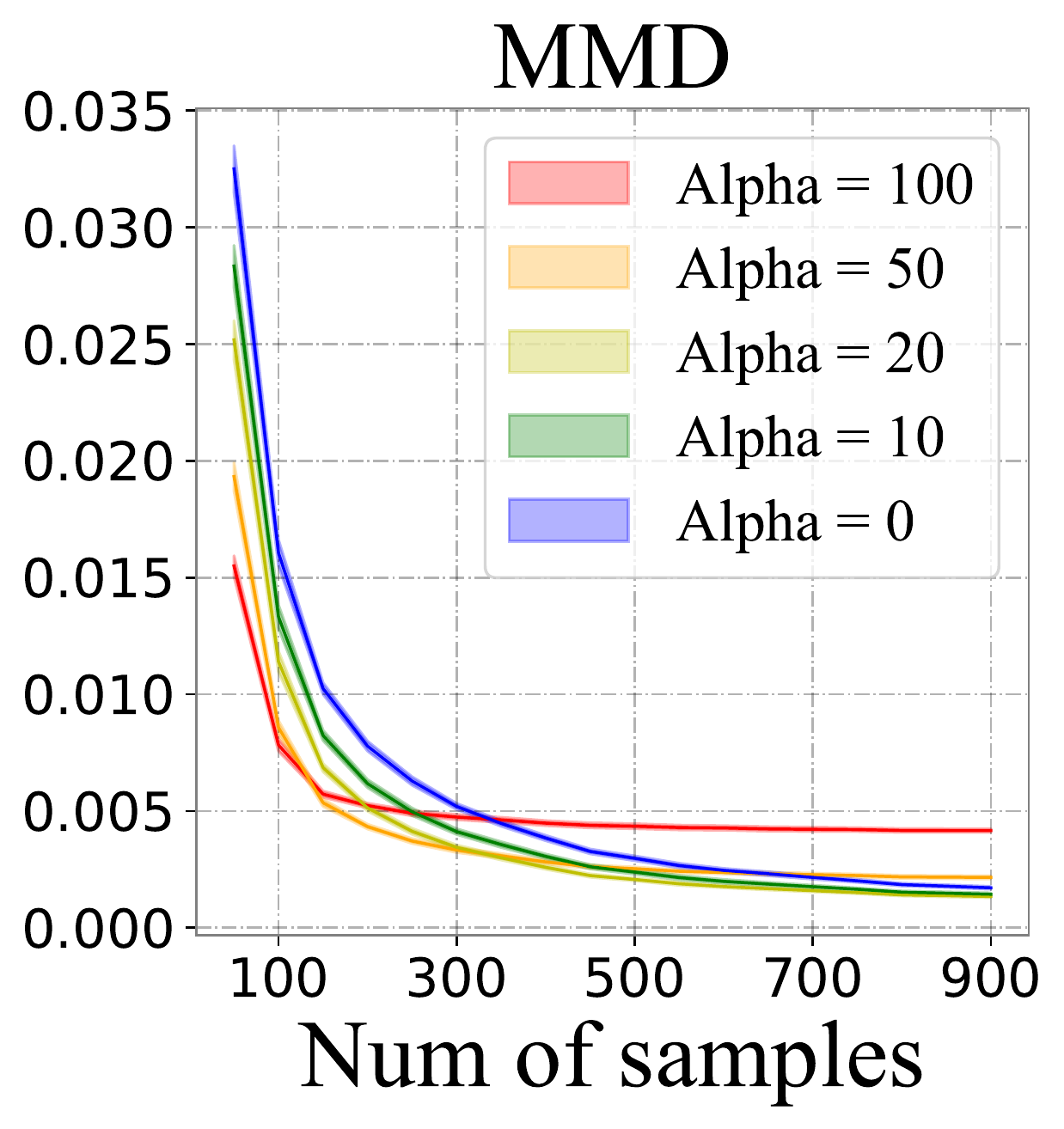}\includegraphics[scale=0.3]{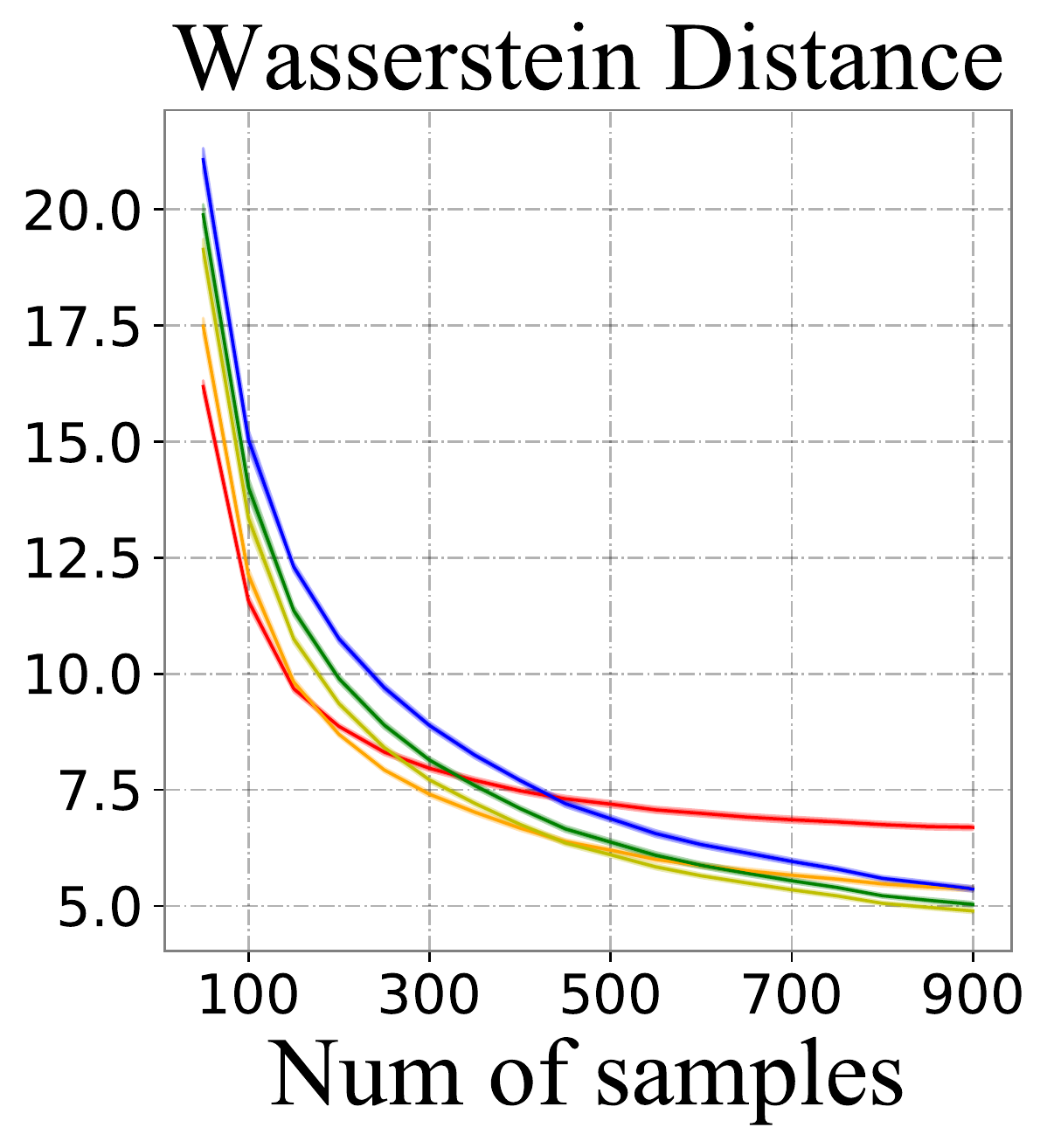}\includegraphics[scale=0.3]{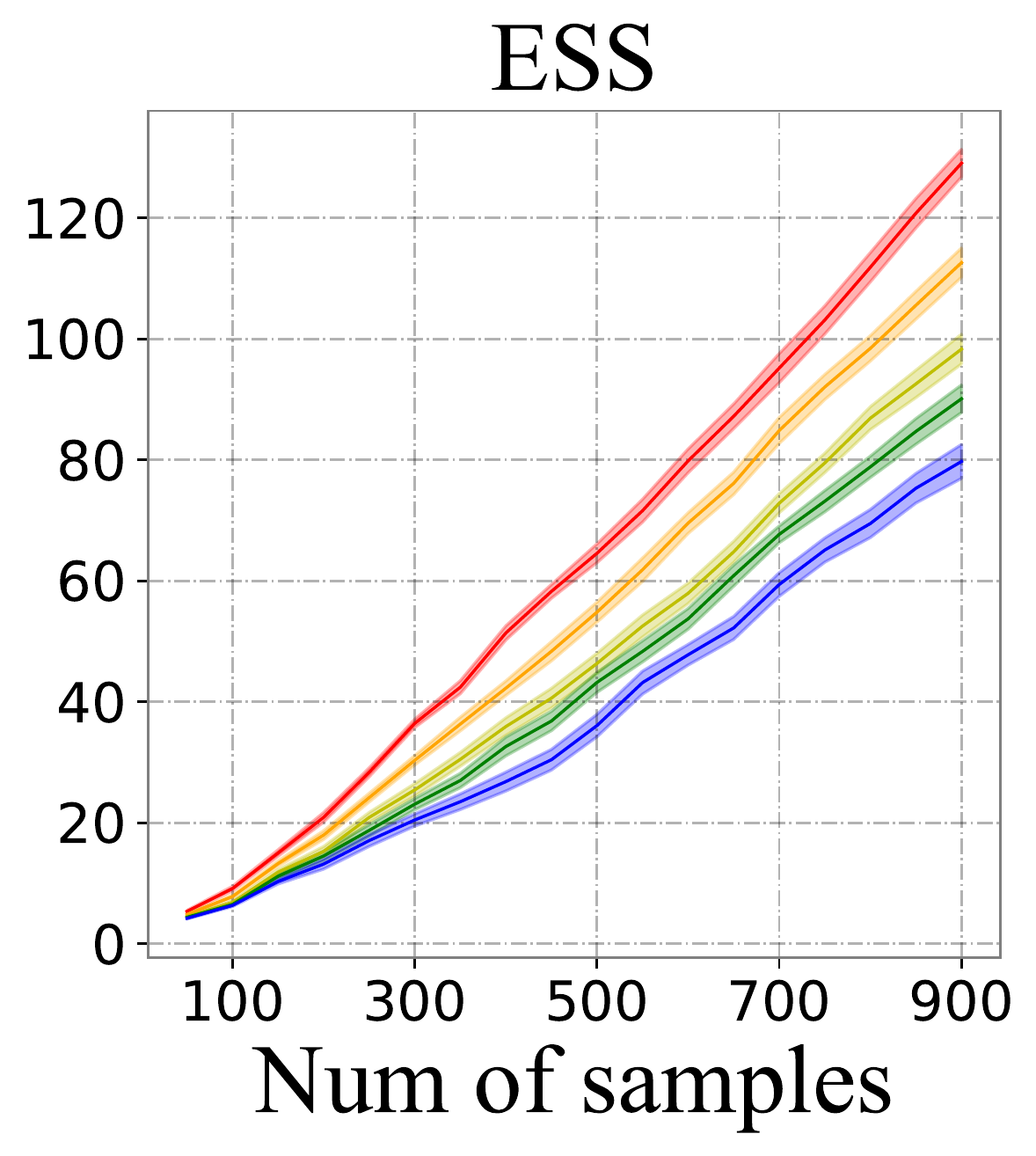}\includegraphics[scale=0.3]{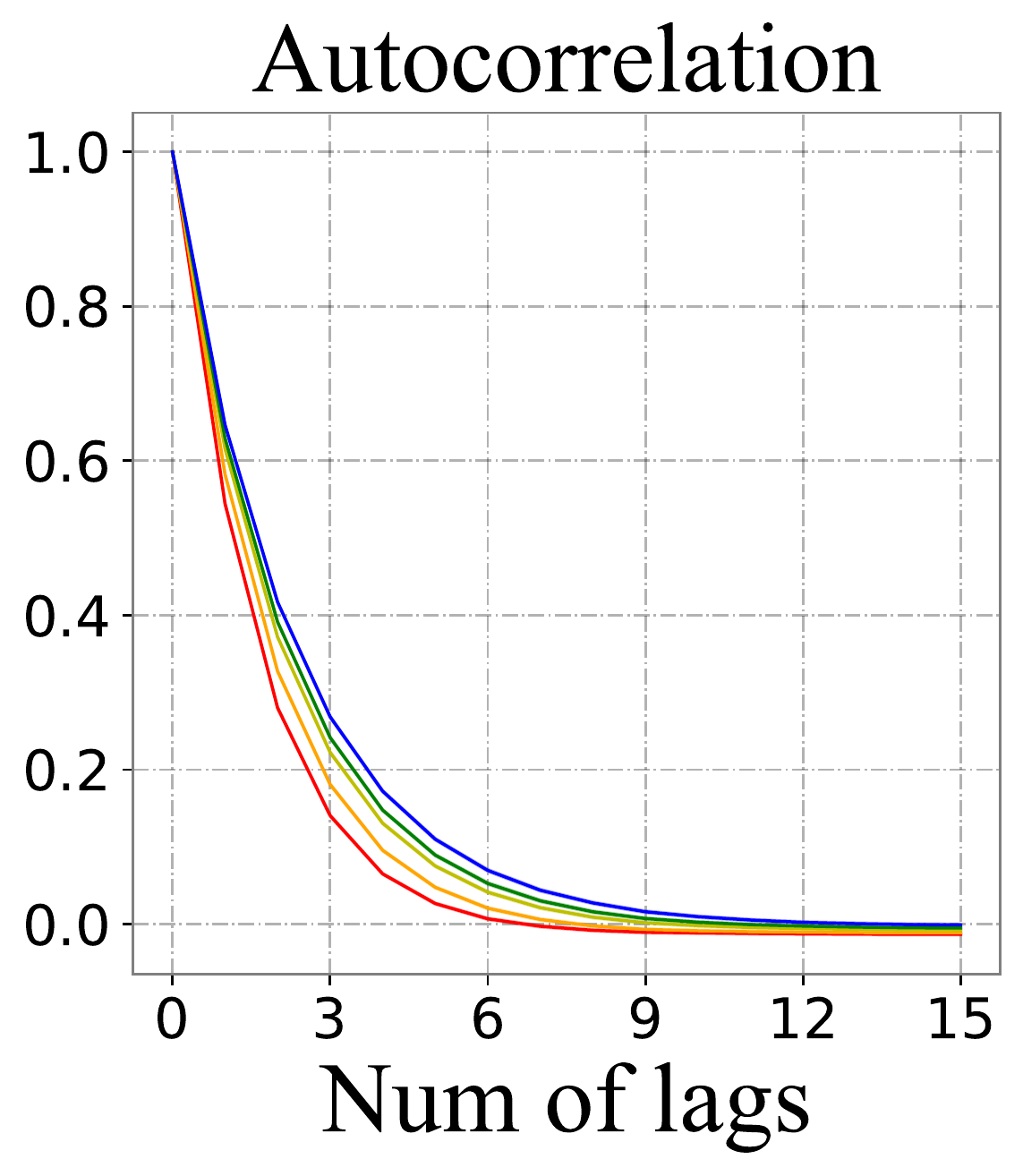}
\par\end{centering}
\caption{Sample quality and autocorrelation of the higher dimensional Gaussian distribution. The auto-correlation is the averaged auto-correlation of all dimensions. \label{fig:Sample-quality-N} }
\end{figure}

To show the performance of SRLD in higher dimensional case with different
value of $\alpha$, we additionally considering the problem on sampling from Gaussian distribution with $d=100$ and covariance $\boldsymbol{\Sigma} = 0.5\textbf{I}$. We run SRLD with $\alpha=100, 50, 20, 10, 0$ and the case $\alpha=0$ reduces to Langevin. We collect 1 sample every 10 iterations. The other experiment setting
is the same as the toy examples in the main text. The results are summarized at Figure \ref{fig:Sample-quality-N}. In this experiment, we set one SRLD with an inappropriate $\alpha=100$. For this chain, the repulsive gradient gives strong repulsive force and thus has the largest ESS and the fastest decay of autocorrelation. While the inappropriate value $\alpha$ induces too much extra approximation error and thus its performance is not as good as these with smaller $\alpha$ (see MMD and Wasserstein distance). This phenomenon matches our theoretical finding.

\subsection{Synthetic higher dimensional Mixture of Gaussian Experiment}  \label{sec:highdimgaussian}
\begin{figure}[tb]
\begin{centering}
\includegraphics[scale=0.3]{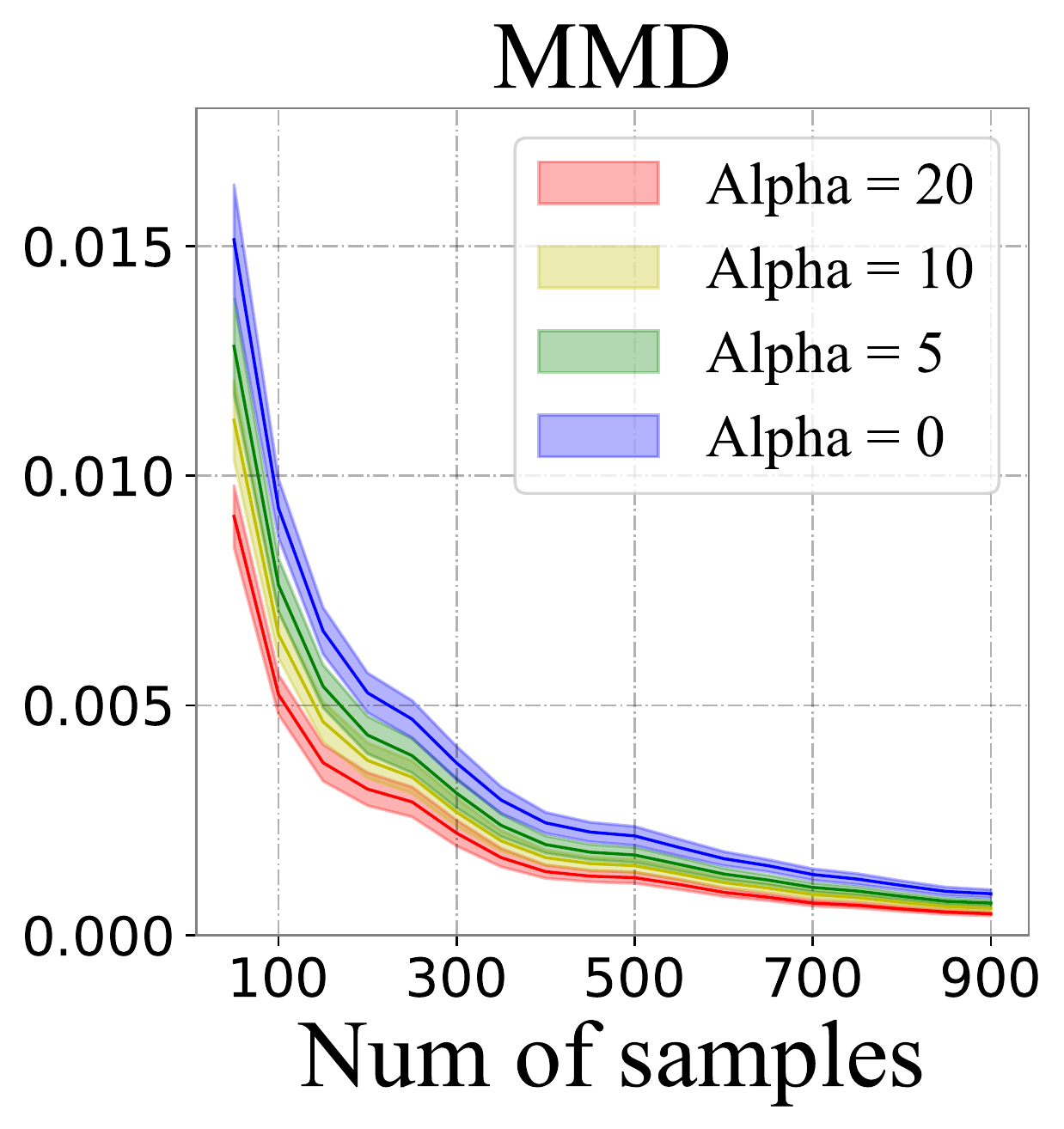}\includegraphics[scale=0.3]{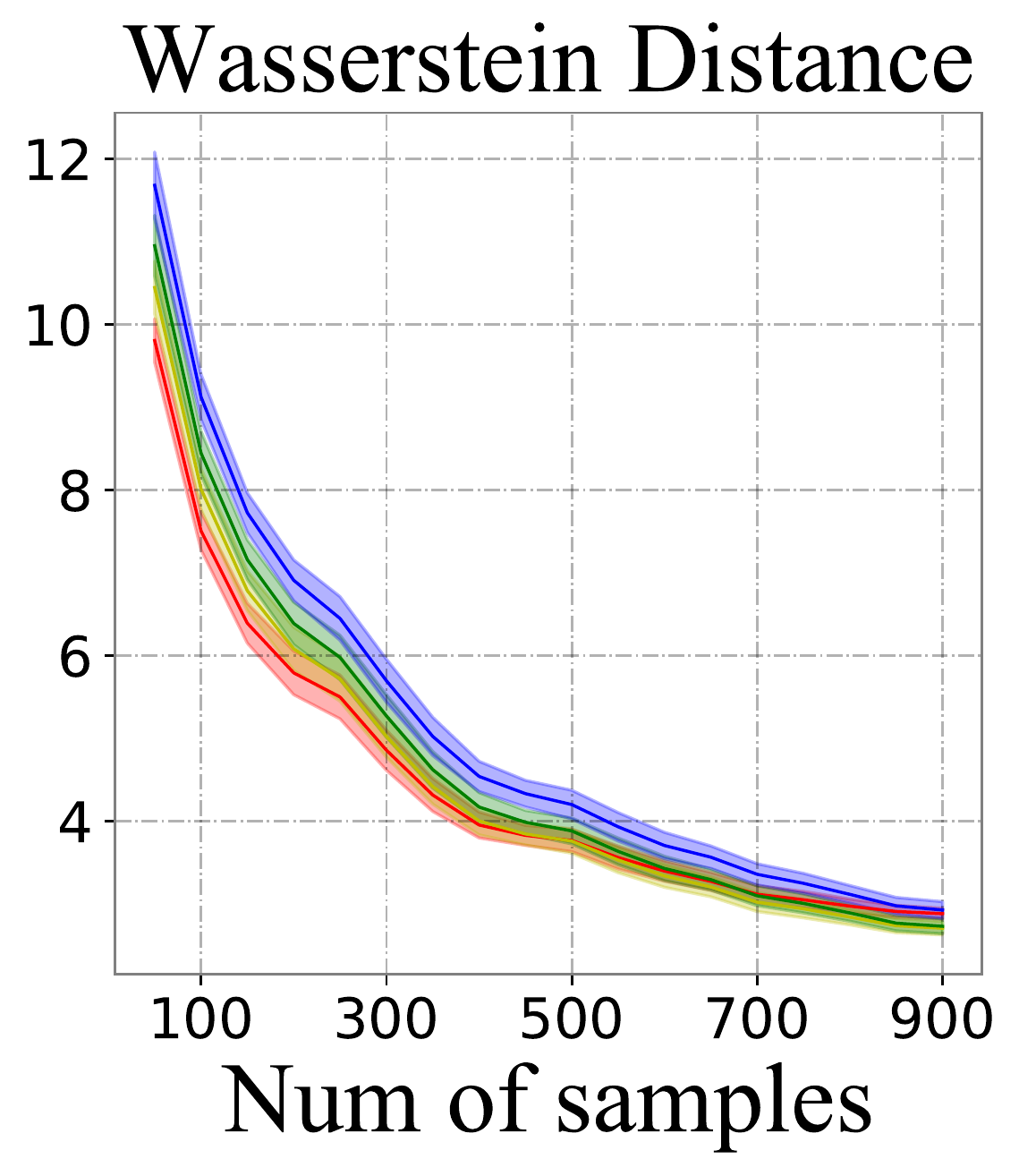}\includegraphics[scale=0.3]{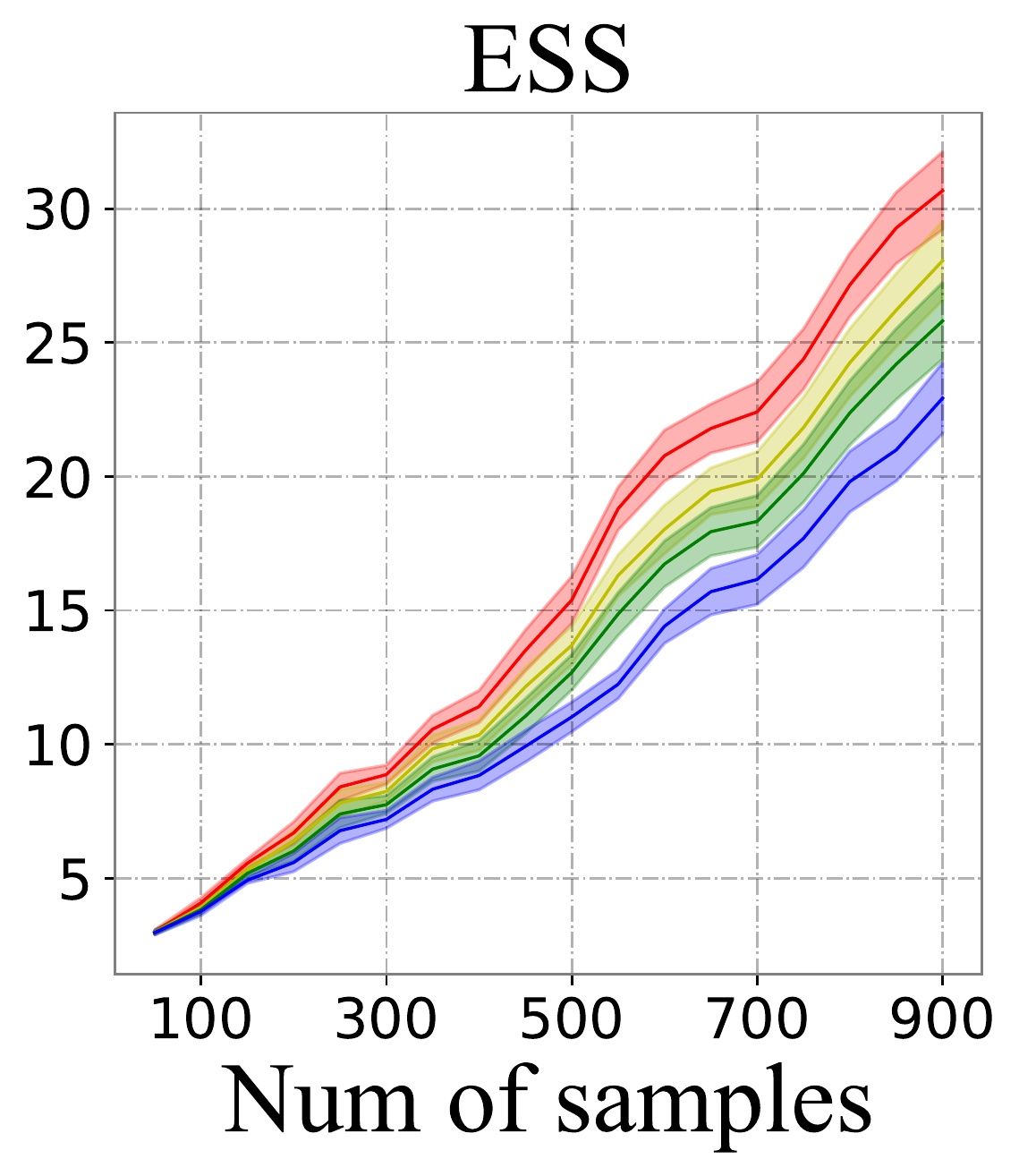}\includegraphics[scale=0.3]{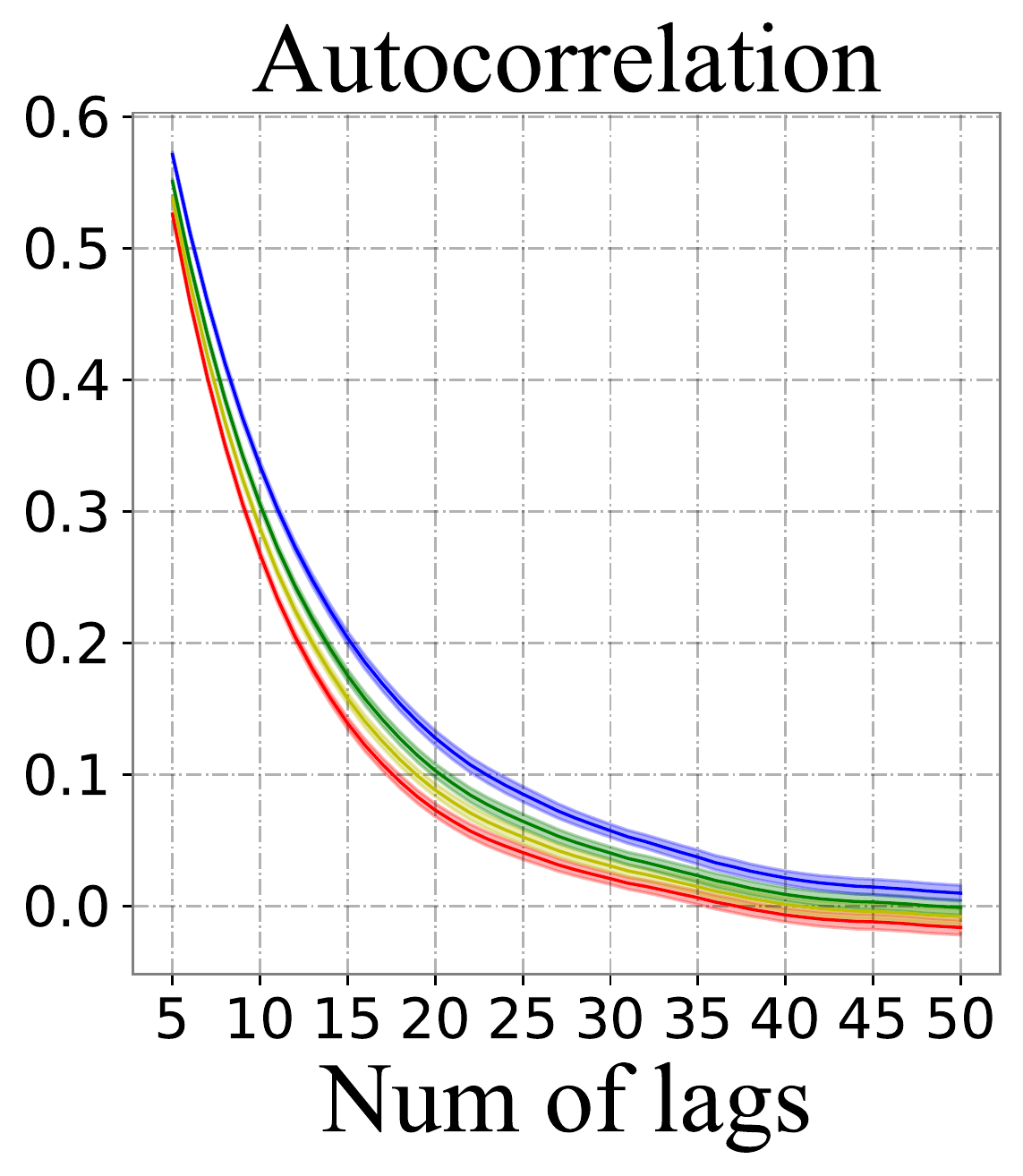}
\par\end{centering}
\caption{Sample quality and autocorrelation of the higher dimensional mixture distribution. The auto-correlation is the averaged auto-correlation of all dimensions. \label{fig:Sample-quality-Hmix} }
\end{figure}

We also consider sampling from the mixture of Gaussian with $d=20$. The
target density is set to be
\[
\rho^{*}(\boldsymbol{\theta})\propto\frac{1}{2}\exp\left(-0.5\left\Vert \boldsymbol{\theta}-\sqrt{2/d}\boldsymbol{1}\right\Vert ^{2}\right)+\frac{1}{2}\exp\left(-0.5\left\Vert \boldsymbol{\theta}+\sqrt{2/d}\boldsymbol{1}\right\Vert ^{2}\right),
\]
where $\boldsymbol{\theta}=[\theta_{1},...,\theta_{20}]^{\top}$ and
$\boldsymbol{1}=[1,...,1]^{\top}$. And thus the mean of the two mixture
component is with distance $2\sqrt{2}$. We run SRLD with $\alpha=20,10,5,0$
(and when $\alpha=0$, it reduces to LD). The other experiment setting
is the same as the low dimensional mixture Gaussian case. Figure [\ref{fig:Sample-quality-Hmix}]
summarizes the result. As shown in the figure, when $\alpha$ becomes
larger, the repulsive forces helps the sampler better explore the
density region.

\section{BNN on UCI Datasets: Experiment Settings and Additional Results} \label{sec:uci_appendix}
We first give detailed experiment settings. We set a $\Gamma(1, 0.1)$ prior for the inverse output variance. 
We set the mini-batch size to be 100. We run 50000 iterations for each methods, and for LD and SRLD, the first 40000 iteration is discarded as burn-in.
We use a thinning factor of $c_\eta = c/\eta=100$ and in total we collect 100 samples from the posterior distribution. For each dataset, we generate 3 extra data splits for tuning the step size for each method. the number of past samples $M$ to be 10.
In all experiments, we use RBF kernel with bandwidth set by the median trick as suggested in \cite{liu2016stein}. We use $\alpha=10$ for all the data sets.
For SVGD, we use the original implementation with 20 particles by \cite{liu2016stein}.

We show some additional experiment result on posterior inference on UCI datasets. As mentioned in Section \ref{sec:uci}, the comparison between SVGD and SRLD is not direct as SVGD is a multiple-chain method with fewer particles and SRLD is a single chain method with more samples. To show more detailed comparison, we compare the SVGD with SRLD using the first 20, 40, 60, 80 and 100 samples, denoted as SRLD-$n$ where $n$ is the number of samples used. Table \ref{tb:SRLD_SVGD_RMSE} shows the result of averaged test RMSE and table \ref{tb:SRLD_SVGD_ll} shows the result of averaged test loglikelihood. For SRLD with different number of samples, the value is set to be boldface if it has better average performance than SVGD. If it is statistical significant with significant level 0.05 using a matched pair t-test, we add an underline on it.

\begin{figure}[h]
\begin{centering}
\includegraphics[scale=0.3]{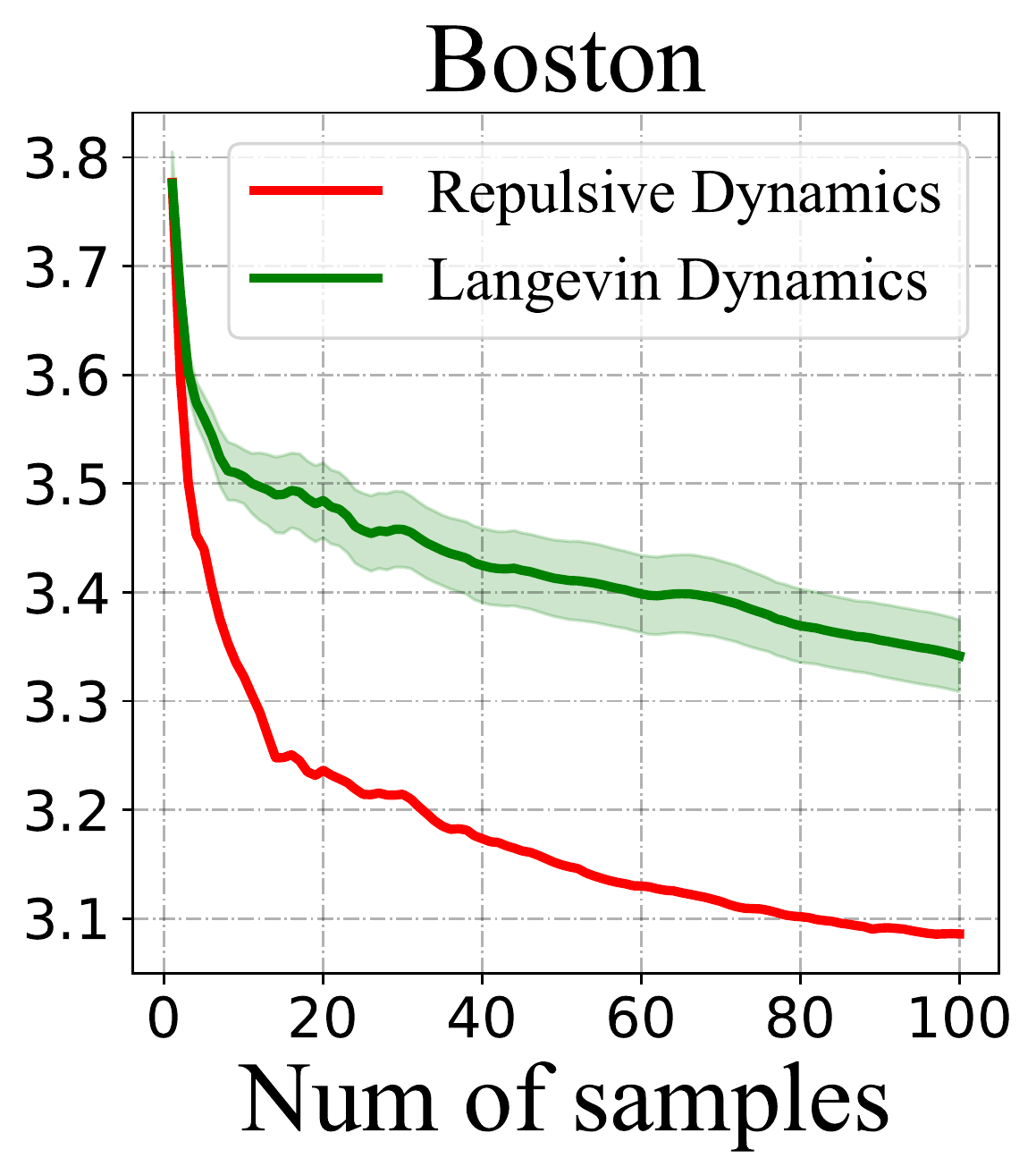}\includegraphics[scale=0.3]{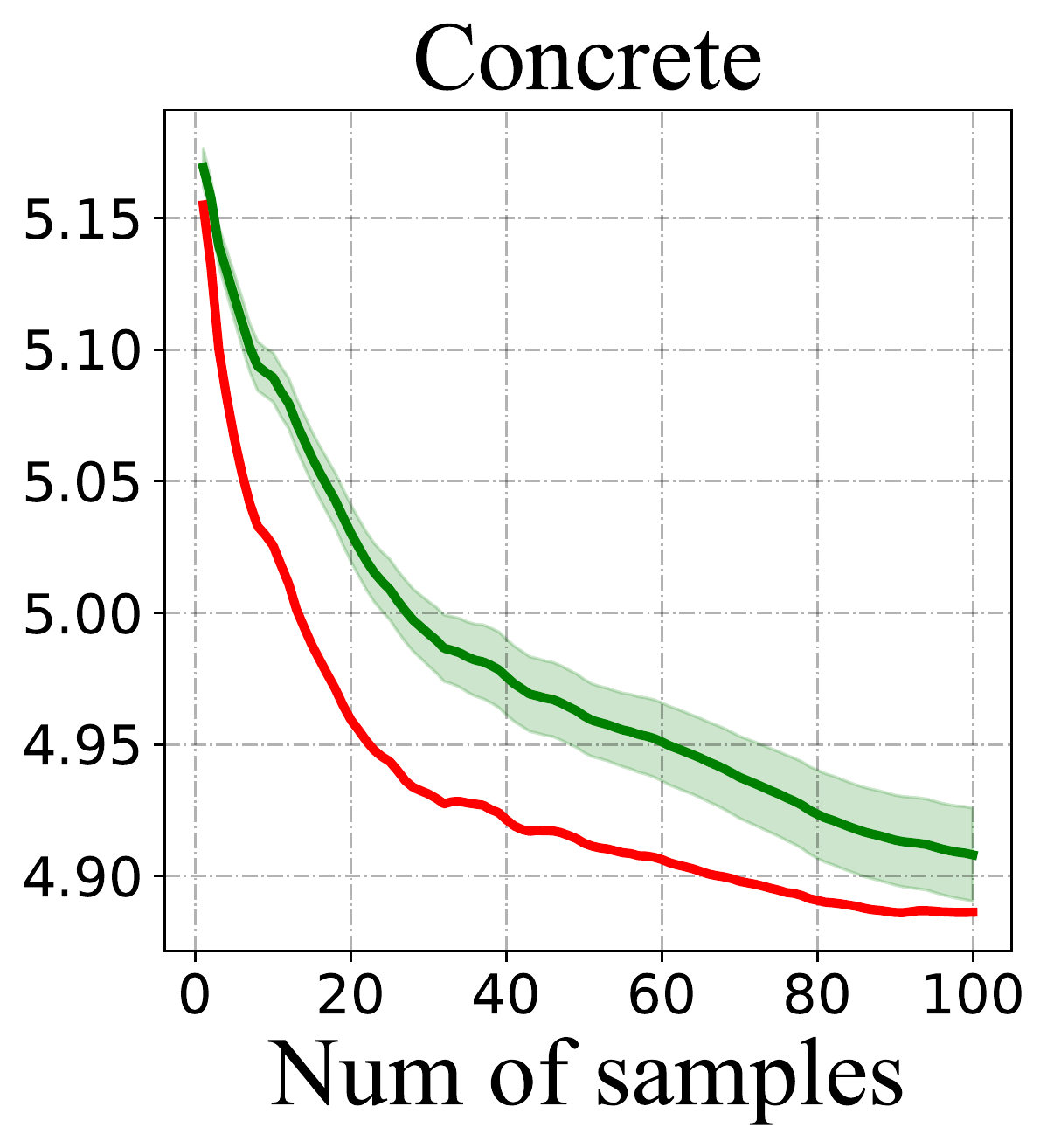}\includegraphics[scale=0.3]{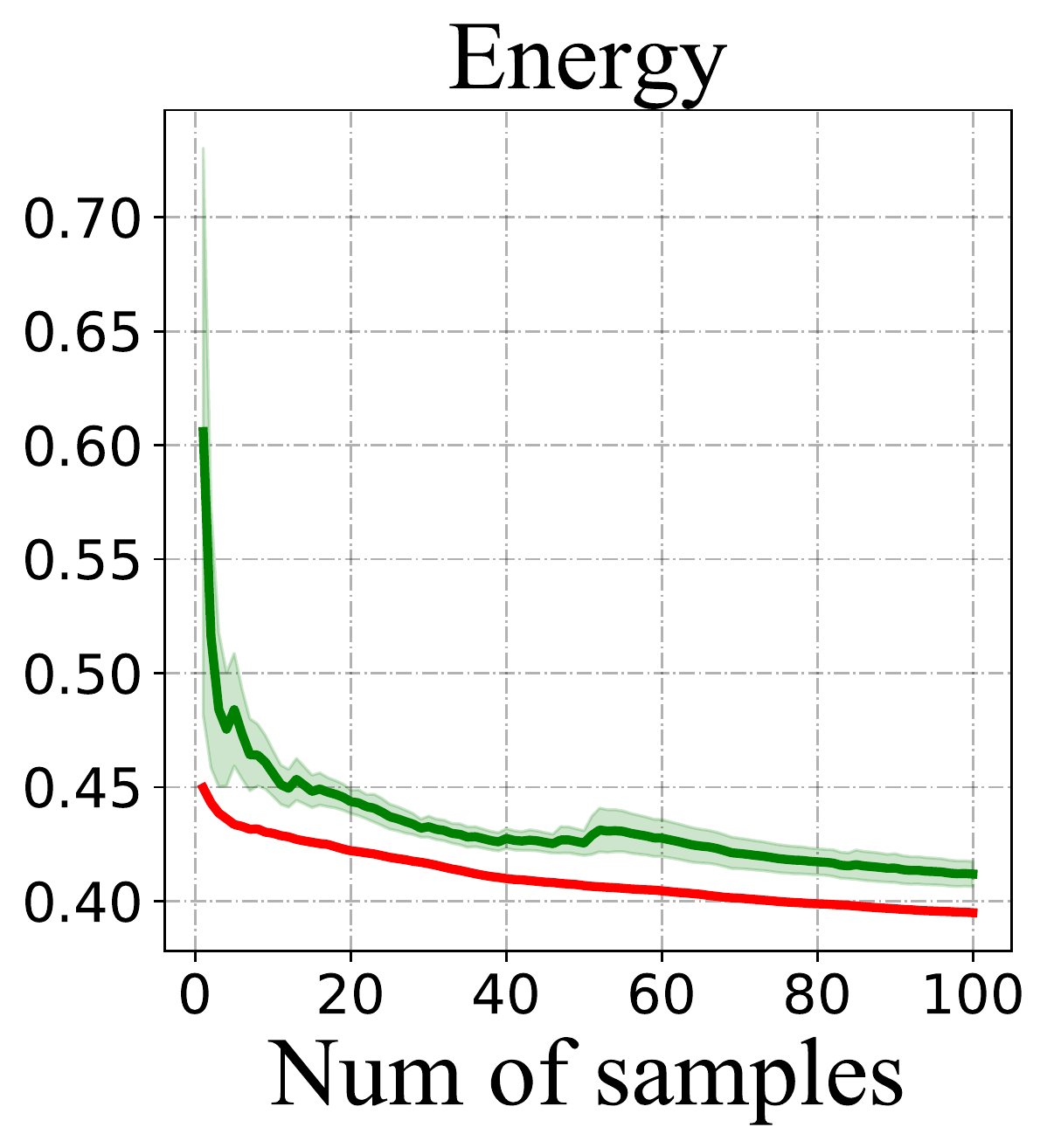}\includegraphics[scale=0.3]{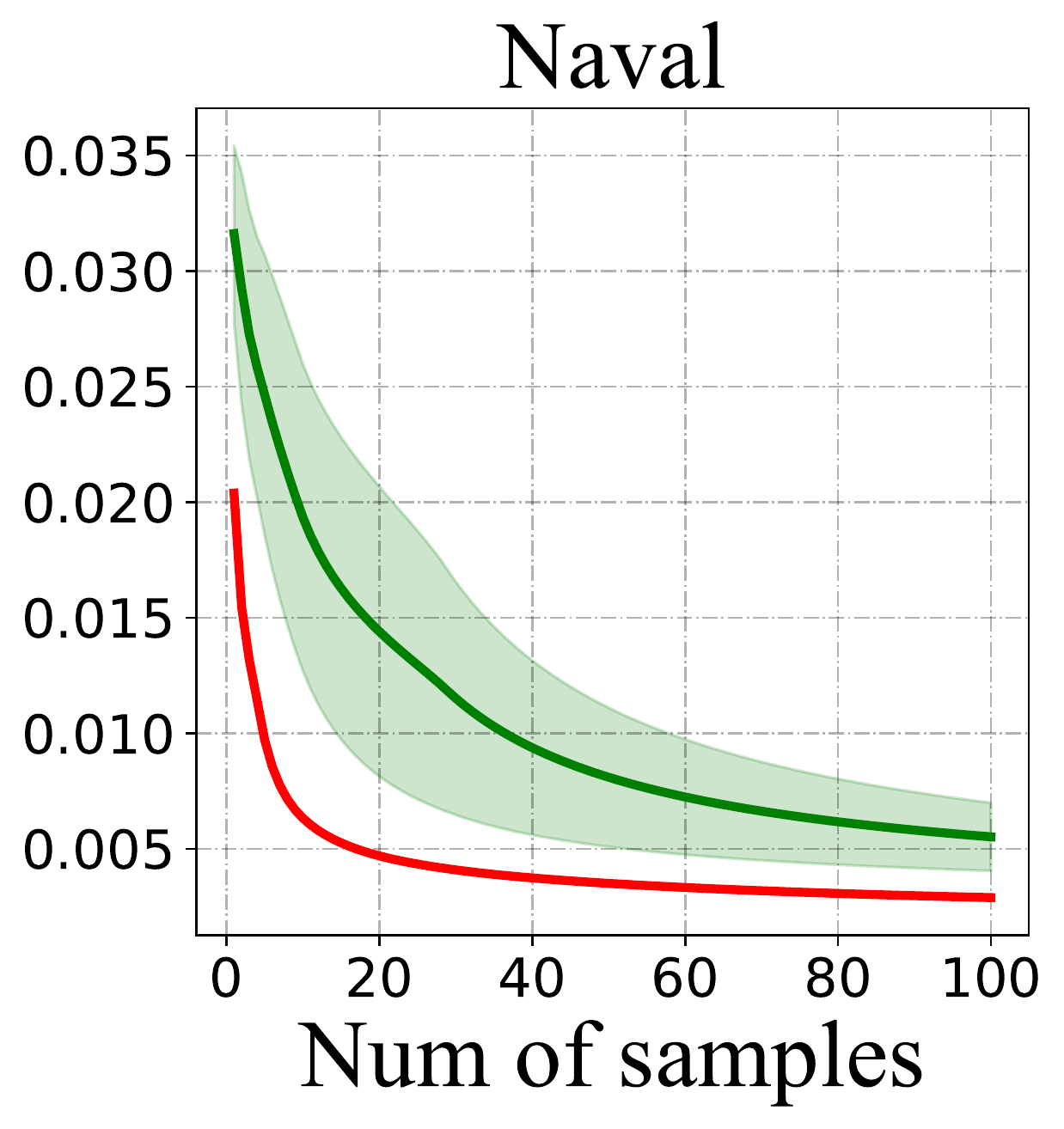}
\par\end{centering}
\begin{centering}
\includegraphics[scale=0.3]{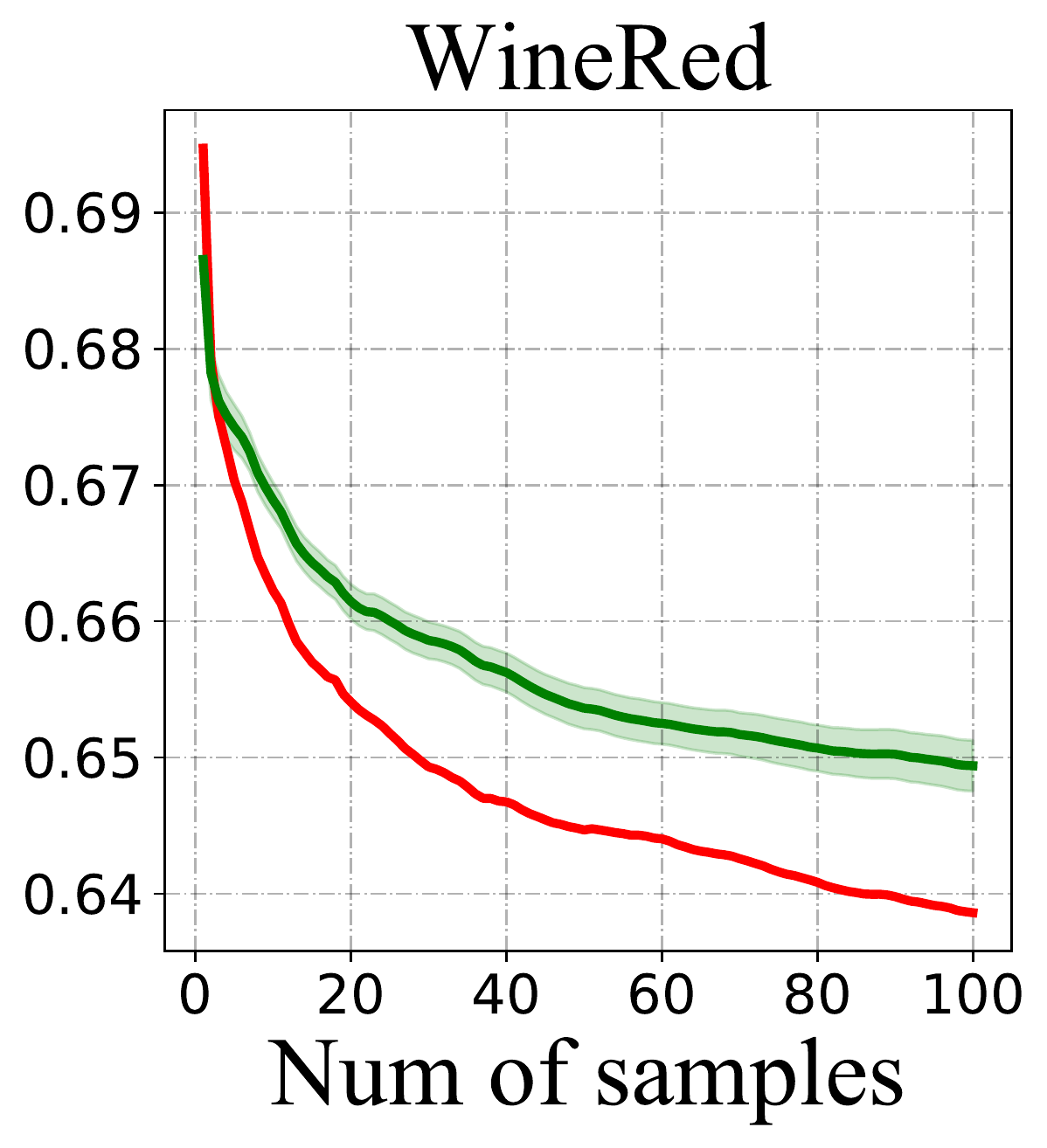}\includegraphics[scale=0.3]{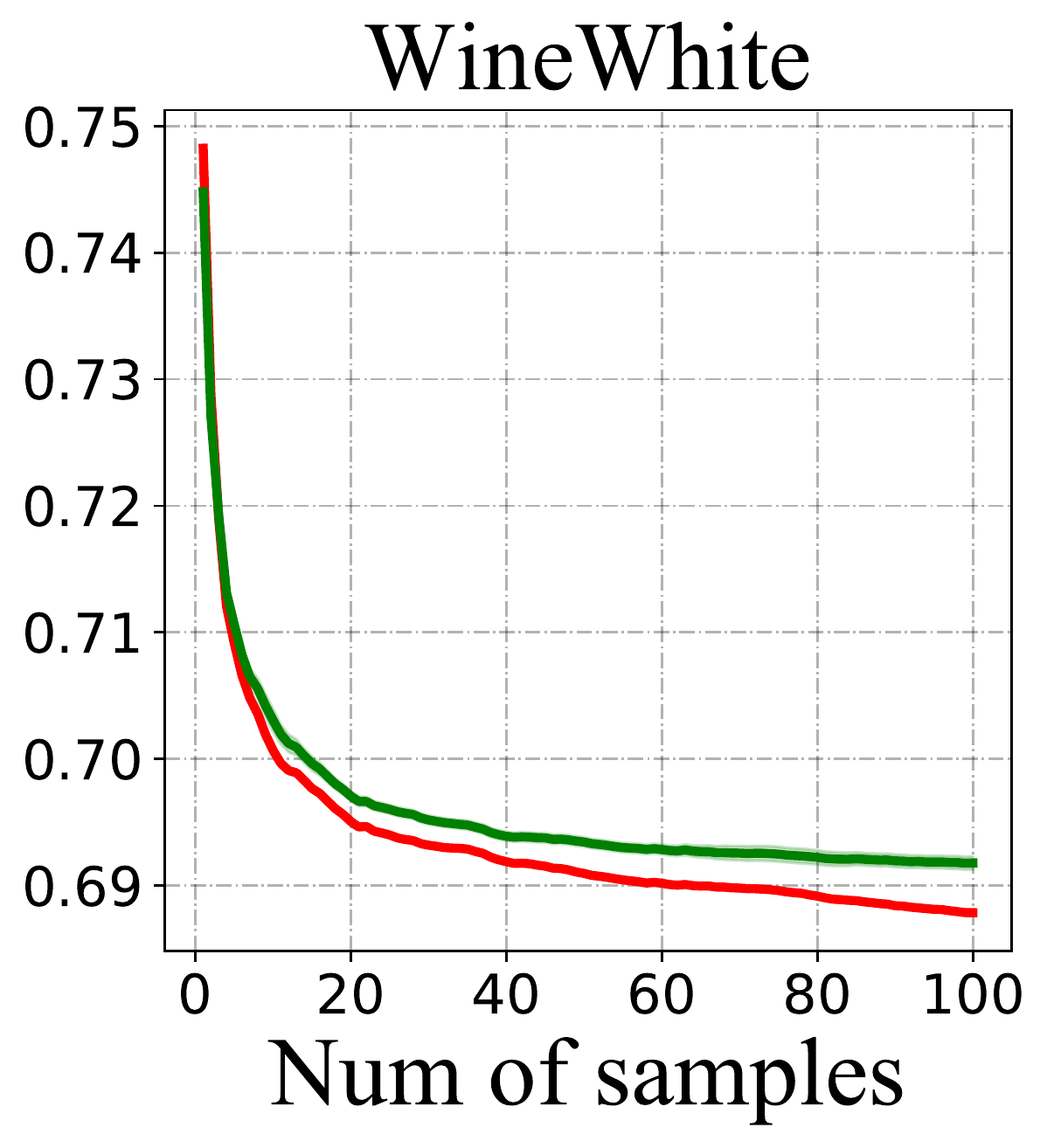}\includegraphics[scale=0.3]{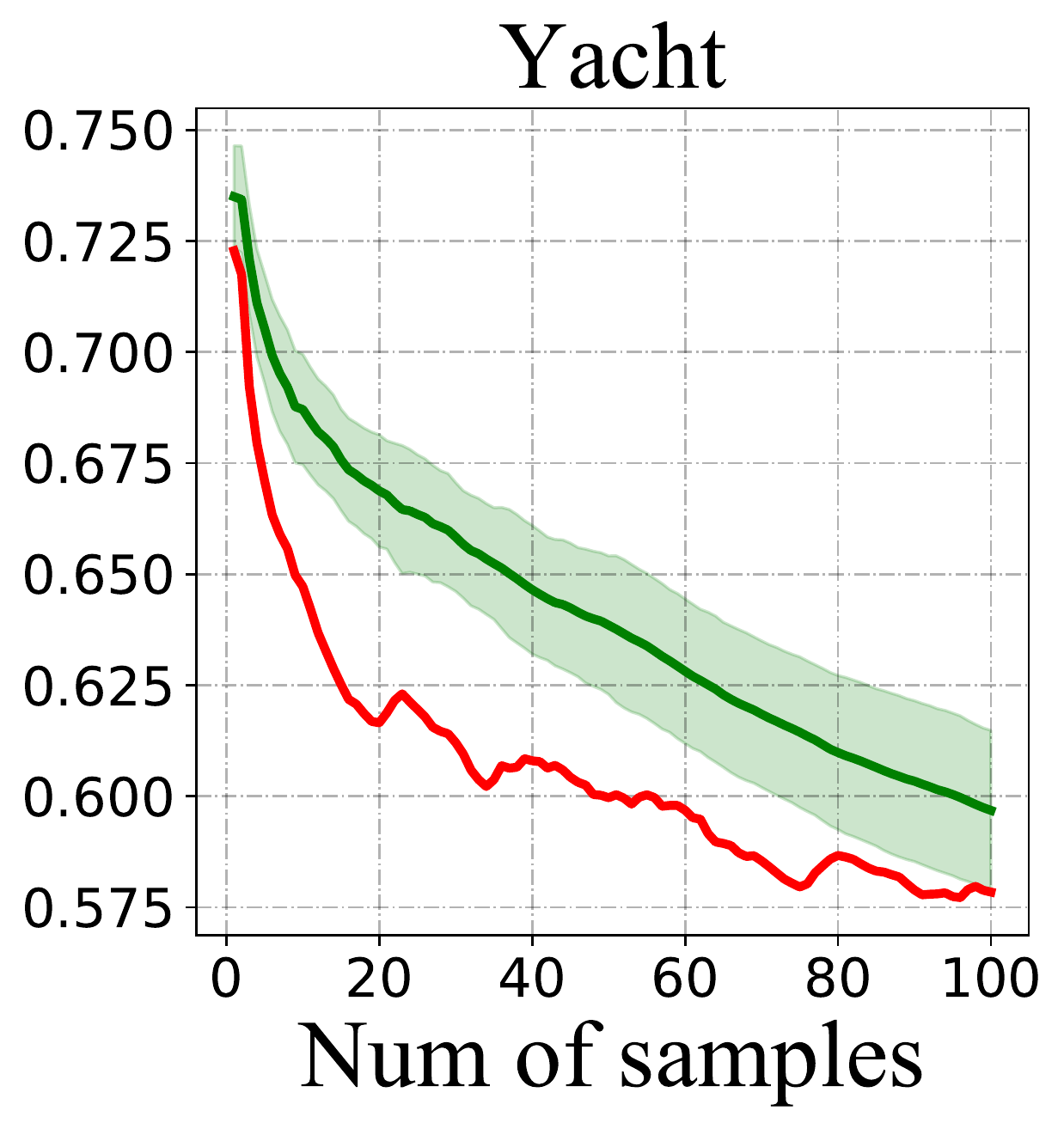}
\par\end{centering}
\begin{centering}
\caption{Comparison between SRLD and Langevin dynamics on test RMSE. The results are computed based on 20 repeats. The error bar is calculated based on RMSE of SRLD - RMSE of Langevin dynamics in 20 repeats to rule out the variance of different data splitting} \label{fig: SRLD_LD_RMSE}
\par\end{centering}
\end{figure}

\begin{figure}[h]
\begin{centering}
\includegraphics[scale=0.3]{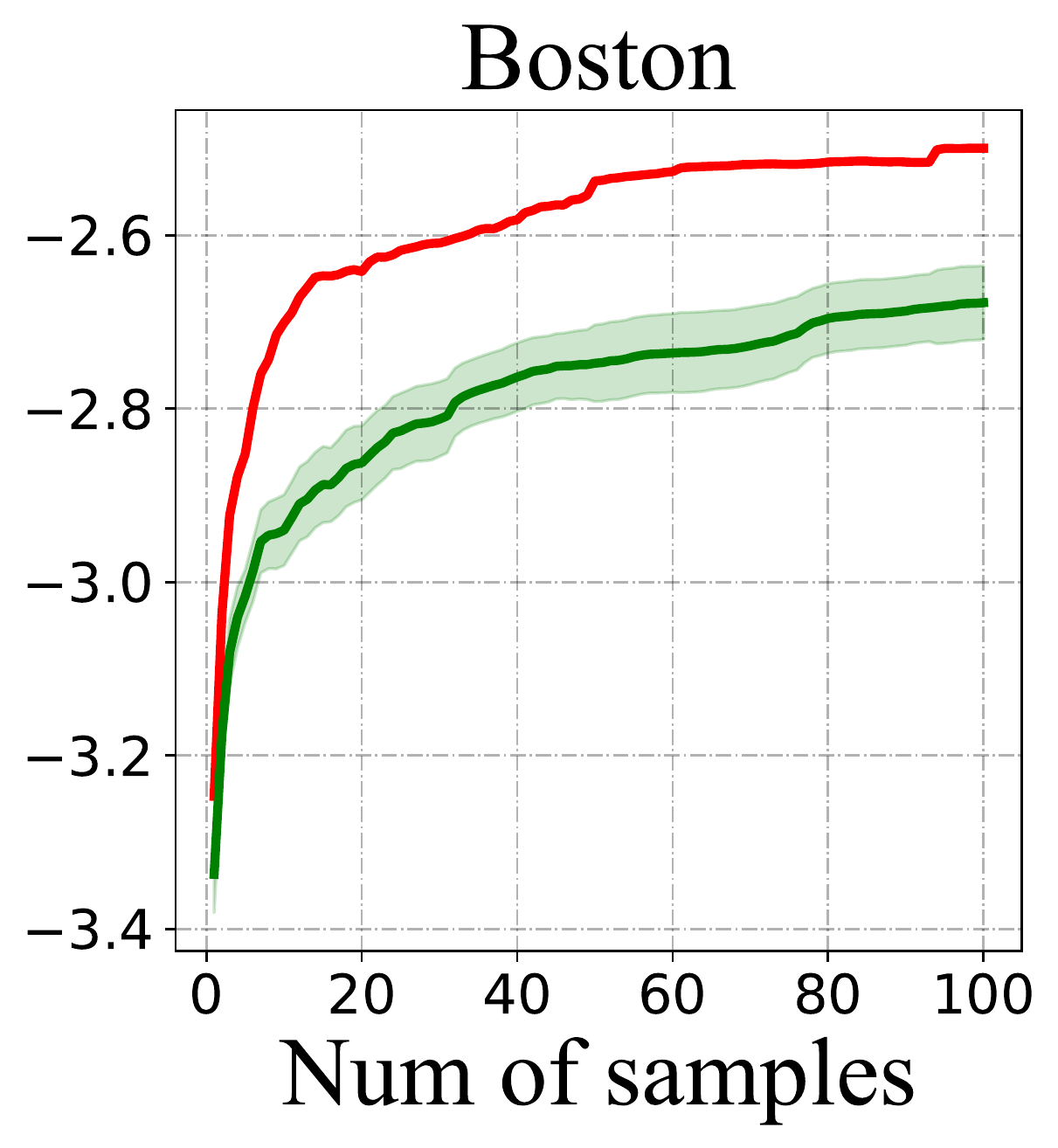}\includegraphics[scale=0.3]{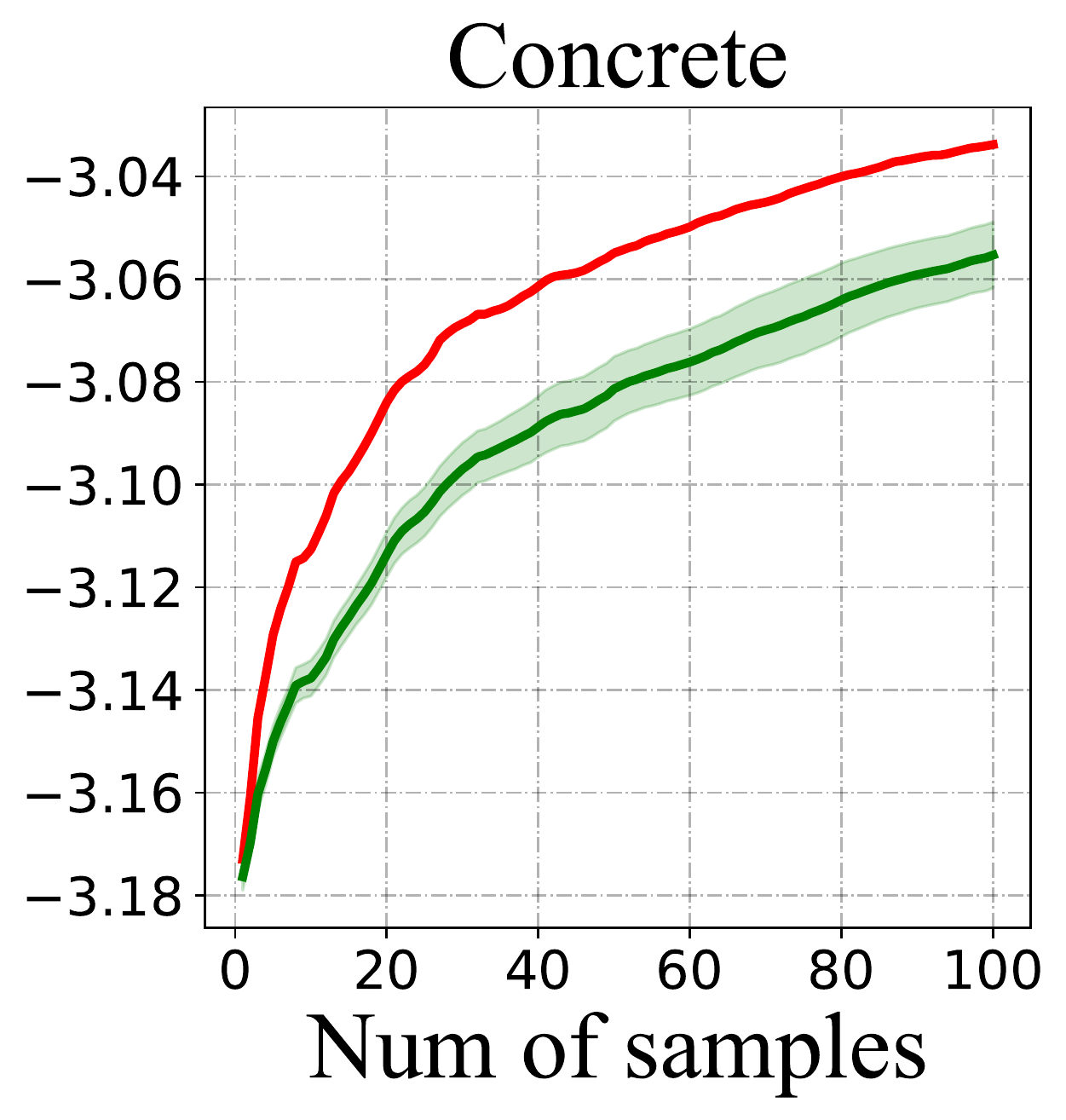}\includegraphics[scale=0.3]{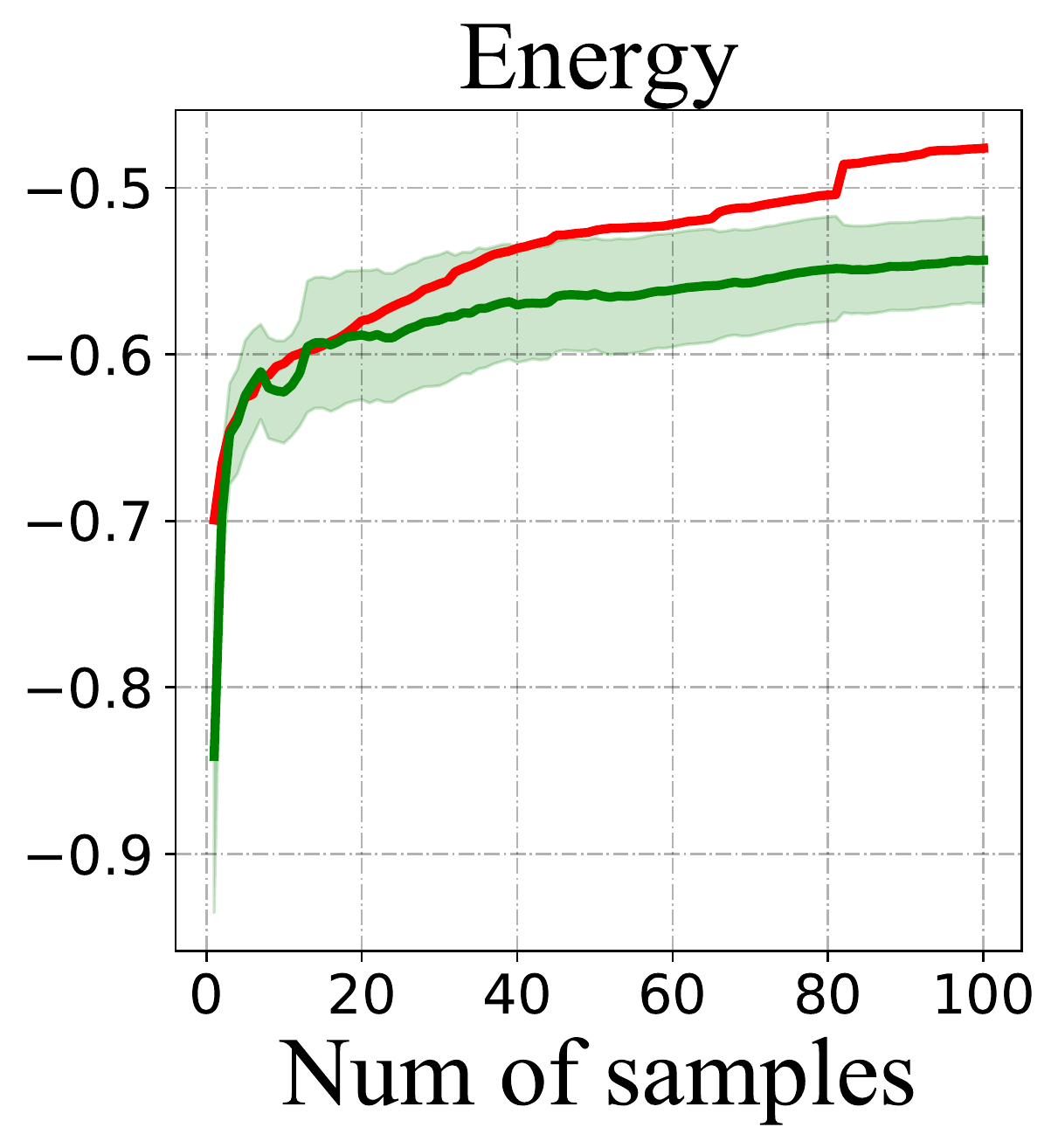}\includegraphics[scale=0.3]{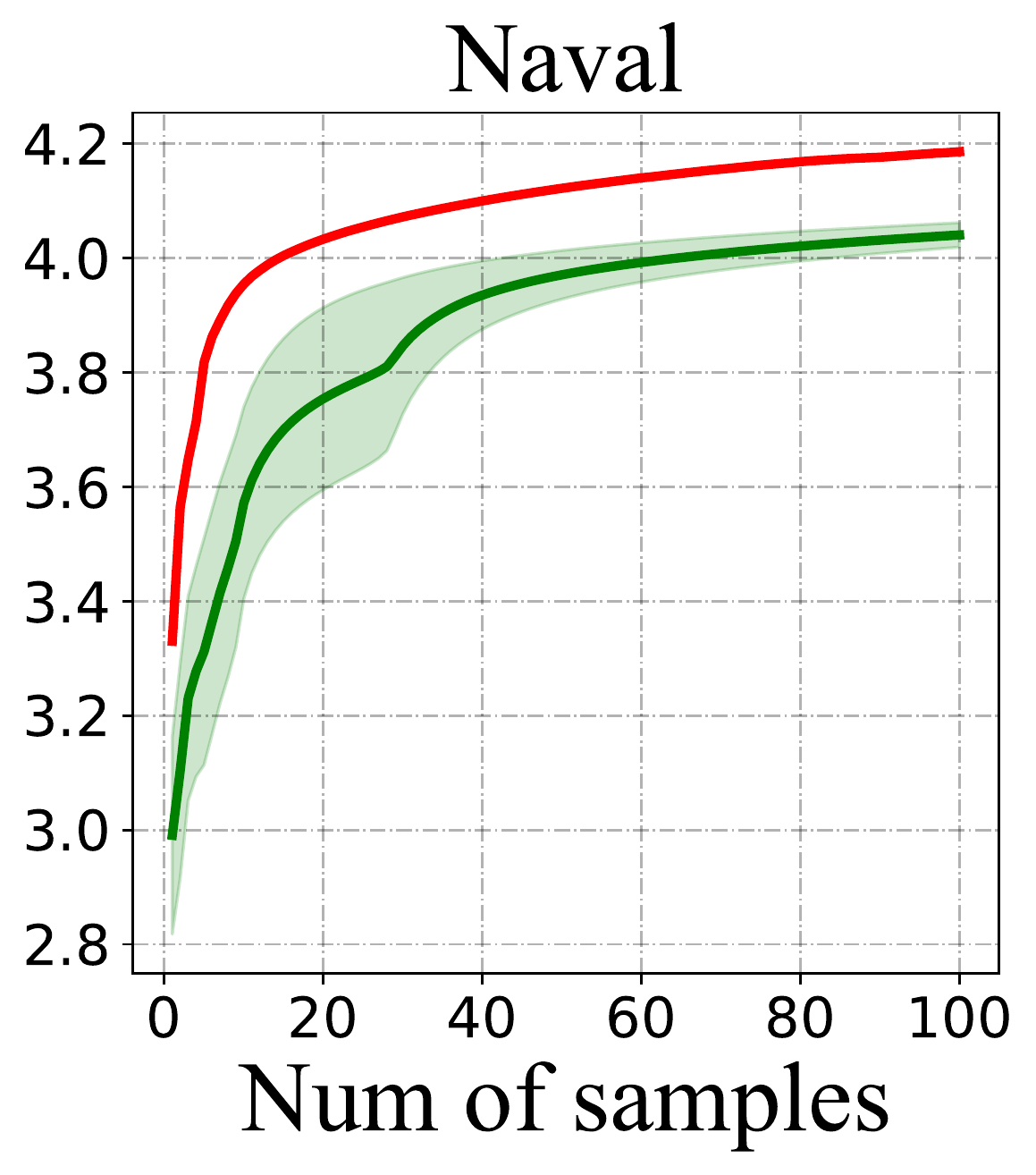}
\par\end{centering}
\begin{centering}
\includegraphics[scale=0.3]{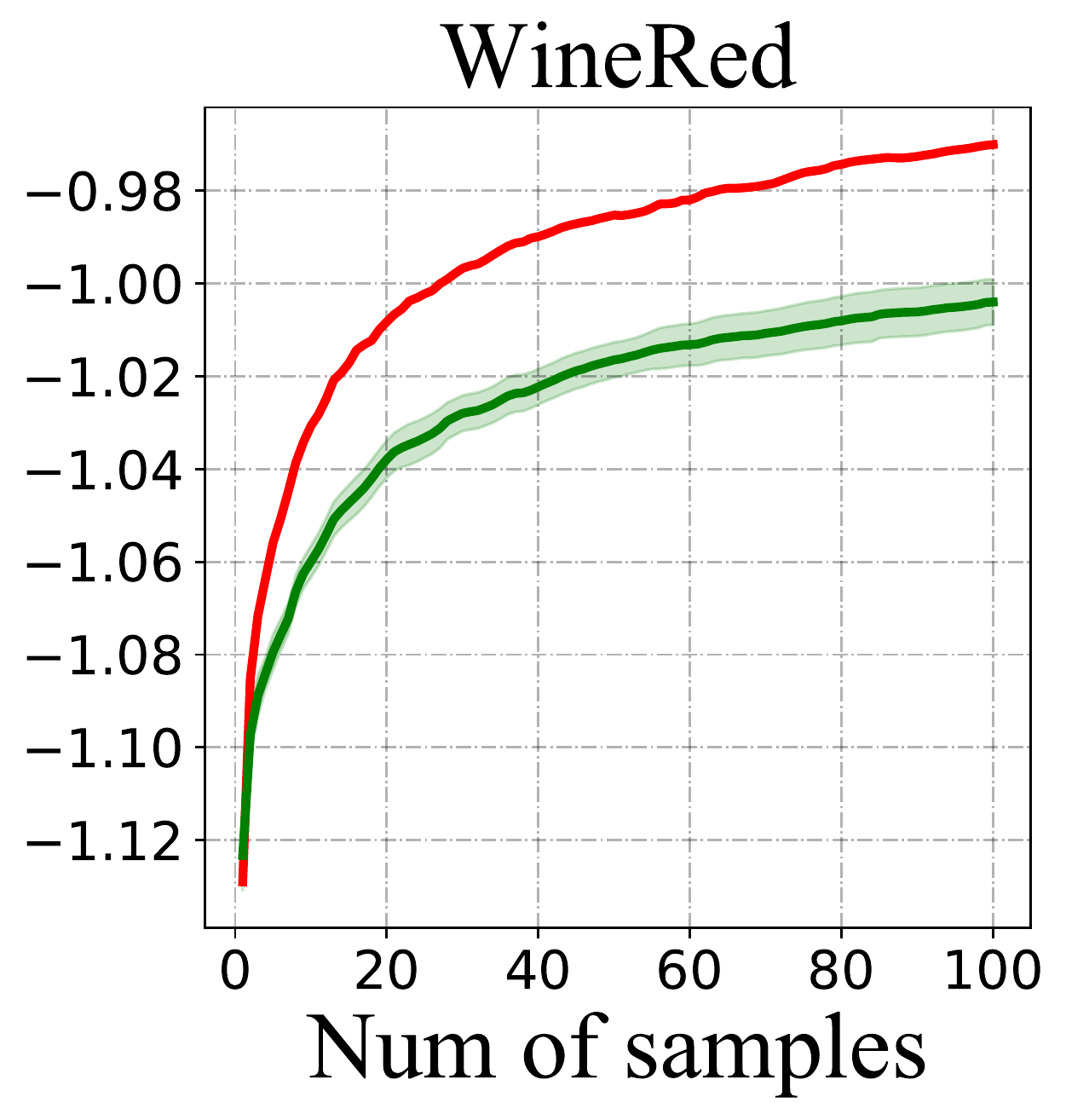}\includegraphics[scale=0.3]{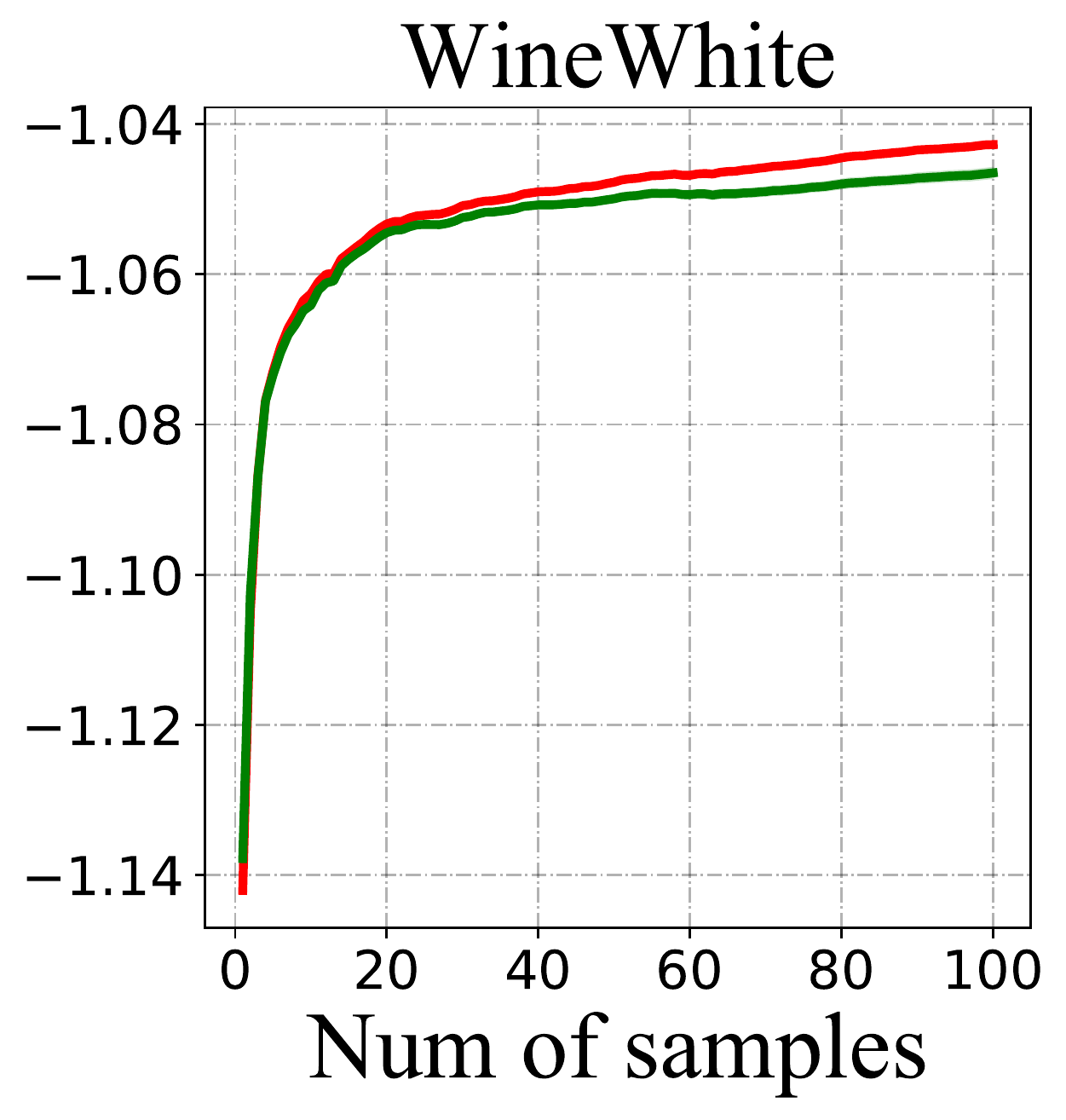}\includegraphics[scale=0.3]{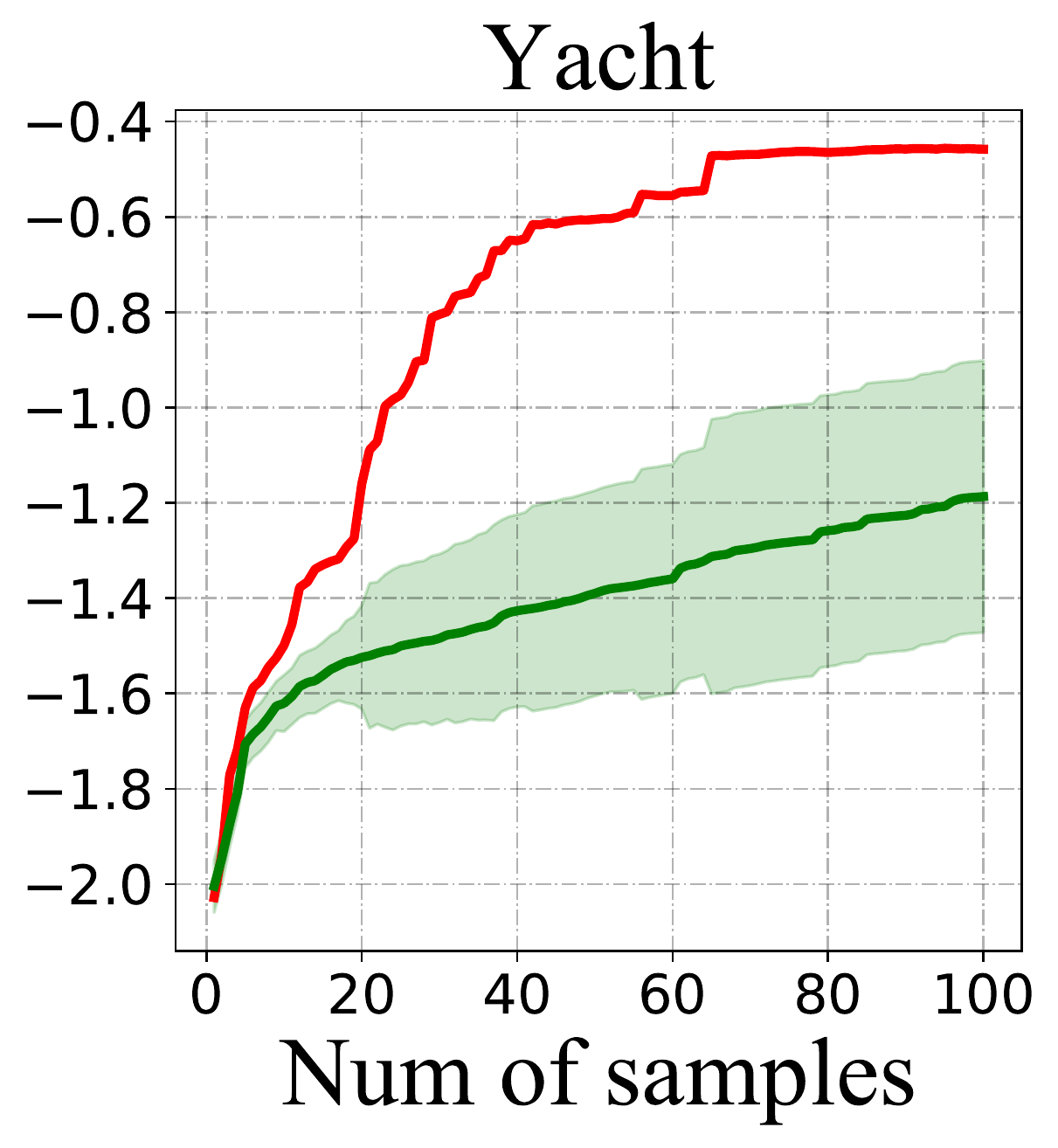}
\par\end{centering}
\centering{}\caption{Comparison between SRLD and Langevin dynamics on test log-likelihood. The results are computed based on 20 repeats. The error bar is calculated based on log-likelihood of SRLD - log-likelihood of Langevin dynamics in 20 repeats to rule out the variance of data splitting.}\label{fig: SRLD_LD_ll}
\end{figure}

Figure \ref{fig: SRLD_LD_RMSE} and \ref{fig: SRLD_LD_ll} give some visualized result on the comparison with Langevin dynamics and SRLD. To rule out the variance of different splitting on the dataset, the errorbar is calculated based on the difference between RMSE of SRLD and RMSE of Langevin dynamcis in 20 repeats (And similarily for test log-likelihood). And we only applied the error bar on Langevin dynamics.

\begin{table}[h]
\scriptsize
\setlength\tabcolsep{3.pt}
\begin{centering}
\begin{tabular}{c|cccccc}
\hline 
\multirow{2}{*}{Dataset} & \multicolumn{6}{c}{Ave Test RMSE}\tabularnewline
 & SRLD-20 & SRLD-40 & SRLD-60 & SRLD-80 & SRLD-100 & SVGD\tabularnewline
\hline 
Boston & $\bf{3.236\pm0.174}$ & $\bf{3.173\pm0.176}$ & $\bf{3.130\pm0.173}$ & $\underline{\bf{3.101\pm0.179}}$ & $\underline{{\bf 3.086\pm0.181}}$ & $3.300\pm0.142$ \tabularnewline
Concrete & ${\bf 4.959\pm0.109}$ & ${\bf 4.921\pm0.111}$ & ${\bf 4.906\pm0.109}$ & ${\bf 4.891\pm0.108}$ & ${\bf 4.886\pm0.108}$ & $4.994\pm0.171$\tabularnewline
Energy & ${\bf 0.422\pm0.016}$ & ${\bf 0.409\pm0.016}$ & $\underline{{\bf 0.405\pm0.016}}$ & $\underline{{\bf 0.399\pm0.016}}$ & $\underline{{\bf 0.395\pm0.016}}$ & $0.428\pm0.016$\tabularnewline
Naval & ${\bf 0.005\pm0.001}$ & $\underline{{\bf 0.004\pm0.000}}$ & $\underline{{\bf 0.003\pm0.000}}$ & $\underline{{\bf 0.003\pm0.000}}$ & $\underline{{\bf 0.003\pm0.000}}$ & $0.006\pm0.000$\tabularnewline
WineRed & ${\bf 0.654\pm0.009}$ & $\underline{{\bf \mathbf{0.647\pm0.009}}}$ & $\underline{{\bf 0.644\pm0.009}}$ & $\underline{{\bf 0.641\pm0.009}}$ & $\underline{{\bf 0.639\pm0.009}}$ & $0.655\pm0.008$\tabularnewline
WineWhite & $0.695\pm0.003$ & $0.692\pm0.003$ & $0.690\pm0.003$ & $0.689\pm0.002$ & $0.688\pm0.003$ & $\underline{{\bf 0.655\pm0.008}}$\tabularnewline
Yacht & $0.616\pm0.055$ & $0.608\pm0.052$ & $0.597\pm0.051$ & ${\bf 0.587\pm0.054}$ & ${\bf 0.578\pm0.054}$ & $0.593\pm0.071$\tabularnewline
\hline 
\end{tabular}
\par\end{centering}
\caption{Comparing SRLD with different number of samples with SVGD on test RMSE. The results are computed over 20 trials. For SRLD, the value is set to be boldface if it has better average performance than SVGD. The value if with underline if it is significantly better than SVGD with significant level 0.05 using a matched pair t-test.}\label{tb:SRLD_SVGD_RMSE}
\end{table}

\begin{table}[h]
\scriptsize
\setlength\tabcolsep{3.pt}
\begin{centering}
\begin{tabular}{c|cccccc}
\hline 
\multirow{2}{*}{Dataset} & \multicolumn{6}{c}{Ave Test LL}\tabularnewline
 & SRLD-20 & SRLD-40 & SRLD-60 & SRLD-80 & SRLD-100 & SVGD\tabularnewline
\hline 
Boston & $\underline{{\bf-2.642\pm.088}}$ & $\underline{{\bf-2.582\pm0.084}}$ & $\underline{{\bf-2.527\pm0.612}}$ & $\underline{{\bf-2.516\pm0.062}}$ & $\underline{{\bf -2.500\pm0.054}}$ & $-4.276\pm0.217$ \tabularnewline
Concrete & $\underline{{\bf -3.084\pm0.036}}$ & $\underline{{\bf -3.061\pm0.034}}$ & $\underline{{\bf -3.050\pm0.033}}$ & $\underline{{\bf -3.040\pm0.031}}$ & $\underline{{\bf -3.034\pm0.031}}$ & $-5.500\pm0.398$\tabularnewline
Energy & $\underline{{\bf -0.580\pm0.053}}$ & $\underline{{\bf -0.536\pm0.048}}$ & $\underline{{\bf -0.522\pm0.046}}$ & $\underline{{\bf -0.504\pm0.044}}$ & $\underline{{\bf -0.476\pm0.036}}$ & $-0.781\pm0.094$\tabularnewline
Naval & $\underline{{\bf 4.033\pm0.230}}$ & $\underline{{\bf 4.100\pm0.171}}$ & $\underline{{\bf 4.140\pm0.015}}$ & $\underline{{\bf 4.167\pm0.014}}$ & $\underline{{\bf 4.186\pm0.015}}$ & $3.056\pm0.034$\tabularnewline
WineRed & $\underline{{\bf -1.008\pm0.019}}$ & $\underline{{\bf -0.990\pm0.017}}$ & $\underline{{\bf -0.982\pm0.016}}$ & $\underline{{\bf -0.974\pm0.016}}$ & $\underline{{\bf -0.970\pm0.016}}$ & $-1.040\pm0.018$\tabularnewline
WineWhite & $-1.053\pm0.004$ & $-1.049\pm0.004$ & $-1.047\pm0.004$ & $-1.044\pm0.004$ & $-1.043\pm0.004$ & $\underline{{\bf -1.040\pm0.019}}$\tabularnewline
Yacht & $\underline{{\bf -1.160\pm0.256}}$ & $\underline{{\bf -0.650\pm0.173}}$ & $\underline{{\bf -0.556\pm0.096}}$ & $\underline{{\bf -0.465\pm0.037}}$ & $\underline{{\bf -0.458\pm0.036}}$ & $-1.281\pm0.279$\tabularnewline
\hline 
\end{tabular}
\par\end{centering}
\caption{Comparing SRLD with different number of samples with SVGD on test log-likelihood. The results are computed over 20 trials. For SRLD, the value is set to be boldface if it has better average performance than SVGD. The value if with underline if it is significantly better than SVGD with significant level 0.05 using a matched pair t-test.} \label{tb:SRLD_SVGD_ll}
\end{table}

\newpage
\section{Contextual Bandit: Experiment Settings and More Background}\label{sec:contexual_bandits}

Contextual bandit is a class of online learning problems that can be viewed as a simple reinforcement learning problem without transition. For a completely understanding of contextual bandit problems, we refer the readers to the Chapter 4 of \citep{bubeck2012regret}. Here we include the main idea for completeness. In contextual bandit problems, the agent needs to find out the best action given some observed context (a.k.a the optimal policy in reinforcement learning). Formally, we define $\mathcal{S}$ as the context set and $K$ as the number of action. Then we can concretely describe the contextual bandit problems as follows: for each time-step $t=1, 2, \cdots, N$, where $N$ is some pre-defined time horizon (and can be given to the agent), the environment provides a context $s_t\in \mathcal{S}$ to the agent, then the agent should choose one action $a_t\in \{1, 2, \cdots, K\}$ based on context $s_t$. The environment will return a (stochastic) reward $r(s_t, a_t)$ to the agent based on the context $s_t$ and the action $a_t$ that similar to the reinforcement learning setting. And notice that, the agent can adjust the strategy at each time-step, so that this kinds of problems are called ``online'' learning problem.

Solving the contextual bandit problems is equivalent to find some algorithms that can minimize the pseudo-regret \citep{bubeck2012regret}, which is defined as:
\begin{equation}
    \overline{R}_N^{\mathcal{S}} = \max_{\pi:\mathcal{S}\to \{1, 2, \cdots, K\}}\mathbb{E}\left[\sum_{t=1}^N r(s_t, g(s_t)) - \sum_{t=1}^N r(s_t, a_t)\right].
    \label{eq:regret}
\end{equation}
where $\pi$ denotes the deterministic mapping from the context set $\mathcal{S}$ to actions $\{1, 2, \cdots, K\}$ (readers can view $\pi$ as a deterministic policy in reinforcement learning). 
Intuitively, this pseudo-regret measures the difference of cumulative reward between the action sequence $a_t$ and the best action sequence $\pi(s_t)$. Thus, an algorithm that can minimize the pseudo-regret \eqref{eq:regret} can also find the best $\pi$.

Posterior sampling \citep[a.k.a. Thompson sampling;][]{Thompson1933On} is one of the classical yet successful algorithms that can achieve the state-of-the-art performance in practice \citep{chapelle2011empirical}. It works by first placing an user-specified prior $\mu^0_{s, a}$ on the reward $r(s, a)$, and each turn make decision based on the posterior distribution and update it, i.e. update the posterior distribution $\mu^t_{s, a}$ with the observation $r(s_{t-1}, a_{t-1})$ at time $t-1$ where $a_{t-1}$ is selected with the posterior distribution: each time, the action is selected with the following way:
$$a_t = \mathop{\arg\max}_{a\in \{1, 2, \cdots, K\}}{\hat{r}(s_t, a)},\quad  \hat{r}(s_t, a)\sim \mu^t_{s, a}.$$
i.e., greedy select the action based on the sampled reward from the posterior, thus called ``Posterior Sampling''. Algorithm \ref{alg:ts} summarizes the whole procedure of Posterior Sampling.

\begin{algorithm}
\caption{Posterior sampling for contextual bandits}
\label{alg:ts}
\begin{algorithmic}
\STATE {\bf Input:} Prior distribution $\mu^0_{s, a}$, time horizon $N$
\FOR{time $t = 1, 2, \cdots, N$}
\STATE observe a new context $s_t\in \mathcal{S}$,
\STATE sample the reward of each action $\hat{r}(s_t, a)\sim \mu^t_{s, a}$, $a\in\{1, 2, \cdots, K\}$,
\STATE select action $a_t = \mathop{\arg\max}_{a\in\{1, 2, \cdots, K\}}{\hat{r}(s_t, a)}$ and get the reward $r(s_t, a_t)$,
\STATE update the posterior $\mu^{t+1}_{s_t, a_t}$ with $r(s_t, a_t)$.
\ENDFOR
\end{algorithmic}
\end{algorithm}

Notice that all of the reinforcement learning problems face the \emph{exploration-exploitation dilemma}, so as the contextual bandit problem. Posterior sampling trade off the exploration and exploitation with the uncertainty provided by the posterior distribution. So if the posterior uncertainty is not estimated properly, posterior sampling will perform poorly. To see this, if we over-estimate the uncertainty, we can explore too-much sub-optimal actions, while if we under-estimate the uncertainty, we can fail to find the optimal actions. Thus, it is a good benchmark for evaluating the uncertainty provided by different inference methods.

Though in principle all of the MCMC methods return the samples follow the true posterior if we can run infinite MCMC steps, in practice we can only obtain finite samples as we only have finite time to run the MCMC sampler. In this case, the auto-correlation issue can lead to the under-estimate the uncertainty, which will cause the failure on all of the reinforcement learning problems that need exploration.

Here, we test the uncertainty provided by vanilla Langevin dynamics and Self-repulsive Langevin dynamics on two of the benchmark contextual bandit problems suggested by \citep{DBLP:conf/iclr/RiquelmeTS18}, called \emph{mushroom} and \emph{wheel}. One can read \citep{DBLP:conf/iclr/RiquelmeTS18} to find the detail introduction of this two contextual bandit problems. For completeness, we include it as follows:

\textbf{Mushroom} Mushroom bandit utilizes the data from Mushroom dataset \citep{schlimmer1981mushroom}, which includes different kinds of poisonous mushroom and safe mushroom with 22 attributes that can indicate whether the mushroom is poisonous or not. \citet{pmlr-v37-blundell15} first introduced the mushroom bandit by designing the following reward function: eating a safe mushroom will give a $+5$ reward, while eating a poisonous mushroom will return a reward $+5$ and $-35$ with equal chances. The agent can also choose not to eat the mushroom, which always yield a $0$ reward. Same to \citep{DBLP:conf/iclr/RiquelmeTS18}, we use 50000 instances in this problem.

\textbf{Wheel}
To highlight the need for exploration, \citep{DBLP:conf/iclr/RiquelmeTS18} designs the wheel bandit, that can control the need of exploration with some ``exploration parameter'' $\delta \in (0, 1)$.
The context set $\mathcal{S}$ is the unit circle $\|s\|_2\leq 1$ in $\mathbb{R}^2$, and each turn the context $s_t$ is uniformly sampled from $\mathcal{S}$.
$K=5$ possible actions are provided: the first action yields a constant reward $r\sim \mathcal{N}(\mu_1, \sigma^2)$; the reward corresponding to other actions is determined by the provided context $s$:
\begin{itemize}
    \item For $s\in\mathcal{S}$ s.t. $\|s\|_2 \le \delta$, all of the four other actions return a suboptimal reward sampled from $\mathcal{N}(\mu_2, \sigma^2)$ for $\mu_2 < \mu_1$. 
    \item For $s\in\mathcal{S}$ s.t. $\|s\|_2 > \delta $, according to the quarter the context $s$ is in, one of the four actions becomes optimal. This optimal action gives a reward of $\mathcal{N}(\mu_3,\sigma^2)$ for $\mu_3\gg\mu_1$, and another three actions still yield the suboptimal reward $\mathcal{N}(\mu_2,\sigma^2)$. 
\end{itemize}
Following the setting from \citep{DBLP:conf/iclr/RiquelmeTS18}, we set $\mu_1=1.2$, $\mu_2=1.0$, and $\mu_3=50$.

When $\delta$ approaches $1$, the inner circle $\|s\|_2\leq \delta$ will dominate the unit circle and the first action becomes the optimal for most of the context. Thus, inference methods with poorly estimated uncertainty will continuously choose the suboptimal action $a_1$ for all of the contexts without exploration. This phenomenon have been confirmed in \citep{DBLP:conf/iclr/RiquelmeTS18}. In our experiments, as we want to evaluate the quality of uncertainty provided by different methods, we set $\delta=0.95$, which is pretty hard for existing inference methods as shown in \citep{DBLP:conf/iclr/RiquelmeTS18}, and use $50000$ contexts for evaluation.
\begin{figure}[h]
    \centering
    \includegraphics[width=0.3\linewidth]{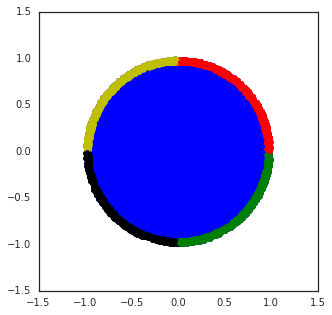}
    \caption{Visualization of the wheel bandit ($\delta=0.95$), taken from \citep{DBLP:conf/iclr/RiquelmeTS18}.}
    \label{fig:wheel}
\end{figure}

\textbf{Experiment Setup} Following \citep{DBLP:conf/iclr/RiquelmeTS18}, we use a feed-forward network with two hidden layer of 100 units and ReLU activation. We use the same step-size and thinning factor $c/\eta=100$ for vanilla Langevin dynamics and SRLD, and set $M=20$, $\alpha = 10$ on both of the mushroom and wheel bandits. The update schedule is similar to \citep{DBLP:conf/iclr/RiquelmeTS18}, and we just change the optimization step in stochastic variational inference methods into MCMC sampler step and replace the warm-up of stochastic variational inference methods in \citet{DBLP:conf/iclr/RiquelmeTS18} with the burn-in phase of the sampling. Similar to other methods in \citep{DBLP:conf/iclr/RiquelmeTS18}, we keep the initial learning rate as $10^{-1}$ for fast burn-in and the step-size for sampling is tuned on the mushroom bandit and keep the same for both the mushroom and wheel bandit. As this is an online posterior inference problem, we only use the last $20$ samples to give the prediction. Notice that, in the original implementation of \citet{DBLP:conf/iclr/RiquelmeTS18}, the authors only update a few steps with new observation after observing enough data, as the posterior will gradually converge to the true reward distribution and little update is needed after observing sufficient data. Similar to their implementation, after observing enough data, we only collect one new sample with the new observation each time. For SVGD, we use 20 particles to make the comparison fair, and also tune the step-size on the mushroom bandit.

\newpage

\section{The Detailed analysis of SRLD} \label{sec:theory_apx}

\subsection{Some additional notation}
We use $\|\cdot\|_{\infty}$ to denote the $\ell_\infty$ vector norm and define the $\mathcal{L}_{\infty}$ norm of a function $f:\mathbb{R}^{d}\to\mathbb{R}^{1}$
as $\left\Vert f\right\Vert _{\mathcal{L}_{\infty}}$. $\mathbb{D}_{\mathrm{TV}}$ denote the Total Variation distance between distribution $\rho_1, \rho_2$ respectively. Also, as $K$ is $\mathbb{R}^{d}\times\mathbb{R}^{d}\to\mathbb{R}^{1}$, we denote $\left\Vert K\right\Vert _{\mathcal{L}_{\infty},\mathcal{L}_{\infty}}=\sup_{\boldsymbol{\theta}_{1},\boldsymbol{\theta}_{2}}K(\boldsymbol{\theta}_{1},\boldsymbol{\theta}_{2})$. For simplicity, we may use $\left\Vert K\right\Vert _{\infty,\infty}$ as $\left\Vert K\right\Vert _{\mathcal{L}_{\infty},\mathcal{L}_{\infty}}$. In the appendix, we also use $\phi[\rho](\th) \coloneqq \boldsymbol{g}(\th;\rho)$, where $\boldsymbol{g}(\th;\rho)$ is defined in the main text. For the clearance, we define $\pi_{M,c/\eta}*\rho_{k} \coloneqq \rho_{k}^{M}$, $\pi_{M,c/\eta}*\tilde{\rho}_{k} \coloneqq \tilde{\rho}_{k}^{M}$ and $\pi_{M,c}*\bar{\rho}_{t} \coloneqq \bar{\rho}_{t}^{M}$, where $\rho_{k}^{M}$, $\tilde{\rho}_{k}^{M}$ and $\bar{\rho}_{t}^{M}$ are defined in main text.
\subsection{Geometric Ergodicity of SRLD}

Before we start the proof of main theorems, we give the following
theorem on the geometric ergodicity of SRLD. It is noticeable that
under this assumption, the practical dynamics follows an $(Mc/\eta+1)$-order
nonlinear autoregressive model when $k\ge Mc/\eta$:
\[
\th_{k+1}=\boldsymbol{\psi}\left(\th_{k},...,\th_{k-Mc/\eta}\right)+\sqrt{2\eta}\boldsymbol{e}_{k},
\]
where 
\[
\boldsymbol{\psi}\left(\th_{k},...,\th_{k-Mc/\eta}\right)=\th_{k}+\eta\left\{ -\nabla V(\th_{k})+\alpha\phi[\frac{1}{M}\sum_{j=1}^{M}\delta_{\th_{k-jc/\eta}}](\th_{k})\right\} .
\]
Further, if we stack the parameter by $\Th_{k}=\left[\th_{k},...,\th_{k-Mc/\eta}\right]^{\top}$
and define $\boldsymbol{\Psi}\left(\Th_{k}\right)=\left[\boldsymbol{\psi}^{\top}\left(\Th_{k}\right),\Th_{k}^{\top}\right]^{\top}$,
we have 
\[
\Th_{k+1}=\boldsymbol{\Psi}\left(\Th_{k}\right)+\sqrt{2\eta}\boldsymbol{E}_{k},
\]
where $\boldsymbol{E}_{k}=\left[\boldsymbol{e}_{k}^{\top},\boldsymbol{0}^{\top},...,\boldsymbol{0}^{\top}\right]^{\top}.$
In this way, we formulate $\Th_{k}$ as a time homogeneous Markov
Chain. In the following analysis, we only analyze the second phase
of SRLD given some initial stacked particles $\Th_{Mc/\eta-1}$.

\begin{thm}[Geometric Ergodicity] \label{thm: ergo}
Under Assumption \ref{asm: RBF} and Assumption \ref{asm: dis}, suppose we choose $\eta$ and
$\alpha$ such that 
\[
\max\left(1-2\eta a_{1}+\eta^{2}b_{1}+\frac{2\alpha\eta}{\sigma}b_{1},\frac{2\alpha\eta}{\sigma}(b_{1}+1)\right)<1,
\]
then the Markov Chain of $\Th_{k}$ is stationary, geometrically ergodic,
i.e., for any $\Th'_{0}=\Th_{Mc/\eta-1}$, we have 
\[
\mathbb{D}_{\mathrm{TV}}\left[P^{k}\left(\cdot,\Th_{0}\right),\Pi\left(\cdot\right)\right]\le Q\left(\Th_{0}\right)e^{-rk},
\]
where $r=\mathcal{O}(\eta)$ is some positive constant, $Q(\Th_{0})$ is constant related
to $\Th_{0}$, $P^{k}$ is the $k$-step Markov transition kernel and $\Pi$ is the stationary distribution.
\end{thm}
We defer the proof to Appendix \ref{sec:ergo}.
\subsection{Moment Bound}
\begin{thm} [Moment Bound] \label{thm: moment}
Under Assumption \ref{asm: dis}, suppose that we have $\mathbb{E}_{\th\sim\rho_{0}}\left\Vert \th\right\Vert ^{2}<\infty$;
and $a_{2}-\alpha\left\Vert K\right\Vert _{\infty}\left(2b_{1}+\frac{4}{\sigma}\right)>0$,
we have 
\begin{align*}
 & \sup_{k}\mathbb{E}_{\th\sim\rho_{k}}\left\Vert \th\right\Vert ^{2}\lor\sup_{k}\ \mathbb{E}_{\th\sim\tilde{\rho}_{k}}\left\Vert \th\right\Vert ^{2}\lor\sup_{t}\mathbb{E}_{\th\sim\bar{\rho}_{t}}\left\Vert \th\right\Vert ^{2}\\
\le & \mathbb{E}_{\th\sim\rho_{0}}\left\Vert \th\right\Vert ^{2}+\frac{b_{1}+1+\eta}{a_{2}-\left\Vert K\right\Vert _{\mathcal{L}_{\infty},\mathcal{L}_{\infty}}\frac{2\alpha}{\sigma}-\alpha\left\Vert K\right\Vert _{\mathcal{L}_{\infty},\mathcal{L}_{\infty}}\left(2b_{1}+\frac{2}{\sigma}\right)}.
\end{align*}
And by Lemma \ref{lem: twid}, we thus have 
\begin{align*}
 & \sup_{k}\ \mathbb{E}_{\th\sim\rho_{k}}\left\Vert \nabla V(\th)\right\Vert ^{2}\lor\sup_{k}\ \mathbb{E}_{\th\sim\tilde{\rho}_{k}}\left\Vert \nabla V(\th)\right\Vert ^{2}\lor\sup_{t}\mathbb{E}_{\th\sim\bar{\rho}_{t}}\left\Vert \nabla V(\th)\right\Vert ^{2}\\
\le & b_{1}\mathbb{E}_{\th\sim\rho_{0}}\left\Vert \th\right\Vert ^{2}+\frac{b_{1}(b_{1}+1+\eta)}{a_{2}-\left\Vert K\right\Vert _{\mathcal{L}_{\infty},\mathcal{L}_{\infty}}\frac{2\alpha}{\sigma}-\alpha\left\Vert K\right\Vert _{\mathcal{L}_{\infty},\mathcal{L}_{\infty}}\left(2b_{1}+\frac{2}{\sigma}\right)}+1
\end{align*}
\end{thm}
The proof can be found at Appendix \ref{sec:moment}.
\subsection{Technical Lemma}

\begin{defi} [$\alpha$-mixing]

For any two $\sigma$-algebras $\mathcal{A}$ and $\mathcal{B}$,
the $\alpha$-mixing coefficient is defined by 
\[
\alpha(\mathcal{A},\mathcal{B})=\underset{A\in\mathcal{A},B\in\mathcal{B}}{\sup}\left|\mathbb{P}\left(A\cap B\right)-\mathbb{P}\left(A\right)\mathbb{P}\left(B\right)\right|.
\]
Let $\left(X_{k},k\ge1\right)$ be a sequence of real random variable
defined on $\left(\Omega,\mathcal{A},\mathbb{P}\right)$. This sequence
is $\alpha$-mixing if 
\[
\alpha(n)\coloneqq\underset{k\ge1}{\sup}\ \alpha\left(\mathcal{M}_{k},\mathcal{G}_{k+n}\right)\to0,\ \mathrm{as}\ n\to\infty,
\]
where $\mathcal{M}_{j}\coloneqq\sigma\left(X_{i},i\le j\right)$ and
$\mathcal{G}_{j}\coloneqq\sigma\left(X_{i},i\ge j\right)$ for $j\ge1$.
Alternatively, as shown by Theorem 4.4 of \cite{bradley2007introduction}
\[
\alpha(n)\coloneqq\frac{1}{4}\sup\left\{ \frac{\mathrm{Cov}\left(f,g\right)}{\left\Vert f\right\Vert _{\mathcal{L}_{\infty}}\left\Vert g\right\Vert _{\mathcal{L}_{\infty}}};\ f\in\mathcal{L}_{\infty}\left(\mathcal{M}_{k}\right),\ g\in\mathcal{L}_{\infty}\left(\mathcal{G}_{k+n}\right)\right\} .
\]
\end{defi}

\begin{defi}[$\beta$-mixing]

For any two $\sigma$-algebras $\mathcal{A}$ and $\mathcal{B}$,
the $\alpha$-mixing coefficient is defined by 
\[
\beta\left(\mathcal{A},\mathcal{B}\right)\coloneqq\sup\frac{1}{2}\sum_{i=1}^{I}\sum_{j=1}^{J}\left|\mathbb{P}\left(A_{i}\cap B_{j}\right)-\mathbb{P}\left(A_{i}\right)\mathbb{P}\left(B_{j}\right)\right|,
\]
where the supremum is taken over all pairs of finite partitions $\left\{ A_{1},...,A_{I}\right\} $
and $\left\{ B_{1},...,B_{J}\right\} $ of $\Omega$ such that $A_{i}\in\mathcal{A}$
and $B_{j}\in\mathcal{B}$ for each $i$, $j$. Let $\left(X_{k},k\ge1\right)$
be a sequence of real random variable defined on $\left(\Omega,\mathcal{A},\mathbb{P}\right)$.
This sequence is $\beta$-mixing if 
\[
\beta(n)\coloneqq\underset{k\ge1}{\sup}\ \beta\left(\mathcal{M}_{k},\mathcal{G}_{k+n}\right)\to0,\ \mathrm{as}\ n\to\infty.
\]
\end{defi}

\begin{prp} [$\beta$-mixing implies $\alpha$-mixing)] \label{prop: mix}

For any two $\sigma$-algebras $\mathcal{A}$ and $\mathcal{B}$,
\[
\alpha\left(\mathcal{A},\mathcal{B}\right)\le\frac{1}{2}\beta\left(\mathcal{A},\mathcal{B}\right).
\]
\end{prp}
This proposition can be found in Equation 1.11 of \cite{bradley2005basic}.

\begin{prp} \label{prop: ergo}
A (strictly) stationary Markov Chain is geometric ergodicity if and
only if $\beta(n)\to0$ at least exponentially fast as $n\to\infty$.
\end{prp}
This proposition is Theorem 3.7 of \cite{bradley2005basic}.

\begin{lem}[Regularity Conditions] \label{lem: twid}
By Assumption \ref{asm: dis}, we have $\left\Vert \nabla V(\th)\right\Vert \le b_{1}\left(\left\Vert \th_{1}\right\Vert +1\right)$ and $\left\Vert \th-\eta\nabla V(\th)\right\Vert \le\left(1-2\eta a_{1}+\eta^{2}b_{1}\right)\left\Vert \th\right\Vert ^{2}+\eta^{2}b_{1}+2\eta b_{1}.$
\end{lem}

\begin{lem}[Properties of RBF Kernel] \label{lem: RBF}
For RBF kernel with bandwidth $\sigma$, we have $\left\Vert K\right\Vert _{\infty,\infty}\le1$ and
\begin{eqnarray*}
\left\Vert K(\th',\th_{1})-K(\th',\th_{2})\right\Vert &\le&\left\Vert e^{-(\cdot)^{2}/\sigma}\right\Vert _{\mathrm{Lip}}\left\Vert \th_{1}-\th_{2}\right\Vert _{2}
\\
\left\Vert \nabla_{\th'}K(\th',\th_{1})-\nabla_{\th'}K(\th',\th_{2})\right\Vert &\le&\left\Vert \frac{2}{\sigma}e^{-(\cdot)^{2}/\sigma}(\cdot)\right\Vert _{\mathrm{Lip}}\left\Vert \th_{1}-\th_{2}\right\Vert _{2}.
\end{eqnarray*}
\end{lem}

\begin{lem}[Properties of Stein Operator] \label{lem: stein1}

For any distribution $\rho$ such that $\mathbb{E}_{\th\sim\rho}\left\Vert \nabla V(\th)\right\Vert <\infty,$
we have
\begin{align*}
\left\Vert \phi[\rho](\cdot)\right\Vert _{\mathrm{Lip}} & \le\left\Vert e^{-(\cdot)^{2}/\sigma}\right\Vert _{\mathrm{Lip}}\mathbb{E}_{\th\sim\rho}\left\Vert \nabla V(\th)\right\Vert +\left\Vert \frac{2}{\sigma}e^{-(\cdot)^{2}/\sigma}(\cdot)\right\Vert _{\mathrm{Lip}},\\
\left\Vert \phi[\rho](\th)\right\Vert  & \le\left\Vert K\right\Vert _{\infty}\mathbb{E}_{\th'\sim\rho}\left[\left\Vert \nabla V(\th')\right\Vert +\frac{2}{\sigma}\left(\left\Vert \th'\right\Vert +\left\Vert \th\right\Vert \right)\right]\\
 & \le\left\Vert K\right\Vert _{\infty}b_{1}+\mathbb{E}_{\th'\sim\rho}\left[\left(\frac{2}{\sigma}+b_{1}\right)\left\Vert \th'\right\Vert \right]+\left\Vert \th\right\Vert .
\end{align*}
\end{lem}

\begin{lem}[Bounded Lipschitz of Stein Operator] \label{lem: stein2}
Given $\th'$, define $\bar{\phi}_{\th'}(\th)\coloneqq\phi[\delta_{\th'}](\th)=K(\th',\th)\nabla V(\th')+\nabla_{1}K(\th',\th)$. We also denote $\bar{\phi}_{\th'}(\th)=[\bar{\phi}_{\th',1}(\th),...,\bar{\phi}_{\th',d}(\th)]^{\top}$. We have
\begin{eqnarray*}
\sum_{i=1}^{d}\left\Vert \bar{\phi}_{\protect\th',i}(\protect\th)\right\Vert _{\mathrm{Lip}}^{2}&\le&2\left\Vert \nabla V(\protect\th')\right\Vert ^{2}\left\Vert e^{-\left\Vert \cdot\right\Vert ^{2}/\sigma}\right\Vert _{\mathrm{Lip}}^{2}+2d\left\Vert \frac{2}{\sigma}e^{-\left\Vert \protect\th\right\Vert ^{2}/\sigma}\theta_{1}\right\Vert _{\mathrm{Lip}}^{2}
\\
\sum_{i=d}^{d}\left\Vert \bar{\phi}_{\th',i}(\th)\right\Vert _{\mathcal{L}_{\infty}}^{2}&\le&2d\left\Vert \frac{2}{\sigma}e^{-\left\Vert \th\right\Vert ^{2}/\sigma}\theta_{1}\right\Vert _{\mathcal{L}_{\infty}}^{2}+2\left\Vert e^{-\left\Vert \cdot\right\Vert ^{2}/\sigma}\right\Vert _{\mathcal{L}_{\infty}}^{2}\left\Vert \nabla V(\th')\right\Vert ^{2}.
\end{eqnarray*}
\end{lem}

\subsection{Proof of Main Theorems}\label{sec:proof_main}
\subsubsection{Proof of Theorem \ref{thm: ergo}}\label{sec:ergo}

The proof of this theorem is by verifying the condition of Theorem
3.2 of \cite{an1996the}. Suppose $\Th=\left[\th_{1},...,\th_{MC+1}\right]$, where $C=c/\eta$,
we have 
\begin{align*}
\left\Vert \boldsymbol{\psi}\left(\Th\right)\right\Vert  & =\left\Vert \th_{1}+\eta\left\{ -\nabla V(\th_{1})+\alpha\phi[\frac{1}{M}\sum_{j=1}^{M}\delta_{\th_{1+jC}}](\th_{k})\right\} \right\Vert \\
 & =\left\Vert \th_{1}-\eta\nabla V(\th_{1})+\frac{\eta\alpha}{M}\sum_{j=1}^{M}\left[e^{-\left\Vert \th_{1+jC}-\th_{1}\right\Vert ^{2}/\sigma}\frac{2}{\sigma}\left(\th_{1}-\th_{1+jC}\right)-e^{-\left\Vert \th_{1+jC}-\th_{1}\right\Vert ^{2}/\sigma}\nabla V(\th_{1+jC})\right]\right\Vert \\
 & \le\left\Vert \th_{1}-\eta\nabla V(\th_{1})+\frac{2}{\sigma}\frac{\eta\alpha}{M}\sum_{j=1}^{M}e^{-\left\Vert \th_{1+jC}-\th_{1}\right\Vert ^{2}/\sigma}\th_{1}\right\Vert \\
 & +\left\Vert \frac{\eta\alpha}{M}\sum_{j=1}^{M}e^{-\left\Vert \th_{1+jC}-\th_{1}\right\Vert ^{2}/\sigma}\frac{2}{\sigma}\left(-\nabla V(\th_{1+jC})-\th_{1+jC}\right)\right\Vert \\
 & \le\left\Vert \th_{1}-\eta\nabla V(\th_{1})\right\Vert +\frac{2\alpha\eta}{\sigma}\left\Vert K\right\Vert _{\infty,\infty}b_{1}(1+\left\Vert \th_{1}\right\Vert )\\
 & +\frac{2\alpha\eta}{M\sigma}\sum_{j=1}^{M}\left\Vert K\right\Vert _{\infty,\infty}b_{1}\left(1+(1+\frac{1}{b_{1}})\left\Vert \th_{1+jC}\right\Vert \right)\\
 & \overset{(1)}{\le}b_{1}(1+\frac{4\alpha\eta}{\sigma}\left\Vert K\right\Vert _{\infty,\infty})+\eta^{2}b_{1}+2\eta b_{1}\\
 & +\left(1-2\eta a_{1}+\eta^{2}b_{1}+\frac{2\alpha\eta}{\sigma}\left\Vert K\right\Vert _{\infty,\infty}b_{1}\right)\left\Vert \th_{1}\right\Vert +\frac{2\alpha\eta}{\sigma}\left\Vert K\right\Vert _{\infty,\infty}(b_{1}+1)\max_{i\in[MC+1]-\{1\}}\left\Vert \th_{1+jC}\right\Vert \\
 & \le b_{1}(1+\frac{4\alpha\eta}{\sigma}\left\Vert K\right\Vert _{\infty,\infty})+\eta^{2}b_{1}+2\eta b_{1}\\
 & +\max\left(1-2\eta a_{1}+\eta^{2}b_{1}+\frac{2\alpha\eta}{\sigma}\left\Vert K\right\Vert _{\infty,\infty}b_{1},\frac{2\alpha\eta}{\sigma}\left\Vert K\right\Vert _{\infty,\infty}(b_{1}+1)\right)\max_{i\in[MC+1]}\left\Vert \th_{1+jC}\right\Vert ,
\end{align*}
where $(1)$ is by Lemma \ref{lem: twid}. Thus, given the step size
$\eta$, if we choose $\eta$, $\alpha$ such that 
\[
\max\left(1-2\eta a_{1}+\eta^{2}b_{1}+\frac{2\alpha\eta}{\sigma}\left\Vert K\right\Vert _{\infty,\infty}b_{1},\frac{2\alpha\eta}{\sigma}\left\Vert K\right\Vert _{\infty,\infty}(b_{1}+1)\right)<1,
\]
then our dynamics is geometric ergodic.

\subsubsection{Proof of Theorem \ref{thm: moment}}\label{sec:moment}
\textbf{Continuous-Time Mean Field Dynamics \eqref{eq:dy3}}
Notice that as our dynamics has two phases and the first phase can
be viewed as an special case of the second phase by setting $\alpha=0$,
here we only analysis the second phase. Define $U_{t}=\underset{s\le t}{\sup}\sqrt{\mathbb{E}\left\Vert \bar{\th}_{s}\right\Vert ^{2}}$,
and thus 
\[
\frac{\partial}{\partial t}U_{t}^{2}\le\mathbb{E}\left\langle \bar{\th}_{t},-V(\bar{\th})+\alpha\phi[\pi_{M,c}*\bar{\rho}_{t}](\bar{\th}_{t})\right\rangle \vee0.
\]
Now we bound $\mathbb{E}\left\langle \bar{\th}_{t},-V(\bar{\th})+\alpha\phi[\pi_{M,c}*\bar{\rho}_{t}](\bar{\th}_{t})\right\rangle $:

\begin{align*}
 & \mathbb{E}\left\langle \bar{\th}_{t},-V(\bar{\th}_{t})+\alpha\phi[\pi_{M,c}*\bar{\rho}_{t}](\bar{\th}_{t})\right\rangle \\
\le & b_{1}-a_{2}\mathbb{E}\left\Vert \bth_{t}\right\Vert ^{2}+\alpha\mathbb{E}\left\Vert \bth_{t}\right\Vert \left\Vert \phi[\pi_{M,c}*\bar{\rho}_{t}](\bar{\th}_{t})\right\Vert \\
\stackrel{(1)}{\le} & b_{1}-a_{2}\mathbb{E}\left\Vert \bth_{t}\right\Vert ^{2}+\alpha\left\Vert K\right\Vert _{\infty}\mathbb{E}\left\Vert \bth_{t}\right\Vert \mathbb{E}_{\th'\sim\pi_{M,c}*\bar{\rho}_{t}}\left[\left\Vert \nabla V(\th')\right\Vert +\frac{2}{\sigma}\left(\left\Vert \th'\right\Vert +\left\Vert \th_{t}\right\Vert \right)\right]\\
\le & b_{1}-a_{2}\mathbb{E}\left\Vert \bth_{t}\right\Vert ^{2}+\alpha\left\Vert K\right\Vert _{\infty}\mathbb{E}\left\Vert \bth_{t}\right\Vert \mathbb{E}_{\th'\sim\pi_{M,c}*\bar{\rho}_{t}}\left[b_{1}\left(\left\Vert \th'\right\Vert +1\right)+\frac{2}{\sigma}\left(\left\Vert \th'\right\Vert +\left\Vert \th_{t}\right\Vert \right)\right]\\
= & b_{1}-\left(a_{2}-\left\Vert K\right\Vert _{\infty}\frac{2\alpha}{\sigma}\right)\mathbb{E}\left\Vert \bth_{t}\right\Vert ^{2}+\alpha\left\Vert K\right\Vert _{\infty}\mathbb{E}\left\Vert \bth_{t}\right\Vert \mathbb{E}_{\th'\sim\pi_{M,c}*\bar{\rho}_{t}}\left(\left(b_{1}+\frac{2}{\sigma}\right)\left\Vert \th'\right\Vert +b_{1}\right)\\
\le & b_{1}-\left(a_{2}-\left\Vert K\right\Vert _{\infty}\frac{2\alpha}{\sigma}\right)U_{t}^{2}+\alpha\left\Vert K\right\Vert _{\infty}\mathbb{E}\left\Vert \bth_{t}\right\Vert \mathbb{E}_{\th'\sim\pi_{M,c}*\bar{\rho}_{t}}\left(\left(b_{1}+\frac{2}{\sigma}\right)\left\Vert \th'\right\Vert +b_{1}\right)\\
\le & b_{1}-\left(a_{2}-\left\Vert K\right\Vert _{\infty}\frac{2\alpha}{\sigma}\right)U_{t}^{2}+\alpha\left\Vert K\right\Vert _{\infty}\left(b_{1}+\frac{2}{\sigma}\right)\frac{1}{M}\sum_{j=1}^{M}U_{t}U_{t-jc}+\alpha\left\Vert K\right\Vert _{\infty}b_{1}U_{t}\\
\le & b_{1}-\left(a_{2}-\left\Vert K\right\Vert _{\infty}\frac{2\alpha}{\sigma}\right)U_{t}^{2}+\alpha\left\Vert K\right\Vert _{\infty}\left(b_{1}+\frac{2}{\sigma}\right)U_{t}^{2}+\alpha\left\Vert K\right\Vert _{\infty}b_{1}(U_{t}^{2}+1)\\
\le & \left(b_{1}+1\right)-\left(a_{2}-\left\Vert K\right\Vert _{\infty}\frac{2\alpha}{\sigma}-\alpha\left\Vert K\right\Vert _{\infty}\left(2b_{1}+\frac{2}{\sigma}\right)\right)U_{t}^{2},
\end{align*}
where $(1)$ is by \ref{lem: stein1}. By the assumption that $\lambda\coloneqq a_{2}-\left\Vert K\right\Vert _{\infty}\frac{2\alpha}{\sigma}-\alpha\left\Vert K\right\Vert _{\infty}\left(2b_{1}+\frac{2}{\sigma}\right)>0$,
we have 
\[
\frac{\partial}{\partial t}U_{t}^{2}\le\left[\left(b_{1}+1\right)-\lambda U_{t}^{2}\right]\lor0.
\]
By Gronwall's inequality, we have $U_{t}^{2}\le U_{0}^{2}+\frac{b_{1}+1}{\lambda}$. (If $\frac{\partial}{\partial t}U_t^2 = 0$, then $U_t$ fix and this bound still holds.)
Notice that in the first phase, as $\alpha=0$, we have $\lambda<a_{2}$
and thus this inequality also holds.

\textbf{Discrete-Time Mean Field Dynamics \eqref{eq:dy2}} Similarly to the analysis of the continuous-time mean field dynamics \eqref{eq:dy3}, we only give proof of the
second phase. Define $U_{k}=\underset{s\le k}{\sup}\sqrt{\mathbb{E}\left\Vert \tth_{s}\right\Vert ^{2}}$,
and thus 
\[
U_{k}^{2}-U_{k-1}^{2}\le\left[2\eta\mathbb{E}\left\langle \tth_{k-1},-\nabla V(\tilde{\th}_{k})+\alpha\phi[\pi_{M,c/\eta}*\tilde{\rho}_{k}](\tilde{\th}_{k})\right\rangle +2\eta^{2}\right]\lor0.
\]
By a similarly analysis, we have bound 
\begin{align*}
 & \mathbb{E}\left\langle \tth_{k-1},-\nabla V(\tilde{\th}_{k})+\alpha\phi[\pi_{M,c/\eta}*\tilde{\rho}_{k}](\tilde{\th}_{k})\right\rangle \\
\le & \left(b_{1}+1\right)-\lambda U_{t}^{2},
\end{align*}
where $\lambda=a_{2}-\left\Vert K\right\Vert _{\infty,\infty}\frac{2\alpha}{\sigma}-\alpha\left\Vert K\right\Vert _{\infty,\infty}\left(2b_{1}+\frac{2}{\sigma}\right)>0$.
And thus we have 
\[
U_{k}^{2}-U_{k-1}^{2}\le\left[2\eta\left[\left(b_{1}+1\right)-\lambda U_{k-1}^{2}\right]+2\eta^{2}\right]\lor0.
\]
It gives that 
\[
U_{k}^{2}\le\frac{b_{1}+1+\eta}{\lambda}+U_{0}^{2}.
\]

\textbf{Practical Dynamics \eqref{eq:srld}} The analysis of Practical Dynamics \eqref{eq:srld} is almost identical to that of the discrete-time mean field dynamics \eqref{eq:dy2} and thus is omitted here.

\subsubsection{Proof of Theorem \ref{thm: stationary} and \ref{thm: stationary2}}\label{sec:proof_stationary}
Notice that the dynamics in Theorem \ref{thm: stationary} is special
case of that in Theorem \ref{thm: stationary2} and thus we only prove
Theorem \ref{thm: stationary2} here. After some algebra, we can show
that the continuity equation of dynamics \eqref{eq:dy4}
is
\[
\partial_{t}\rho_{t}=\nabla\cdot\left(\left[-\left(D(\th)+Q(\th)\right)\nabla V(\th)+\alpha\phi[\pi_{M,c}*\rho_{t}](\th_{t})\right]\rho_{t}+\left(D(\th)+Q(\th)\right)\nabla\rho_{t}\right).
\]
Notice that the limiting distribution satisfies
\begin{align*}
0 & \overset{a.e.}{=}\nabla\cdot\left(\left[-\left(D(\th)+Q(\th)\right)\nabla V(\th)+\alpha\phi[\pi_{M,c}*\rho_{\infty}](\th_{t})\right]\rho_{\infty}+\left(D(\th)+Q(\th)\right)\nabla\rho_{\infty}\right)\\
 & =\nabla\cdot\left(\left[-\left(D(\th)+Q(\th)\right)\nabla V(\th)+\alpha\phi[\rho_{\infty}](\th_{t})\right]\rho_{\infty}+\left(D(\th)+Q(\th)\right)\nabla\rho_{\infty}\right)\\
 & =\nabla\cdot\left(\left[-\left(D(\th)+Q(\th)\right)\nabla V(\th)\right]\rho_{\infty}+\left(D(\th)+Q(\th)\right)\nabla\rho_{\infty}\right)\\
 & +\alpha\nabla\cdot\left(K*\left(\nabla\rho_{\infty}-\nabla V(\th)\rho_{\infty}\right)\rho_{\infty}\right).
\end{align*}
which implies that $\rho_{\infty}\propto\exp(-V(\th))$ is the stationary distribution. 
\subsubsection{Proof of Theorem \ref{thm: discrete}}\label{sec:proof-discretize}
In the later proof we use $c_d$ to represent the quantity
\[
\sqrt{\mathbb{E}_{\th\sim\rho_{0}}\left\Vert \th\right\Vert ^{2}+\frac{b_{1}+1+\eta}{a_{2}-\left\Vert K\right\Vert _{\infty,\infty}\frac{2\alpha}{\sigma}-\alpha\left\Vert K\right\Vert _{\infty,\infty}\left(2b_{1}+\frac{2}{\sigma}\right)}}.
\]
Recall that there are two dynamics: the continuous-time mean field dynamics \eqref{eq:dy3} and the discretized version discrete-time mean field Dynamics \eqref{eq:dy2}.
Notice that here we couple the discrete-time mean field dynamics with the continuous-time mean field system using the same initialization. Given any $T=\eta N$, for any $0\le t\le T$, define $\lt=\lfloor\frac{t}{\eta}\rfloor\eta$.
We introduce an another continuous-time interpolation dynamics: 
\begin{align*}
\hat{\th}_{t} & =\begin{cases}
-\nabla V(\hat{\th}_{\lt})+d\mathcal{B}_{t}, & t\in[0,Mc)\\
-\nabla V(\hat{\th}_{\lt})+\alpha\phi[\pi_{M,c}*\hat{\rho}_{\lt}](\hat{\th}_{\lt})+d\mathcal{B}_{t}, & t\ge Mc,
\end{cases}\\
\hat{\rho}_{t} & =\mathrm{Law}(\hat{\th}_{t}),\\
\hat{\th}_{0} & =\bar{\th}_{0}\sim\bar{\rho}_{0},
\end{align*}
Notice that here we couples this interpolation dynamics with the same
Brownian motion as that of the dynamics of $\bar{\th}_{t}$. By the
definition of $\hat{\th}_{t}$, at any $t_{k}\coloneqq k\eta$ for
some integrate $k\in[N]$, $\hat{\th}_{t_k}$ and $\tilde{\th}_{k}$
has the same distribution. Define $\bar{\rho}_{t}^{\th_{0}}=\mathrm{Law}(\bar{\th}_{t})$
conditioning on $\bar{\th}_{0}=\th_{0}$ and $\hat{\rho}_{t}^{\th_{0}}=\mathrm{Law}(\hat{\th}_{t})$
conditioning on $\hat{\th}_{0}=\th_{0}$. Followed by the argument of proving Lemma 2 in \cite{dalalyan2017theoretical}, if $k\ge\frac{Mc}{\eta}$, we have
\begin{align*}
 & \mathbb{D}_{\mathrm{KL}}\left[\bar{\rho}_{t_{k}}^{\th_{0}}\|\hat{\rho}_{t_{k}}^{\th_{0}}\right]\\
= & \frac{1}{4}\int_{0}^{t_{k}}\mathbb{E}\left\Vert -\nabla V(\hat{\th}_{\ls})+\alpha\phi[\pi_{M,c}*\hat{\rho}_{\ls}](\hat{\th}_{\ls})+\nabla V(\hat{\th}_{s})-\alpha\phi[\pi_{M,c}*\bar{\rho}_{s}](\hat{\th}_{s})\right\Vert ^{2}ds\\
= & \frac{1}{4}\sum_{j=0}^{k-1}\int_{t_{j}}^{t_{j+1}}\mathbb{E}\left\Vert -\nabla V(\hat{\th}_{t_{j}})+\alpha\phi[\pi_{M,c}*\hat{\rho}_{t_{j}}](\hat{\th}_{t_{j}})+\nabla V(\hat{\th}_{s})-\alpha\phi[\pi_{M,c}*\bar{\rho}_{s}](\hat{\th}_{s})\right\Vert ^{2}ds\\
\le & \frac{3}{4}\sum_{j=0}^{k-1}\int_{t_{j}}^{t_{j+1}}\mathbb{E}\left\Vert \nabla V(\hat{\th}_{t_{j}})-\nabla V(\hat{\th}_{s})\right\Vert ^{2}ds\\
+ & \frac{3\alpha^{2}}{4}\sum_{j=0}^{k-1}\int_{t_{j}}^{t_{j+1}}\mathbb{E}\left\Vert \phi[\pi_{M,c}*\hat{\rho}_{t_{j}}](\hat{\th}_{t_{j}})-\phi[\pi_{M,c}*\bar{\rho}_{s}](\hat{\th}_{t_{j}})\right\Vert ^{2}ds\\
\le & \frac{3\alpha^{2}}{4}\sum_{j=0}^{k-1}\int_{t_{j}}^{t_{j+1}}\mathbb{E}\left\Vert \phi[\pi_{M,c}*\bar{\rho}_{s}](\hat{\th}_{t_{j}})-\phi[\pi_{M,c}*\bar{\rho}_{s}](\hat{\th}_{s})\right\Vert ^{2}ds\\
= & I_{1}+I_{2}+I_{3}.
\end{align*}
We bound $I_{1}$, $I_{2}$ and $I_{3}$ separately.

\textbf{Bounding $I_{1}$ and $I_{3}$} By the smoothness of $\nabla V$, we have 
\[
\left\Vert \nabla V(\hat{\th}_{t_{j}})-\nabla V(\hat{\th}_{s})\right\Vert ^{2}\le b_{1}^{2}\left\Vert \hat{\th}_{t_{j}}-\hat{\th}_{s}\right\Vert ^{2}.
\]
And by Lemma \ref{lem: stein1} (Lipschitz of Stein Operator), we
know that 
\begin{align*}
 & \left\Vert \phi[\pi_{M,c}*\bar{\rho}_{s}](\th_{1})-\phi[\pi_{M,c}*\bar{\rho}_{s}](\th_{2})\right\Vert \\
\le & \left[\left\Vert e^{-(\cdot)^{2}/\sigma}\right\Vert _{\mathrm{Lip}}\mathbb{E}_{\th\sim\pi_{M,c}*\bar{\rho}_{s}}\left\Vert \nabla V(\th)\right\Vert +\left\Vert \frac{2}{\sigma}e^{-(\cdot)^{2}/\sigma}(\cdot)\right\Vert _{\mathrm{Lip}}\right]\left\Vert \th_{1}-\th_{2}\right\Vert _{2}.
\end{align*}
And by the Assumption \ref{asm: dis} and that $\bar{\rho}_{s}$ as
finite second moment, we have 
\begin{align*}
 & \left\Vert \phi[\pi_{M,c}*\bar{\rho}_{s}](\th_{1})-\phi[\pi_{M,c}*\bar{\rho}_{s}](\th_{2})\right\Vert \\
\le & Cc_{d}\left\Vert \th_{1}-\th_{2}\right\Vert _{2}.
\end{align*}
Combine the two bounds, we have 
\[
I_{1}+I_{3}\le\frac{3Cc_{d}^{2}}{4}\sum_{j=0}^{k-1}\int_{t_{j}}^{t_{j+1}}\mathbb{E}\left\Vert \hat{\th}_{t_{j}}-\hat{\th}_{s}\right\Vert ^{2}ds.
\]
Notice that $\hat{\th}_{t}=\hat{\th}_{\lt}+\left[-\nabla V(\hat{\th}_{\lt})+\alpha\phi[\pi_{M,c}*\hat{\rho}_{\lt}](\hat{\th}_{\lt})\right](t-\lt)+\int_{\lt}^{t}d\mathcal{B}_{s}$.
By It\^o's lemma, it implies that 
\begin{align*}
I_{1}+I_{3} & \le\frac{3Cc_{d}^{2}}{4}\sum_{j=0}^{k-1}\int_{t_{j}}^{t_{j+1}}\mathbb{E}\left\Vert \hat{\th}_{t_{j}}-\hat{\th}_{s}\right\Vert ^{2}ds\\
 & \le\frac{3Cc_{d}^{2}}{4}\int_{t_{j}}^{t_{j+1}}\left[\mathbb{E}\left\Vert -\nabla V(\hat{\th}_{\ls})+\alpha\phi[\pi_{M,c}*\hat{\rho}_{\ls}](\hat{\th}_{\ls})\right\Vert ^{2}(s-t_{j})^{2}+2d(s-t_{j})\right]ds\\
 & =Cc_{d}^{2}\eta^{3}\sum_{j=0}^{k-1}\mathbb{E}\left\Vert -\nabla V(\hat{\th}_{t_{j}})+\alpha\phi[\pi_{M,c}*\hat{\rho}_{t_{j}}](\hat{\th}_{t_{j}})\right\Vert ^{2}+Cc_{d}^{2}dkh^{2}.
\end{align*}
By the assumption that $\mathbb{E}\left\Vert \tilde{\th}_{t_{j}}\right\Vert $
is finite and $\tilde{\th}_{t_{j}}\overset{d}{=}\hat{\th}_{t_{j}}$,
$\mathbb{E}\left\Vert \hat{\th}_{t_{j}}\right\Vert ^{2}$ is also
finite, we have
\begin{align*}
 & \mathbb{E}\left\Vert -\nabla V(\hat{\th}_{\lt})+\alpha\phi[\pi_{M,c}*\hat{\rho}_{\lt}](\hat{\th}_{\lt})\right\Vert ^{2}\\
\le & 2\mathbb{E}\left\Vert \nabla V(\hat{\th}_{\lt})\right\Vert ^{2}+2\alpha^{2}\mathbb{E}\left\Vert \phi[\pi_{M,c}*\hat{\rho}_{\lt}](\hat{\th}_{\lt})\right\Vert ^{2}\\
\le & 4b_{1}^{2}+4b_{1}^{2}\mathbb{E}\left\Vert \hat{\th}_{\lt}\right\Vert ^{2}+2\alpha^{2}\mathbb{E}\left(\left(\frac{2}{\sigma}+b_{1}\right)\mathbb{E}_{\th'\sim\pi_{M,c}*\hat{\rho}_{\lt}}\left\Vert \th'\right\Vert +\left\Vert \th\right\Vert \right)^{2}\\
\le & c_{d}^{2}C.
\end{align*}
Thus we conclude that 
\[
I_{1}+I_{3}\le Cc_{d}^{2}\left(c_{d}^{2}k\eta^{3}+dk\eta^{2}\right).
\]

\textbf{Bounding $I_{2}$}

\begin{align*}
 & \mathbb{E}\left\Vert \phi[\pi_{M,c}*\hat{\rho}_{t_{j}}](\hat{\th}_{t_{j}})-\phi[\pi_{M,c}*\bar{\rho}_{s}](\hat{\th}_{t_{j}})\right\Vert ^{2}\\
= & \mathbb{E}\left\Vert \frac{1}{M}\sum_{l=1}^{M}\left[\phi[\hat{\rho}_{t_{j}-cl}](\hat{\th}_{t_{j}})-\phi[\bar{\rho}_{s-cl}](\hat{\th}_{t_{j}})\right]\right\Vert ^{2}\\
\le & \frac{1}{M}\sum_{l=1}^{M}\mathbb{E}\left\Vert \phi[\hat{\rho}_{t_{j}-cl}](\hat{\th}_{t_{j}})-\phi[\bar{\rho}_{s-cl}](\hat{\th}_{t_{j}})\right\Vert ^{2}\\
= & \frac{1}{M}\sum_{l=1}^{M}\mathbb{E}\left\Vert \mathbb{E}_{\th\sim\hat{\rho}_{t_{j}-cl}}\bar{\phi}_{\hat{\th}_{t_{j}}}(\th)-\mathbb{E}_{\th\sim\bar{\rho}_{s-cl}}\bar{\phi}_{\hat{\th}_{t_{j}}}(\th)\right\Vert ^{2}\\
= & \frac{1}{M}\sum_{l=1}^{M}\mathbb{E}_{\hat{\th}_{t_{j}}}\sum_{i=1}^{d}\left|\mathbb{E}_{\th\sim\hat{\rho}_{t_{j}-cl}}\bar{\phi}_{\hat{\th}_{t_{j}},i}(\th)-\mathbb{E}_{\th\sim\bar{\rho}_{s-cl}}\bar{\phi}_{\hat{\th}_{t_{j}},i}(\th)\right|^{2}\\
\le & \frac{1}{M}\sum_{l=1}^{M}\mathbb{E}_{\hat{\th}_{t_{j}}}\sum_{i=1}^{d}\left(\left\Vert \bar{\phi}_{\hat{\th}_{t_{j}},i}(\cdot)\right\Vert _{\mathcal{L}_{\infty}}\lor\left\Vert \bar{\phi}_{\hat{\th}_{t_{j}},i}(\cdot)\right\Vert _{\mathrm{Lip}}\right)^{2}\mathbb{D}_{\mathrm{BL}}^{2}\left[\hat{\rho}_{t_{j}-cl},\bar{\rho}_{s-cl}\right]
\end{align*}
By Lemma \ref{lem: stein2} and the Assumption \ref{asm: logcon} that $V$ is at most
quadratic growth and that $\hat{\rho}_{\lt}$ has finite second moment,
we have 
\begin{align*}
 & \mathbb{E}_{\hat{\th}_{t_{j}}}\sum_{i=1}^{d}\left(\left\Vert \bar{\phi}_{\hat{\th}_{t_{j}},i}(\cdot)\right\Vert _{\mathcal{L}_{\infty}}\lor\left\Vert \bar{\phi}_{\hat{\th}_{t_{j}},i}(\cdot)\right\Vert _{\mathrm{Lip}}\right)^{2}\\
= & \mathbb{E}_{\hat{\th}_{t_{j}}}\sum_{i=1}^{d}\left(\left\Vert \bar{\phi}_{\hat{\th}_{t_{j}},i}(\cdot)\right\Vert _{\mathcal{L}_{\infty}}^{2}\lor\left\Vert \bar{\phi}_{\hat{\th}_{t_{j}},i}(\cdot)\right\Vert _{\mathrm{Lip}}^{2}\right)\\
\le & \left[4d\left\Vert \frac{2}{\sigma}e^{-\left\Vert \th\right\Vert ^{2}/\sigma}\theta_{1}\right\Vert _{\mathrm{BL}}^{2}+4\left\Vert e^{-\left\Vert \cdot\right\Vert ^{2}/\sigma}\right\Vert _{\mathrm{BL}}^{2}\mathbb{E}_{\hat{\th}_{t_{j}}}\left\Vert \nabla V(\hat{\th}_{t_{j}})\right\Vert ^{2}\right]\\
\le & C(d+c_{d}^{2}).
\end{align*}
Plug in the above estimation, we have 
\begin{align*}
I_{2} & =\frac{3\alpha^{2}}{4}\sum_{j=0}^{k-1}\int_{t_{j}}^{t_{j+1}}\mathbb{E}\left\Vert \phi[\pi_{M,c}*\hat{\rho}_{t_{j}}](\hat{\th}_{t_{j}})-\phi[\pi_{M,c}*\bar{\rho}_{s}](\hat{\th}_{t_{j}})\right\Vert ^{2}ds\\
 & \le\alpha^{2}C(d+c_{d}^{2})\sum_{j=0}^{k-1}\int_{t_{j}}^{t_{j+1}}\frac{1}{M}\sum_{l=1}^{M}\mathbb{D}_{\mathrm{BL}}^{2}\left[\hat{\rho}_{t_{j}-cl},\bar{\rho}_{s-cl}\right]ds\\
 & \le\alpha^{2}C(d+c_{d}^{2})\sum_{j=0}^{k-1}\frac{1}{M}\sum_{l=1}^{M}\int_{t_{j}}^{t_{j+1}}\mathbb{D}_{\mathrm{KL}}\left[\hat{\rho}_{t_{j}-cl},\bar{\rho}_{s-cl}\right]ds,
\end{align*}
where the last inequality is due to the relation that $\mathbb{D}_{\mathrm{BL}}^{2}\overset{\mathrm{definition}}{\le}\mathbb{D}_{\mathrm{TV}}^{2}\overset{\mathrm{Pinsker's}}{\le}\mathbb{D}_{\mathrm{KL}}$. 

\textbf{Overall Bound} Combine all the estimation, we have 
\begin{align*}
\mathbb{D}_{\mathrm{KL}}\left[\bar{\rho}_{t_{k}}^{\th_{0}}\|\hat{\rho}_{t_{k}}^{\th_{0}}\right] & \le\alpha^{2}C(d+c_{d}^{2})\sum_{j=0}^{k-1}\frac{1}{M}\sum_{l=1}^{M}\int_{t_{j}}^{t_{j+1}}\mathbb{D}_{\mathrm{KL}}\left[\hat{\rho}_{t_{j}-cl},\bar{\rho}_{s-cl}\right]ds+Cc_{d}^{2}\left(c_{d}^{2}k\eta^{3}+dk\eta^{2}\right)\\
 & =\alpha^{2}C(d+c_{d}^{2})\sum_{j=0}^{k-1}\frac{1}{M}\sum_{l=1}^{M}\int_{0}^{\eta}\mathbb{D}_{\mathrm{KL}}\left[\hat{\rho}_{t_{\left(\frac{j\eta-cl}{\eta}\right)}},\bar{\rho}_{t_{\left(\frac{j\eta-cl}{\eta}\right)}+s}\right]ds+Cc_{d}^{2}\left(c_{d}^{2}k\eta^{3}+dk\eta^{2}\right)
\end{align*}
Similar, if $k\le \frac{Mc}{\eta}-1$, we have 
\begin{align*}
 & \mathbb{D}_{\mathrm{KL}}\left[\bar{\rho}_{t_{k}}^{\th_{0}}\|\hat{\rho}_{t_{k}}^{\th_{0}}\right]\\
= & \frac{1}{4}\int_{0}^{t_{k}}\mathbb{E}\left\Vert \nabla V(\hat{\th}_{\ls})-\nabla V(\hat{\th}_{s})\right\Vert ^{2}ds\\
\le & \frac{b_{1}^{2}}{4}\sum_{j=0}^{k-1}\int_{t_{j}}^{t_{j+1}}\mathbb{E}\left\Vert \hat{\th}_{t_{j}}-\hat{\th}_{s}\right\Vert ^{2}ds\\
\text{\ensuremath{\le}} & \frac{b_{1}^{2}\eta^{3}}{12}\sum_{j=0}^{k-1}\mathbb{E}\left\Vert \nabla V(\hat{\th}_{t_{j}})\right\Vert ^{2}+\frac{dkb_{1}^{2}\eta^{2}}{4}\\
\le & \frac{b_{1}^{2}\eta^{3}kc_{d}^{2}}{12}+\frac{dkb_{1}^{2}\eta^{2}}{4}.
\end{align*}
Define 
\[
u_{k}=\underset{s\in\left[t_{k},t_{k+1}\right]}{\sup}\ \mathbb{D}_{\mathrm{KL}}\left[\bar{\rho}_{\ls}^{\th_{0}}\|\hat{\rho}_{s}^{\th_{0}}\right],
\]
 and $U_{k}=\underset{l\in\{0,...,k\}}{\max}\ u_{l}$. We conclude
that for $k\ge\frac{Mc}{\eta}$, for any $k'\le k$, 
\begin{align*}
u_{k'} & \le\alpha^{2}C(d+c_{d}^{2})\sum_{j=0}^{k-1}\frac{1}{M}\sum_{l=1}^{M}\int_{0}^{h}\mathbb{D}_{\mathrm{KL}}\left[\hat{\rho}_{t_{\left(\frac{j\eta-cl}{\eta}\right)}},\bar{\rho}_{t_{\left(\frac{j\eta-cl}{\eta}\right)}+s}\right]ds+Cc_{d}^{2}\left(c_{d}^{2}k\eta^{3}+dk\eta^{2}\right)\\
 & \le\alpha^{2}C(d+c_{d}^{2})\sum_{j=0}^{k-1}\frac{1}{M}\sum_{l=1}^{M}\eta u_{\left(\frac{j\eta-cl}{\eta}\right)}+Cc_{d}^{2}\left(c_{d}^{2}k\eta^{3}+dk\eta^{2}\right)\\
 & \le\alpha^{2}C(d+c_{d}^{2})\eta\sum_{j=0}^{k-1}U_{j}+Cc_{d}^{2}\left(c_{d}^{2}k\eta^{3}+dk\eta^{2}\right).
\end{align*}
For $k<\frac{Mc}{\eta}$, which is a simpler case, we have
\[
U_{k}\le C\left(\eta^{3}kc_{d}^{2}+dk\eta^{2}\right)<CMc\left(\eta c_{d}^{2}+d\right)\eta.
\]
We bound the case when $k\ge\frac{Mc}{\eta}$,
\[
U_{k}\le\alpha^{2}C(d+c_{d}^{2})\eta\sum_{j=0}^{k-1}U_{j}+Cc_{d}^{2}\left(c_{d}^{2}k\eta^{3}+dk\eta^{2}\right).
\]
If we take $\eta$ sufficiently small, such that $c_{d}^{2}k\eta^{3}\le dk\eta^{2}$,
we have 
\begin{align*}
U_{k} & \le\alpha^{2}C(d+c_{d}^{2})\eta\sum_{j=0}^{k-1}U_{j}+2Cc_{d}^{2}dk\eta^{2}\\
 & \le\alpha^{2}C(d+c_{d}^{2})\eta\sum_{j=0}^{k-1}\left(U_{j}+\eta\right).
\end{align*}
Define $\eta'=\alpha^{2}C(d+c_{d}^{2})\eta$ and we can choose $\eta$
small enough such that $\eta'<1/2$ and $\eta<1/2$. Without loss
of generality, we also assume $\eta'\ge\eta$ and thus we have 
\[
U_{k}\le\eta'\sum_{j=0}^{k-1}\left(U_{j}+\eta'\right).
\]
Also we assume $U_{k}\ge\eta'$, otherwise we conclude that $U_{k}<\eta'$.
We thus have $U_{k}\le q\sum_{j=0}^{k-1}U_{j}$, where $q=2\eta'$.
Suppose that $U_{\frac{Mc}{\eta}-1}=x\le CMc\left(\eta c_{d}^{2}+d\right)\eta$
and some algebra (which reduces to Pascal's triangle) shows that
\[
U_{k}\le xq(1+q)^{k-\frac{Mc}{\eta}}.
\]
We conclude that $U_{k}\le xq(1+q)^{k-1}$. Notice that $q=2\alpha^{2}C(d+c_{d}^{2})\eta.$
Thus for any $k\ge Mc/\eta$,
\begin{align*}
U_{k} & \le xq(1+q)^{k-\frac{Mc}{\eta}}\\
 & =xq(1+q)^{\left(k\eta-Mc\right)/\eta}\\
 & =xq(1+q)^{2\alpha^{2}C(d+c_{d}^{2})\left(k\eta-Mc\right)/q}\\
 & \le x2\alpha^{2}C(d+c_{d}^{2})e^{2\alpha^{2}C(d+c_{d}^{2})\left(k\eta-Mc\right)}\eta\\
 & \le CMc\alpha^{2}\left(\eta c_{d}^{2}+d\right)(d+c_{d}^{2})e^{2\alpha^{2}C(d+c_{d}^{2})\left(k\eta-Mc\right)}\eta^{2},
\end{align*}
for sufficiently small $\eta$. Combine the above two estimations,
we have 
\[
U_{k}\le\begin{cases}
C\left(\eta^{3}kc_{d}^{2}+dk\eta^{2}+\eta\right) & k\le Mc/\eta-1\\
CMc\alpha^{2}\left(\eta c_{d}^{2}+d\right)(d+c_{d}^{2})e^{2\alpha^{2}C(d+c_{d}^{2})\left(k\eta-Mc\right)}\eta^{2}+C\eta & k\ge Mc/\eta
\end{cases}.
\]
Notice that now we have $U_{k}=\underset{l\in\{0,...,k\}}{\max}\underset{s\in\left[0,\eta\right]}{\sup}\ \mathbb{D}_{\mathrm{KL}}\left[\bar{\rho}_{l\eta+s}^{\th_{0}}\|\tilde{\rho}_{l\eta}^{\th_{0}}\right]$,
which is a function of $\th_{0}$. We then bound $\bar{U}_{k}=\underset{l\in\{0,...,k\}}{\max}\underset{s\in\left[0,\eta\right]}{\sup}\ \mathbb{D}_{\mathrm{KL}}\left[\bar{\rho}_{l\eta+s}\|\tilde{\rho}_{l\eta}\right]$.
Notice that the KL divergence has the following variational representation:
\[
\mathbb{D}_{\mathrm{KL}}[\rho_{1}\|\rho_{2}]=\sup_{f}\left[\mathbb{E}_{\rho_{1}}f-\mathbb{E}_{\rho_{2}}e^{f}\right],
\]
where the $f$ is chosen in the set that $\mathbb{E}_{\rho_{1}}f$
and $\mathbb{E}_{\rho_{2}}e^{f}$ exist. And thus we have 
\begin{align*}
\mathbb{D}_{\mathrm{KL}}[\bar{\rho}_{l\eta+s}\|\tilde{\rho}_{l\eta}] & =\sup_{f}\left[\mathbb{E}_{\th_{0}\sim\rho_{0}}\left(\mathbb{E}_{\bar{\rho}_{l\eta+s}^{\th_{0}}}f-\mathbb{E}_{\tilde{\rho}_{l\eta}^{\th_{0}}}e^{f}\right)\right]\\
 & \le\mathbb{E}_{\th_{0}\sim\rho_{0}}\sup_{f}\left[\left(\mathbb{E}_{\bar{\rho}_{l\eta+s}^{\th_{0}}}f-\mathbb{E}_{\tilde{\rho}_{l\eta}^{\th_{0}}}e^{f}\right)\right].
\end{align*}
And thus $\bar{U}_{k}\le U_{k}$. Also the inequality that 
\[
\bar{U}_{k}=\underset{l\in\{0,...,k\}}{\max}\underset{s\in\left[0,\eta\right]}{\sup}\ \mathbb{D}_{\mathrm{KL}}\left[\bar{\rho}_{l\eta+s}\|\tilde{\rho}_{l\eta}\right]\ge\underset{l\in\{0,...,k\}}{\max}\ \mathbb{D}_{\mathrm{KL}}\left[\bar{\rho}_{l\eta}\|\tilde{\rho}_{l\eta}\right]
\]
holds naturally by definition. We complete the proof.

\subsubsection{Proof of Theorem \ref{thm: particle}}\label{sec:proof-particle}
The constant $h_1$ is defined as
\[
h_{1} =\left\Vert \frac{2}{\sigma}e^{-\left\Vert \protect\th\right\Vert ^{2}/\sigma}\theta_{1}\right\Vert _{\mathrm{BL}}^{2}\lor\left\Vert e^{-\left\Vert \cdot\right\Vert ^{2}/\sigma}\right\Vert _{\mathrm{BL}}^{2}\lor\left\Vert \frac{2}{\sigma}e^{-(\cdot)^{2}/\sigma}(\cdot)\right\Vert _{\mathrm{Lip}}
\]
Now we start the proof. We couple the process of $\th_{k}$ and $\tilde{\th}_{k}$
by the same gaussian noise $\boldsymbol{e}_{k}$ in every iteration
and same initialization $\tilde{\th}_{0}=\th_{0}$. For $k\le Mc/\eta-1$,
$\mathbb{E}\left\Vert \boldsymbol{\theta}_{k}-\tilde{\th}_{k}\right\Vert ^{2}=0$
and for $k\ge Mc/\eta$ we have the following inequality,
\begin{align*}
 & \mathbb{E}\left\Vert \boldsymbol{\theta}_{k+1}-\tilde{\th}_{k+1}\right\Vert ^{2}-\mathbb{E}\left\Vert \boldsymbol{\theta}_{k}-\tilde{\th}_{k}\right\Vert ^{2}\\
= & 2\eta\mathbb{E}\left\langle \boldsymbol{\theta}_{k}-\tilde{\th}_{k},-\nabla V(\th_{k})+\nabla V(\tilde{\th}_{k})\right\rangle \\
+ & 2\eta\alpha\mathbb{E}\left\langle \boldsymbol{\theta}_{k}-\tilde{\th}_{k},\phi[\frac{1}{M}\sum_{j=1}^{M}\delta_{\th_{k-jc/\eta}}](\th_{k})-\phi[\pi_{M,c/\eta}*\tilde{\rho}_{k}](\tilde{\th}_{k})\right\rangle \\
+ & \eta^{2}\mathbb{E}\left\Vert -\nabla V(\th_{k})+\alpha\phi[\frac{1}{M}\sum_{j=1}^{M}\delta_{\th_{k-jc/\eta}}](\th_{k})+\nabla V(\tilde{\th}_{k})-\alpha\phi[\pi_{M,c/\eta}*\tilde{\rho}_{k}](\tilde{\th}_{k})\right\Vert ^{2}.
\end{align*}
By the log-concavity, we have 
\begin{align*}
 & \mathbb{E}\left\langle \boldsymbol{\theta}_{k}-\tilde{\th}_{k},-\nabla V(\th_{k})+\nabla V(\tilde{\th}_{k})\right\rangle \\
\le & -L\mathbb{E}\left\Vert \boldsymbol{\theta}_{k}-\tilde{\th}_{k}\right\Vert ^{2},
\end{align*}
for some positive constant $L$. And also, as $\eta$ is small, the
last term on the right side of the equation is small term. Thus our
main target is to bound the second term. We decompose the second term
on the left side of the equation by 
\begin{align*}
 & \mathbb{E}\left\langle \boldsymbol{\theta}_{k}-\tilde{\th}_{k},\phi[\frac{1}{M}\sum_{j=1}^{M}\delta_{\th_{k-jc}}](\th_{k})-\phi[\pi_{M,c/\eta}*\tilde{\rho}_{k}](\tilde{\th}_{k})\right\rangle \\
= & \mathbb{E}\left\langle \boldsymbol{\theta}_{k}-\tilde{\th}_{k},\phi[\frac{1}{M}\sum_{j=1}^{M}\delta_{\th_{k-jc/\eta}}](\th_{k})-\phi[\pi_{M,c/\eta}*\rho_{k}](\th_{k})\right\rangle \\
+ & \mathbb{E}\left\langle \boldsymbol{\theta}_{k}-\tilde{\th}_{k},\phi[\pi_{M,c/\eta}*\rho_{k}](\th_{k})-\phi[\pi_{M,c/\eta}*\tilde{\rho}_{k}](\th_{k})\right\rangle \\
+ & \mathbb{E}\left\langle \boldsymbol{\theta}_{k}-\tilde{\th}_{k},\phi[\pi_{M,c/\eta}*\tilde{\rho}_{k}](\th_{k})-\phi[\pi_{M,c/\eta}*\tilde{\rho}_{k}](\tilde{\th}_{k})\right\rangle \\
= & I_{1}+I_{2}+I_{3}.
\end{align*}
We bound $I_{1}$, $I_{2}$ and $I_{3}$ independently.

\textbf{Bounding $I_{1}$}

By Holder's inequality, 
\begin{align*}
I_{1} & \le\mathbb{E}\left[\left\Vert \boldsymbol{\theta}_{k}-\tilde{\th}_{k}\right\Vert \left\Vert \phi[\frac{1}{M}\sum_{j=1}^{M}\delta_{\th_{k-jc/\eta}}](\th_{k})-\phi[\pi_{M,c/\eta}*\rho_{k}](\th_{k})\right\Vert \right]\\
 & \le\sqrt{\mathbb{E}\left\Vert \boldsymbol{\theta}_{k}-\tilde{\th}_{k}\right\Vert ^{2}}\sqrt{\mathbb{E}\left\Vert \phi[\frac{1}{M}\sum_{j=1}^{M}\delta_{\th_{k-jc/\eta}}](\th_{k})-\phi[\pi_{M,c/\eta}*\rho_{k}](\th_{k})\right\Vert ^{2}}.
\end{align*}
We bound the second term on the right side of the inequality. Define
\[
a_{2}=\underset{k}{\sup}\ \frac{\mathbb{E}\left\Vert \phi[\frac{1}{M}\sum_{j=1}^{M}\delta_{\th_{k-jc/\eta}}](\th_{k})-\phi[\pi_{M,c/\eta}*\rho_{k}](\th_{k})\right\Vert ^{2}}{\underset{\left\Vert \th\right\Vert \le B}{\sup}\mathbb{E}\left\Vert \phi[\frac{1}{M}\sum_{j=1}^{M}\delta_{\th_{k-jc/\eta}}](\th)-\phi[\pi_{M,c/\eta}*\rho_{k}](\th)\right\Vert ^{2}}
\]
and by the regularity assumption we know that $a_2<\infty$.
Define $\phi[\frac{1}{M}\sum_{j=1}^{M}\delta_{\th_{k-jc/\eta}}](\th)-\phi[\pi_{M,c/\eta}*\rho_{k}](\th)=\phi^{*}[\frac{1}{M}\sum_{j=1}^{M}\delta_{\th_{k-jc/\eta}}]$
and since the stein operator is linear functional of the distribution,
we have 
\[
\mathbb{E}\phi^{*}[\frac{1}{M}\sum_{j=1}^{M}\delta_{\th_{k-jc/\eta}}](\th)=0,
\]
given any $\th$. By Theorem \ref{thm: ergo} that $\Th_{k}$ is geometric
ergodicity and thus is $\beta$-mixing with exponentially fast decay
rate by Proposition \ref{prop: ergo}. And by Proposition \ref{prop: mix},
we know that $\Th_{k}$ is also $\alpha$-mixing with exponentially
fast decay rate. We have the following estimation 
\begin{align*}
 & \mathbb{E}\left\Vert \phi[\frac{1}{M}\sum_{j=1}^{M}\delta_{\th_{k-jc/\eta}}](\th_{k})-\phi[\pi_{M,c/\eta}*\rho_{k}](\th_{k})\right\Vert ^{2}\\
\le & a_{2}\underset{\left\Vert \th\right\Vert \le B}{\sup}\mathbb{E}\left\Vert \phi^{*}[\frac{1}{M}\sum_{j=1}^{M}\delta_{\th_{k-jc/\eta}}](\th)\right\Vert ^{2}\\
\le & \frac{a_{2}}{M^{2}}\underset{\left\Vert \th\right\Vert \le B}{\sup}\mathbb{E}\sum_{k=1}^{M}\left\Vert \phi^{*}[\delta_{\th_{t-kc/\eta}}](\th)\right\Vert ^{2}\\
+ & \frac{a_{2}}{M^{2}}\underset{\left\Vert \th\right\Vert \le B}{\sup}\mathbb{E}\sum_{k\neq j}\left\langle \phi^{*}[\delta_{\th_{t-kc/\eta}}](\th),\phi^{*}[\delta_{\th_{t-jc/\eta}}](\th)\right\rangle \\
\le & \frac{Ca_{2}}{M}\left[\frac{e^{-rc}\left(1-e^{-rMc}\right)}{1-e^{rc}}+1\right],
\end{align*}
for some positive constant $r$ that characterize the decay rate of
$\alpha$ mixing. Notice that here $\eta$ is canceled because the decay rate of mixing is $\mathcal{O}(\eta)$ (on the power of exponential) and $c/\eta=\mathcal{O}(\eta^{-1})$. Combine this two estimations, we have 
\[
I_{1}\le\sqrt{\mathbb{E}\left\Vert \boldsymbol{\theta}_{k}-\tilde{\th}_{k}\right\Vert ^{2}}\sqrt{\frac{a_{2}C}{M}\left[\frac{e^{-rc}\left(1-e^{-rMc}\right)}{1-e^{rc}}+1\right]}.
\]
\textbf{Bounding $I_{2}$}
By Holder's inequality, we have 
\begin{align*}
I_{2} & \le\sqrt{\mathbb{E}\left\Vert \boldsymbol{\theta}_{k}-\tilde{\th}_{k}\right\Vert ^{2}}\sqrt{\mathbb{E}\left\Vert \phi[\pi_{M,c/\eta}*\rho_{k}](\th_{k})-\phi[\pi_{M,c/\eta}*\tilde{\rho}_{k}](\th_{k})\right\Vert ^{2}}.
\end{align*}
We bound the second term in the right side of the inequality. 
\begin{align*}
 & \mathbb{E}\left\Vert \phi[\pi_{M,c/\eta}*\rho_{k}](\th_{k})-\phi[\pi_{M,c/\eta}*\tilde{\rho}_{k}](\th_{k})\right\Vert ^{2}\\
= & \mathbb{E}\left\Vert \frac{1}{M}\sum_{j=1}^{M}\left[\phi[\rho_{k-jc/\eta}](\th_{k})-\phi[\tilde{\rho}_{k-jc/\eta}](\th_{k})\right]\right\Vert ^{2}\\
\le & \frac{1}{M}\sum_{j=1}^{M}\mathbb{E}\left\Vert \phi[\rho_{k-jc/\eta}](\th_{k})-\phi[\tilde{\rho}_{k-jc/\eta}](\th_{k})\right\Vert ^{2}\\
= & \frac{1}{M}\sum_{j=1}^{M}\mathbb{E}_{\th_{k}}\left\Vert \mathbb{E}_{\th\sim\rho_{k-jc/\eta}}\bar{\phi}_{\th_{k}}(\th)-\mathbb{E}_{\th\sim\tilde{\rho}_{k-jc/\eta}}\bar{\phi}_{\th_{k}}(\th)\right\Vert ^{2}\\
= & \frac{1}{M}\sum_{j=1}^{M}\mathbb{E}_{\th_{k}}\sum_{i=1}^{d}\left|\mathbb{E}_{\th\sim\rho_{k-jc/\eta}}\bar{\phi}_{\th_{k},i}(\th)-\mathbb{E}_{\th\sim\tilde{\rho}_{k-jc/\eta}}\bar{\phi}_{\th_{k},i}(\th)\right|^{2}\\
\le & \frac{1}{M}\sum_{j=1}^{M}\mathbb{E}_{\th_{k}}\sum_{i=1}^{d}\left(\left\Vert \bar{\phi}_{\th_{k},i}(\cdot)\right\Vert _{\mathcal{L}_{\infty}}\lor\left\Vert \bar{\phi}_{\th_{k},i}(\cdot)\right\Vert _{\mathrm{Lip}}\right)^{2}\mathbb{D}_{\mathrm{BL}}^{2}\left[\rho_{k-jc/\eta},\tilde{\rho}_{k-jc/\eta}\right].
\end{align*}
By Lemma \ref{lem: stein2}, we have 
\begin{align*}
 & \sum_{i=1}^{d}\left(\left\Vert \bar{\phi}_{\th_{k},i}(\cdot)\right\Vert _{\mathcal{L}_{\infty}}\lor\left\Vert \bar{\phi}_{\th_{k},i}(\cdot)\right\Vert _{\mathrm{Lip}}\right)^{2}\\
= & \sum_{i=1}^{d}\left(\left\Vert \bar{\phi}_{\th_{k},i}(\cdot)\right\Vert _{\mathcal{L}_{\infty}}^{2}\lor\left\Vert \bar{\phi}_{\th_{k},i}(\cdot)\right\Vert _{\mathrm{Lip}}^{2}\right)\\
\le & \left[4d\left\Vert \frac{2}{\sigma}e^{-\left\Vert \th\right\Vert ^{2}/\sigma}\theta_{1}\right\Vert _{\mathrm{BL}}^{2}+4\left\Vert e^{-\left\Vert \cdot\right\Vert ^{2}/\sigma}\right\Vert _{\mathrm{BL}}^{2}\left\Vert \nabla V(\th_{k})\right\Vert ^{2}\right].
\end{align*}
Plug in the above estimation and by the relation that $\mathbb{D}_{\mathrm{BL}}\le\mathbb{W}_{1}\le\mathbb{W}_{2}$,
we have
\begin{align*}
 & \mathbb{E}\left\Vert \phi[\pi_{M,c}*\rho_{k}](\th_{k})-\phi[\pi_{M,c}*\tilde{\rho}_{k}](\th_{k})\right\Vert ^{2}\\
\le & \left[4d\left\Vert \frac{2}{\sigma}e^{-\left\Vert \th\right\Vert ^{2}/\sigma}\theta_{1}\right\Vert _{\mathrm{BL}}^{2}+4\left\Vert e^{-\left\Vert \cdot\right\Vert ^{2}/\sigma}\right\Vert _{\mathrm{BL}}^{2}\mathbb{E}_{\th_{k}}\left\Vert \nabla V(\th_{k})\right\Vert ^{2}\right]\frac{1}{M}\sum_{j=1}^{M}\mathbb{D}_{\mathrm{BL}}^{2}\left[\rho_{k-cj},\tilde{\rho_{k-cj}}\right]\\
\le & \left[4d\left\Vert \frac{2}{\sigma}e^{-\left\Vert \th\right\Vert ^{2}/\sigma}\theta_{1}\right\Vert _{\mathrm{BL}}^{2}+4\left\Vert e^{-\left\Vert \cdot\right\Vert ^{2}/\sigma}\right\Vert _{\mathrm{BL}}^{2}\mathbb{E}_{\th_{k}}\left\Vert \nabla V(\th_{k})\right\Vert ^{2}\right]\frac{1}{M}\sum_{j=1}^{M}\mathbb{W}_{2}^{2}\left[\rho_{k-cj},\tilde{\rho_{k-cj}}\right].
\end{align*}
And combined all the estimation and by the definition of Wasserstein-distance,
we conclude that 
\begin{align*}
I_{2} & \le\sqrt{4d\left\Vert \frac{2}{\sigma}e^{-\left\Vert \th\right\Vert ^{2}/\sigma}\theta_{1}\right\Vert _{\mathrm{BL}}^{2}+4\left\Vert e^{-\left\Vert \cdot\right\Vert ^{2}/\sigma}\right\Vert _{\mathrm{BL}}^{2}\mathbb{E}_{\th_{k}}\left\Vert \nabla V(\th_{k})\right\Vert ^{2}}\sqrt{\frac{1}{M}\sum_{j=1}^{M}\mathbb{W}_{2}^{2}\left[\rho_{k-cj},\tilde{\rho}_{k-cj}\right]}\\
 & \le\sqrt{4d\left\Vert \frac{2}{\sigma}e^{-\left\Vert \th\right\Vert ^{2}/\sigma}\theta_{1}\right\Vert _{\mathrm{BL}}^{2}+4\left\Vert e^{-\left\Vert \cdot\right\Vert ^{2}/\sigma}\right\Vert _{\mathrm{BL}}^{2}\mathbb{E}_{\th_{k}}\left\Vert \nabla V(\th_{k})\right\Vert ^{2}}\sqrt{\frac{1}{M}\sum_{j=1}^{M}\mathbb{E}\left\Vert \th_{k-cj}-\tilde{\th}_{k-cj}\right\Vert ^{2}}.
\end{align*}

\textbf{Bounding $I_{3}$}

By Holder's inequality, 
\begin{align*}
I_{3} & \le\sqrt{\mathbb{E}\left\Vert \boldsymbol{\theta}_{k}-\tilde{\th}_{k}\right\Vert ^{2}}\sqrt{\mathbb{E}\left\Vert \phi[\pi_{M,c/\eta}*\tilde{\rho}_{k}](\th_{k})-\phi[\pi_{M,c/\eta}*\tilde{\rho}_{k}](\tilde{\th}_{k})\right\Vert ^{2}}.
\end{align*}
We bound the last term on the right side of the inequality. By assumption
and Lemma \ref{lem: stein1}, we have
\begin{align*}
 & \mathbb{E}\left\Vert \phi[\pi_{M,c/\eta}*\tilde{\rho}_{k}](\th_{k})-\phi[\pi_{M,c/\eta}*\tilde{\rho}_{k}](\tilde{\th}_{k})\right\Vert ^{2}\\
\le & \left[\left\Vert e^{-(\cdot)^{2}/\sigma}\right\Vert _{\mathrm{Lip}}\mathbb{E}_{\th\sim\tilde{\rho}_{k}}\left\Vert \nabla V(\th)\right\Vert +\left\Vert \frac{2}{\sigma}e^{-(\cdot)^{2}/\sigma}(\cdot)\right\Vert _{\mathrm{Lip}}\right]^{2}\mathbb{E}\left\Vert \th_{k}-\tilde{\th}_{k}\right\Vert ^{2}.
\end{align*}
And combine the estimation, we have 
\begin{align*}
I_{3} & \le\left[\left\Vert e^{-(\cdot)^{2}/\sigma}\right\Vert _{\mathrm{Lip}}\mathbb{E}_{\th\sim\tilde{\rho}_{k}}\left\Vert \nabla V(\th)\right\Vert +\left\Vert \frac{2}{\sigma}e^{-(\cdot)^{2}/\sigma}(\cdot)\right\Vert _{\mathrm{Lip}}\right]\mathbb{E}\left\Vert \th_{k}-\tilde{\th}_{k}\right\Vert ^{2}.
\end{align*}

\textbf{Overall Bound} Combine all the results, we have the following bound: for $k\ge Mc$,
\begin{align*}
 & \mathbb{E}\left\Vert \boldsymbol{\theta}_{k+1}-\tilde{\th}_{k+1}\right\Vert ^{2}-\mathbb{E}\left\Vert \boldsymbol{\theta}_{k}-\tilde{\th}_{k}\right\Vert ^{2}\\
\le & -2\eta L\mathbb{E}\left\Vert \boldsymbol{\theta}_{k}-\tilde{\th}_{k}\right\Vert ^{2}\\
+ & 2\eta\alpha\sqrt{\mathbb{E}\left\Vert \boldsymbol{\theta}_{k}-\tilde{\th}_{k}\right\Vert ^{2}}\frac{c_{1}}{\sqrt{M}}\\
+ & 2\text{\ensuremath{\eta\alpha}}c_{2}\sqrt{\frac{1}{M}\sum_{j=1}^{M}\mathbb{E}\left\Vert \th_{k-jc/\eta}-\tilde{\th}_{k-jc/\eta}\right\Vert ^{2}\mathbb{E}\left\Vert \boldsymbol{\theta}_{k}-\tilde{\th}_{k}\right\Vert ^{2}}\\
+ & 2\eta\alpha c_{3}\mathbb{E}\left\Vert \th_{k}-\tilde{\th}_{k}\right\Vert ^{2}\\
+ & \eta^{2}c_{4},
\end{align*}
where 
\[
c_{1}=\sqrt{a_{2}C\left[\frac{e^{-rc}\left(1-e^{-rMc}\right)}{1-e^{rc}}+1\right]},
\]
\[
c_{2}=\sqrt{4d\left\Vert \frac{2}{\sigma}e^{-\left\Vert \th\right\Vert ^{2}/\sigma}\theta_{1}\right\Vert _{\mathrm{BL}}^{2}+4\left\Vert e^{-\left\Vert \cdot\right\Vert ^{2}/\sigma}\right\Vert _{\mathrm{BL}}^{2}\underset{k}{\sup}\ \mathbb{E}_{\th_{k}}\left\Vert \nabla V(\th_{k})\right\Vert ^{2}},
\]
\[
c_{3}=\left[\left\Vert e^{-(\cdot)^{2}/\sigma}\right\Vert _{\mathrm{Lip}}\underset{k}{\sup}\ \mathbb{E}_{\th\sim\tilde{\rho}_{k}}\left\Vert \nabla V(\th)\right\Vert +\left\Vert \frac{2}{\sigma}e^{-(\cdot)^{2}/\sigma}(\cdot)\right\Vert _{\mathrm{Lip}}\right],
\]
and 
\[
c_{4}=\underset{k\ge Mc/\eta}{\sup}\ \mathbb{E}\left\Vert \nabla V(\th_{k})+\alpha\phi[\frac{1}{M}\sum_{j=1}^{M}\delta_{\th_{k-jc/\eta}}](\th_{k})-\nabla V(\tilde{\th}_{k})-\alpha\phi[\pi_{M,c/\eta}*\tilde{\rho}_{k}](\tilde{\th}_{k})\right\Vert ^{2}.
\]
Define $u_{k}=\sqrt{\mathbb{E}\left\Vert \boldsymbol{\theta}_{k}-\tilde{\th}_{k}\right\Vert ^{2}}$
and $U_{k}=\underset{l\in[k]}{\sup}\ u_{l}^ {}$, we have 
\[
U_{k+1}^{2}\le qU_{k}^{2}+\frac{2\eta\alpha c_{1}}{\sqrt{M}}U_{k}+\eta^{2}c_{4},
\]
where $q=(1-2\eta(L-\alpha c_{2}-\alpha c_{3}))$. By the assumption that $\alpha \le L/(c_2+c_3)$, $q<1$. Now we prove the
bound of $U_{k}$ by induction. We take the hypothesis that $U_{k}^{2}\le\frac{\left(\frac{2\eta\alpha c_{1}}{\sqrt{M}}+(1-q)\eta\left(c_{4}+\frac{1}{1-q}\right)\right)^{2}}{\left(1-q\right)^{2}}$
and notice that the hypothesis holds for $U_{0}=0$. By the hypothesis,
we have 
\begin{align*}
U_{k+1}^{2} & \le q\frac{\left(\frac{2\eta\alpha c_{1}}{\sqrt{M}}+(1-q)\eta\left(c_{4}+\frac{1}{1-q}\right)\right)^{2}}{\left(1-q\right)^{2}}+\frac{2\eta\alpha c_{1}}{\sqrt{M}}\frac{\left(\frac{2\eta\alpha c_{1}}{\sqrt{M}}+(1-q)\eta\left(c_{4}+\frac{1}{1-q}\right)\right)}{\left(1-q\right)}+\eta^{2}\left(c_{4}+\frac{1}{1-q}\right)\\
 & =q\frac{\left(\frac{2\eta\alpha c_{1}}{\sqrt{M}}+(1-q)\eta\left(c_{4}+\frac{1}{1-q}\right)\right)^{2}}{\left(1-q\right)^{2}}+\frac{1}{1-q}\left(\frac{2\eta\alpha c_{1}}{\sqrt{M}}\right)^{2}+\frac{2\eta\alpha c_{1}}{\sqrt{M}}\eta\left(c_{4}+\frac{1}{1-q}\right)+\eta^{2}\left(c_{4}+\frac{1}{1-q}\right)\\
 & =q\frac{\left(\frac{2\eta\alpha c_{1}}{\sqrt{M}}+(1-q)\eta\left(c_{4}+\frac{1}{1-q}\right)\right)^{2}}{\left(1-q\right)^{2}}\\
 & +\frac{1-q}{\left(1-q\right)^{2}}\left[\left(\frac{2\eta\alpha c_{1}}{\sqrt{M}}\right)^{2}+(1-q)\frac{2\eta\alpha c_{1}}{\sqrt{M}}\eta\left(c_{4}+\frac{1}{1-q}\right)+(1-q)\eta^{2}\left(c_{4}+\frac{1}{1-q}\right)\right]\\
 & \le q\frac{\left(\frac{2\eta\alpha c_{1}}{\sqrt{M}}+(1-q)\eta\left(c_{4}+\frac{1}{1-q}\right)\right)^{2}}{\left(1-q\right)^{2}}\\
 & +\frac{1-q}{\left(1-q\right)^{2}}\left[\left(\frac{2\eta\alpha c_{1}}{\sqrt{M}}\right)^{2}+(1-q)\frac{2\eta\alpha c_{1}}{\sqrt{M}}\eta\left(c_{4}+\frac{1}{1-q}\right)+(1-q)^{2}\eta^{2}\left(c_{4}+\frac{1}{1-q}\right)^{2}\right]\\
 & =q\frac{\left(\frac{2\eta\alpha c_{1}}{\sqrt{M}}+(1-q)\eta\left(c_{4}+\frac{1}{1-q}\right)\right)^{2}}{\left(1-q\right)^{2}}+(1-q)\frac{\left(\frac{2\eta\alpha c_{1}}{\sqrt{M}}+(1-q)\eta\left(c_{4}+\frac{1}{1-q}\right)\right)^{2}}{\left(1-q\right)^{2}}\\
 & =\frac{\left(\frac{2\eta\alpha c_{1}}{\sqrt{M}}+(1-q)\eta\left(c_{4}+\frac{1}{1-q}\right)\right)^{2}}{\left(1-q\right)^{2}},
\end{align*}
where the last second inequality holds by $(1-q)\left(c_{4}+\frac{1}{1-q}\right)\ge1$.
Thus we complete the argument of induction and we have, for any $k$,
\begin{align*}
U_{k}^{2} & \le\frac{\left(\frac{2\eta\alpha c_{1}}{\sqrt{M}}+(1-q)\eta\left(c_{4}+\frac{1}{1-q}\right)\right)^{2}}{\left(1-q\right)^{2}}\\
 & \le2\frac{\frac{4\eta^{2}\alpha^{2}c_{1}^{2}}{M}+2(1-q)^{2}\eta^{2}\left(c_{4}+\frac{1}{1-q}\right)^{2}}{\left(1-q\right)^{2}}\\
 & =\frac{2\alpha^{2}c_{1}^{2}}{(L-\alpha c_{2}-\alpha c_{3})^{2}}\frac{1}{M}+4\eta^{2}\left(c_{4}+2\eta(L-\alpha c_{2}-\alpha c_{3})\right)^{2}.
\end{align*}
And it implies that $\mathbb{W}_{2}^{2}[\rho_{k},\tilde{\rho}_{k}]\le u_{k}\le U_{k}\le\frac{2\alpha^{2}c_{1}^{2}}{(L-\alpha c_{2}-\alpha c_{3})^{2}}\frac{1}{M}+4\eta^{2}\left(c_{4}+2\eta(L-\alpha c_{2}-\alpha c_{3})\right)^{2}.$

\subsection{Proof of Technical Lemmas}

\subsubsection{Proof of Lemma \ref{lem: twid}}

For the first part:
\begin{align*}
 & \left\Vert \nabla V(\th)\right\Vert \\
\le & \left\Vert \nabla V(\th)-\nabla V(\boldsymbol{0})\right\Vert +\left\Vert \nabla V(\boldsymbol{0})\right\Vert \\
\le & b_{1}\left(\left\Vert \th_{1}\right\Vert +1\right).
\end{align*}
For the second part:
\begin{align*}
 & \left\Vert \protect\th-\eta\nabla V(\protect\th)\right\Vert \protect\\
= & \left\langle \protect\th-\eta\nabla V(\protect\th),\protect\th-\eta\nabla V(\protect\th)\right\rangle \protect\\
= & \left\Vert \protect\th\right\Vert ^{2}+2\eta\left\langle \protect\th,-\nabla V(\protect\th)\right\rangle +\eta^{2}\left\Vert \nabla V(\protect\th)\right\Vert ^{2}\protect\\
\le & \left\Vert \protect\th\right\Vert ^{2}+2\eta\left(-a_{1}\left\Vert \protect\th\right\Vert ^{2}+b_{1}\right)+\eta^{2}b_{1}(1+\left\Vert \protect\th\right\Vert ^{2})\protect\\
= & \left(1-2\eta a_{1}+\eta^{2}b_{1}\right)\left\Vert \protect\th\right\Vert ^{2}+\eta^{2}b_{1}+2\eta b_{1}.
\end{align*}

\subsubsection{Proof of Lemma \ref{lem: RBF}}

It is obvious that $\left\Vert K\right\Vert _{\infty,\infty}\le1$.

\begin{align*}
 & \left\Vert K(\th',\th_{1})-K(\th',\th_{2})\right\Vert \\
 & \left\Vert e^{-\left\Vert \th'-\th_{1}\right\Vert ^{2}/\sigma}-e^{-\left\Vert \th'-\th_{2}\right\Vert ^{2}/\sigma}\right\Vert \\
\le & \left\Vert e^{-(\cdot)^{2}/\sigma}\right\Vert _{\mathrm{Lip}}\left\Vert \th_{1}-\th_{2}\right\Vert _{2}.
\end{align*}
And
\begin{align*}
 & \left\Vert \nabla_{\th'}K(\th',\th_{1})-\nabla_{\th'}K(\th',\th_{2})\right\Vert \\
= & \left\Vert \frac{2}{\sigma}e^{-\left\Vert \th'-\th_{1}\right\Vert ^{2}/\sigma}\left(\th'-\th_{1}\right)-\frac{2}{\sigma}e^{-\left\Vert \th'-\th_{2}\right\Vert ^{2}/\sigma}\left(\th'-\th_{2}\right)\right\Vert \\
\le & \left\Vert \frac{2}{\sigma}e^{-(\cdot)^{2}/\sigma}(\cdot)\right\Vert _{\mathrm{Lip}}\left\Vert \th_{1}-\th_{2}\right\Vert _{2}.
\end{align*}

\subsubsection{Proof of Lemma \ref{lem: stein1}}

For any distribution $\rho$ such that $\int\left\Vert \nabla_{\th}V(\th)\right\Vert \rho(\th)d\th<\infty$,
\begin{align*}
 & \left\Vert \phi[\rho](\th_{1})-\phi[\rho](\th_{2})\right\Vert \\
= & \left\Vert \mathbb{E}_{\th\sim\rho}\left\{ -\left[K(\th,\th_{1})-K(\th,\th_{2})\right]\nabla V(\th)+\nabla_{1}K(\th,\th_{1})-\nabla_{1}K(\th,\th_{2})\right\} \right\Vert \\
\le & \left\Vert e^{-(\cdot)^{2}/\sigma}\right\Vert _{\mathrm{Lip}}\mathbb{E}_{\th\sim\rho}\left\Vert \nabla V(\th)\right\Vert \left\Vert \th_{1}-\th_{2}\right\Vert _{2}\\
+ & \left\Vert \frac{2}{\sigma}e^{-(\cdot)^{2}/\sigma}(\cdot)\right\Vert _{\mathrm{Lip}}\left\Vert \th_{1}-\th_{2}\right\Vert _{2}.
\end{align*}
For proving the second result, we notice that 
\begin{align*}
\left\Vert \phi[\rho](\th)\right\Vert  & =\mathbb{E}_{\th'\sim\rho}\left[K(\th',\th)\nabla V(\th')+\nabla_{1}K(\th',\th)\right]\\
 & \le\left\Vert K\right\Vert _{\infty}\mathbb{E}_{\th'\sim\rho}\left[\left\Vert \nabla V(\th')\right\Vert +\frac{2}{\sigma}\left(\left\Vert \th'\right\Vert +\left\Vert \th\right\Vert \right)\right]\\
 & \le\left\Vert K\right\Vert _{\infty}b_{1}+\mathbb{E}_{\th'\sim\rho}\left[\left(\frac{2}{\sigma}+b_{1}\right)\left\Vert \th'\right\Vert +\left\Vert \th\right\Vert \right].
\end{align*}

\subsubsection{Proof of Lemma \ref{lem: stein2}}

Given any $\th'$,

\begin{align*}
 & \sum_{i=1}^{d}\left\Vert \bar{\phi}_{\th',i}(\th)\right\Vert _{\mathrm{Lip}}^{2}\\
= & \sum_{i=1}^{d}\left[\underset{\th_{1}\neq\th_{2}}{\sup}\frac{\left|\bar{\phi}_{\th',i}(\th_{1})-\bar{\phi}_{\th',i}(\th_{2})\right|}{\left\Vert \th_{1}-\th_{2}\right\Vert _{2}}\right]^{2}\\
= & \sum_{i=1}^{d}\underset{\th_{1}\neq\th_{2}}{\sup}\frac{\left|\bar{\phi}_{\th',i}(\th_{1})-\bar{\phi}_{\th',i}(\th_{2})\right|^{2}}{\left\Vert \th_{1}-\th_{2}\right\Vert _{2}^{2}}\\
\le & 2\sum_{i=1}^{d}\underset{\th_{1}\neq\th_{2}}{\sup}\frac{\left|\left(e^{-\left\Vert \th'-\th_{1}\right\Vert ^{2}/\sigma}-e^{-\left\Vert \th'-\th_{2}\right\Vert ^{2}/\sigma}\right)\frac{\partial}{\partial\th_{i}'}V(\th')\right|^{2}}{\left\Vert \th_{1}-\th_{2}\right\Vert _{2}^{2}}\\
+ & 2\sum_{i=1}^{d}\underset{\th_{1}\neq\th_{2}}{\sup}\frac{\left|\frac{2}{\sigma}e^{-\left\Vert \th'-\th_{1}\right\Vert ^{2}/\sigma}(\th_{1,i}-\th_{i}')-\frac{2}{\sigma}e^{-\left\Vert \th'-\th_{2}\right\Vert ^{2}/\sigma}(\th_{2,i}-\th_{i}')\right|^{2}}{\left\Vert \th_{1}-\th_{2}\right\Vert _{2}^{2}}.
\end{align*}
For the first term on the right side of the inequality, 
\begin{align*}
 & \sum_{i=1}^{d}\underset{\th_{1}\neq\th_{2}}{\sup}\frac{\left|\left(e^{-\left\Vert \th'-\th_{1}\right\Vert ^{2}/\sigma}-e^{-\left\Vert \th'-\th_{2}\right\Vert ^{2}/\sigma}\right)\frac{\partial}{\partial\theta_{i}'}V(\th')\right|^{2}}{\left\Vert \th_{1}-\th_{2}\right\Vert _{2}^{2}}\\
= & \sum_{i=1}^{d}\left|\frac{\partial}{\partial\theta_{i}}V(\th')\right|^{2}\underset{\th_{1}\neq\th_{2}}{\sup}\frac{\left|\left(e^{-\left\Vert \th'-\th_{1}\right\Vert ^{2}/\sigma}-e^{-\left\Vert \th'-\th_{2}\right\Vert ^{2}/\sigma}\right)\right|^{2}}{\left\Vert \th_{1}-\th_{2}\right\Vert _{2}^{2}}\\
= & \left\Vert \nabla V(\th')\right\Vert ^{2}\left\Vert e^{-\left\Vert \cdot\right\Vert ^{2}/\sigma}\right\Vert _{\mathrm{Lip}}^{2}.
\end{align*}
To bound the second term, by the symmetry of each coordinates, we
have
\begin{align*}
 & \sum_{i=1}^{d}\underset{\th_{1}\neq\th_{2}}{\sup}\frac{\left|\frac{2}{\sigma}e^{-\left\Vert \th'-\th_{1}\right\Vert ^{2}/\sigma}(\theta_{1,i}-\theta'_{i})-\frac{2}{\sigma}e^{-\left\Vert \th'-\th_{2}\right\Vert ^{2}/\sigma}(\th_{1,i}-\th'_{i})\right|^{2}}{\left\Vert \th_{1}-\th_{2}\right\Vert _{2}^{2}}\\
= & d\left\Vert \frac{2}{\sigma}e^{-\left\Vert \th\right\Vert ^{2}/\sigma}\theta_{1}\right\Vert _{\mathrm{Lip}}^{2}.
\end{align*}
This finishes the first part of the lemma. 
\begin{align*}
 & \sum_{i=d}^{d}\left\Vert \bar{\phi}_{\th',i}(\th)\right\Vert _{\mathcal{L}_{\infty}}^{2}\\
= & \sum_{i=d}^{d}\left\Vert e^{-\left\Vert \th'-\th\right\Vert ^{2}/\sigma}\left(\frac{2}{\sigma}\th_{i}-\frac{2}{\sigma}\th_{i}'-\frac{\partial}{\partial\th_{i}'}V(\th')\right)\right\Vert _{\mathcal{L}_{\infty}}^{2}\\
\le & \sum_{i=d}^{d}2\left\Vert \frac{2}{\sigma}e^{-\left\Vert \th'-\th\right\Vert ^{2}/\sigma}\left(\th_{i}-\th_{i}'\right)\right\Vert _{\mathcal{L}_{\infty}}^{2}+\sum_{i=d}^{d}2\left\Vert e^{-\left\Vert \th'-\th\right\Vert ^{2}/\sigma}\frac{\partial}{\partial\th_{i}'}V(\th')\right\Vert _{\mathcal{L}_{\infty}}^{2}\\
\le & \sum_{i=d}^{d}2\left\Vert \frac{2}{\sigma}e^{-\left\Vert \th'-\th\right\Vert ^{2}/\sigma}\left(\th_{i}-\th_{i}'\right)\right\Vert _{\mathcal{L}_{\infty}}^{2}+2\left\Vert e^{-\left\Vert \cdot\right\Vert ^{2}/\sigma}\right\Vert _{\mathcal{L}_{\infty}}^{2}\left\Vert \nabla V(\th')\right\Vert ^{2}\\
\le & 2d\left\Vert \frac{2}{\sigma}e^{-\left\Vert \th\right\Vert ^{2}/\sigma}\theta_{1}\right\Vert _{\mathcal{L}_{\infty}}^{2}+2\left\Vert e^{-\left\Vert \cdot\right\Vert ^{2}/\sigma}\right\Vert _{\mathcal{L}_{\infty}}^{2}\left\Vert \nabla V(\th')\right\Vert ^{2}.
\end{align*}

\end{document}